\documentclass{hld2024} % Include author names
\usepackage{xcolor}

\title[Reactivation: Empirical NTK Dynamics Under Task Shifts]{Reactivation: Empirical NTK Dynamics Under Task Shifts}

\hldauthor{\Name{Yuzhi Liu}\textsuperscript{†}
\thanks{Equal contribution. Email \{yuzliu, chenzixu, zirzhang\}@student.ethz.ch, yufei.liu@inf.ethz.ch}\\
\Name{Zixuan Chen}\textsuperscript{‡} \footnotemark[1] \\
\Name{Zirui Zhang}\textsuperscript{‡} \footnotemark[1] \\
\Name{Yufei Liu}\textsuperscript{‡} \footnotemark[1]\\ 
\Name{Giulia Lanzillotta}\textsuperscript{‡} \\
\addr{\textsuperscript{†}Department of Mathematics, ETH Zurich, Switzerland}\\
\addr{\textsuperscript{‡}Department of Computer Science, ETH Zurich, Switzerland}
}
\begin{document}

\maketitle

\vspace{-5mm}
\begin{abstract}%
The Neural Tangent Kernel (NTK) offers a powerful tool to study the functional dynamics of neural networks. In the so-called lazy, or kernel regime, the NTK remains static during training and the network function is linear in the static neural tangents feature space. 
The evolution of the NTK during training is necessary for feature learning, a key driver of deep learning success. The study of the NTK dynamics has led to several critical discoveries in recent years, in generalization and scaling behaviours. However, this body of work has been limited to the single task setting, where the data distribution is assumed constant over time. In this work, we present a comprehensive empirical analysis of NTK dynamics in continual learning, where the data distribution shifts over time. Our findings  highlight continual learning as a rich and underutilized testbed for probing the dynamics of neural training. At the same time, they challenge the validity of static-kernel approximations in theoretical treatments of continual learning, even at large scale.

\end{abstract}

%\begin{keywords}%
%  List of keywords%
%\end{keywords}

\section{Introduction}
\vspace{-1mm}
Continual learning is central to real-world applications where models must learn from a stream of tasks without retraining from scratch or forgetting previous knowledge. While architectural and algorithmic advances have improved performance in such settings, our theoretical understanding of how and why continual learning works remains limited. In particular, the dominant tools for analyzing neural network learning dynamics—developed primarily for stationary, single-task settings—may not capture the behaviors that emerge when data distributions shift over time.

The study of learning dynamics in neural networks seeks to characterize how model parameters, internal representations, and predictions evolve throughout training. This perspective reveals the implicit biases of gradient-based optimization and has helped explain phenomena such as generalization, feature learning, and convergence. 

In recent years, powerful tools have emerged for analyzing learning dynamics. The Neural Tangent Kernel (NTK) framework \citep{Jacot_NEURIPS2018_NTK}, for example, shows that infinitely wide neural networks behave like kernel machines with fixed kernels throughout training. However, practical networks typically operate in finite-width regimes, where the NTK evolves during training, enabling feature learning. \citet{fort2020deep} and \citet{zhou2025coneeffectlearningdynamics} empirically demonstrated that NTKs in realistic architectures undergo substantial change at early training stages, correlating with stronger representation learning and improved performance. This has led to a conceptual dichotomy between the \textbf{lazy (kernel)} regime—where internal representations remain fixed—and the \textbf{rich (feature learning)} regime—where features evolve to adapt to data structure.

Beyond the lazy/rich distinction, recent studies have explored how NTKs evolve. A key phenomenon is \emph{kernel alignment}, where NTK eigenvectors align with task-relevant directions over time, improving learning efficiency and generalization \citep{baratin2021implicit, shan2021theory}. In parallel, works on loss landscape geometry reveal \emph{progressive sharpening}—a rise in curvature early in training—linked to changes in the NTK through its connection to the Hessian spectrum \citep{Jastrzebski2020break-even, cohen2021gradient}.

Bringing this lens to continual and sequential learning holds significant promise. Understanding the mechanisms that drive interference and forgetting—potentially from minimal assumptions about task similarity— could illuminate their root causes and guide the design of algorithms that are both more robust and more efficient.
Some theoretical works have modeled continual learning in the lazy regime, where NTKs remain static \citep{karakida2022learningcurvescontinuallearning, doan2021theoreticalanalysiscatastrophicforgetting, bennani2020generalisation}. While analytically convenient, these assumptions overlook the dynamic, evolving nature of representations in realistic networks, leaving a gap between theory and observed behavior.

A deeper challenge is that most existing theory assumes \emph{stationarity}, i.e., that training data is drawn from a fixed distribution. This assumption breaks down in continual learning, where tasks and data distributions shift over time. This raises a central and largely unanswered question:
%\lyz{last two paragraphs -- more precise}
\begin{quote}
\textit{Do existing theories of learning dynamics—particularly those based on the NTK—extend meaningfully to non-stationary environments?}
\end{quote}

\vspace{-2mm}
\subsection{Contributions}
This work provides a systematic, empirical investigation of Neural Tangent Kernel (NTK) dynamics in the context of continual learning—a setting that challenges the conventional assumption of stationary data distributions. Our contributions are as follows:
\begin{enumerate}
    \item We evaluate how NTK dynamics of past data respond to changes in network width, learning rate, training duration, and critically, task similarity, across single and multiple task switches.
    \item We demonstrate that task transitions consistently trigger abrupt shifts in the NTK of past data, even under a lazy learning regime, revealing a reactivation of feature learning at each task boundary.
    \item Through controlled experiments, we distinguish between different types of distributional shifts, showing that the introduction of semantically novel classes leads to significantly greater NTK change.
\end{enumerate}

By systematically characterizing the evolution of NTKs in non-stationary regimes, our experiments provide compelling evidence that challenges the static distribution assumption foundational to the original NTK theory and many previous theoretical works. Indeed, we observe that even in networks with relatively large width, the NTK of past data can undergo abrupt changes when exposed to data from a new distribution—a phenomenon we term \emph{re-activation}. These findings not only reveal critical limitations in existing theoretical frameworks in continual learning but also open up a promising and underexplored direction: developing theories that explicitly account for data distribution shifts. Our work lays the groundwork for advancing continual learning and offers valuable insights into the training dynamics of neural networks.

\section{Experiments and Results}
\label{sec:experiment}

\subsection{Metrics used and parametrization.}
% In this section, we focus on how learning dynamics evolve during task switches. To measure NTK evolution, we track changes in the maximum eigenvalues of NTK matrices, and the kernel distance relative to the initial NTK matrix, using Equation \ref{eq:CKA}. We also employ kernel Alignment (Equation \ref{eq:alignment}) to assess how well the NTK matrix aligns with the training labels in feature learning. To explore how the training phase at the point of task switch influences subsequent learning dynamics, we vary the learning rates and widths across experiments.

We review the definition and some fundamental ideas related to the Neural Tangent Kernel in Appendix \ref{sec:theory}. Here, we introduce the main metrics used in our experiments. We run image classification experiments on CIFAR and ImageNet across multiple seeds. More training details can be found in Appendix \ref{sec:experiment_details}. In the following sections we present results from experiments involving two sequential tasks, and in Appendix \ref{sec:complementary-results}, we report results on more task sequences. Crucially, all metrics are evaluated solely on the data from the first task throughout the continual experiment.
\paragraph{Kernel Spectral Norm} It is equivalent to the max eigenvalue of NTK. We show in Appendix \ref{sec:ntk_eigenvalues} that the NTK spectral norm controls the convergence rate in certain eigenmodes.

\paragraph{Kernel Distance} Following \citet{fort2020deep,cortes2012algorithms,kornblith2019similarity}, we define the kernel distance based on Centered Kernel Alignment $\operatorname{CKA}(\cdot, \cdot)$ (definition in the Appendix \ref{appendix:cka_def}) as $S(\Theta, \Theta') \triangleq 1 - \operatorname{CKA}(\Theta, \Theta')$.
\paragraph{Kernel Velocity} The kernel velocity $v(t) \triangleq S\big(\Theta_t, \Theta_{t + dt}\big)/dt$ quantifies the  rate of change of NTKs at time $t$.
%This metric provides insights into different learning regimes: small kernel velocity corresponds to the \emph{lazy} regime with nearly constant kernels, while larger kernel velocity indicates \emph{rich} feature learning dynamics.
\paragraph{Kernel Alignment} The kernel alignment $
    A(t)\triangleq \operatorname{CKA}(\Theta_t,\mathbf{y}\mathbf{y}^\top )$\cite{cortes2012algorithms} measures the similarity between the NTK and the target label kernel $\mathbf{y}\mathbf{y}^\top$ at time $t$, where $\mathbf{y}$ is the label vector.

\paragraph{Parametrization} We consider the \emph{Pytorch standard parametrization} in our experiments, which we operate under two regimes: a lazy learning regime, obtained by scaling the learning rate inversely with the width and a feature learning regime, for which we employ the \emph{Kaiming uniform} initialization, which reflects common practice in continual learning. 
We review in detail the parametrization used in Appendix \ref{subsec:net-param}.

\subsection{Task Shifts Reactivate Learning Dynamics}
\label{sec: reactivation}

% \begin{figure}[h!]
%     \centering
%     % 第一排
%     \subfigure[Accuracy]{
%         \includegraphics[height=0.21\textwidth]{images/kaiming/width_comparison_accuracy_cnn_CIFAR10_inc5-5_e160_b32_kaiming_sgd_s32.pdf}
%     }
%     \hfill
%     \subfigure[Alignment]{
%         \includegraphics[height=0.2\textwidth]{images/kaiming/width_comparison_alignment_cnn_CIFAR10_inc5-5_e160_b32_kaiming_sgd_s32.pdf}
%     }
%     \hfill
%     \subfigure[Kernel Distance]{
%         \includegraphics[height=0.21\textwidth]{images/kaiming/width_comparison_cka_cnn_CIFAR10_inc5-5_e160_b32_kaiming_sgd_s32.pdf}
%     }

%     % 第二排
%     \subfigure[Max Eigenvalue]{
%         \includegraphics[height=0.19\textwidth]{images/kaiming/width_comparison_ntk_max_eigenvalues_cnn_CIFAR10_inc5-5_e160_b32_kaiming_sgd_s32.pdf}
%     }
%     \hspace{0.01\textwidth}
%     \subfigure[Velocity (dt=10)]{
%         \includegraphics[height=0.19\textwidth]{images/kaiming/width_comparison_velocity_dt10_cnn_CIFAR10_inc5-5_e160_b32_kaiming_sgd_s32.pdf}
%     }
%     \hspace{0.01\textwidth}
%     \subfigure[Velocity (dt=10) Zoomed In]{
%         \includegraphics[height=0.2\textwidth]{images/kaiming/width_comparison_velocity_dt10_cnn_CIFAR10_inc5-5_e160_b32_kaiming_sgd_s32_zoom_in.pdf}
%     }

%     \caption{Comparison of different metrics across network widths for CNN trained on CIFAR10 with kaiming setting. The number of epochs per task is set to 160. (a) Test accuracy, (b) Alignment, (c) Kernel distance, (d) Maximum eigenvalue of NTK, (e) Kernel velocity with dt=10, (f) Kernel velocity (zoom in).}
%     \label{fig:kaiming_5}
% \end{figure}

\begin{figure}[h!]
    \vspace{-2mm}
    \centering
    % First row Width comparision
    % First row: Width comparison (feature learning)
    \includegraphics[width=0.30\textwidth]{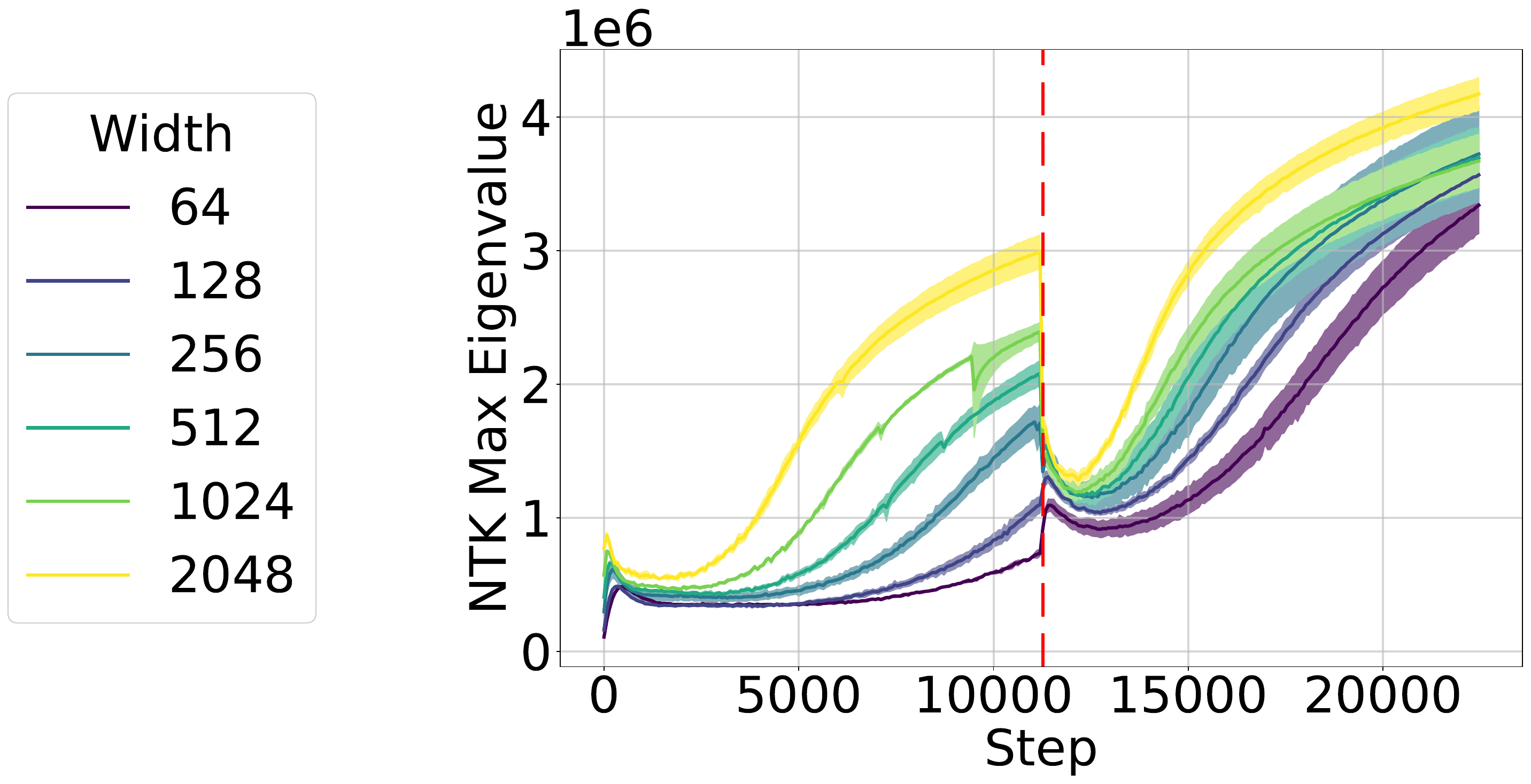}
    \hfill
    \includegraphics[width=0.20\textwidth]{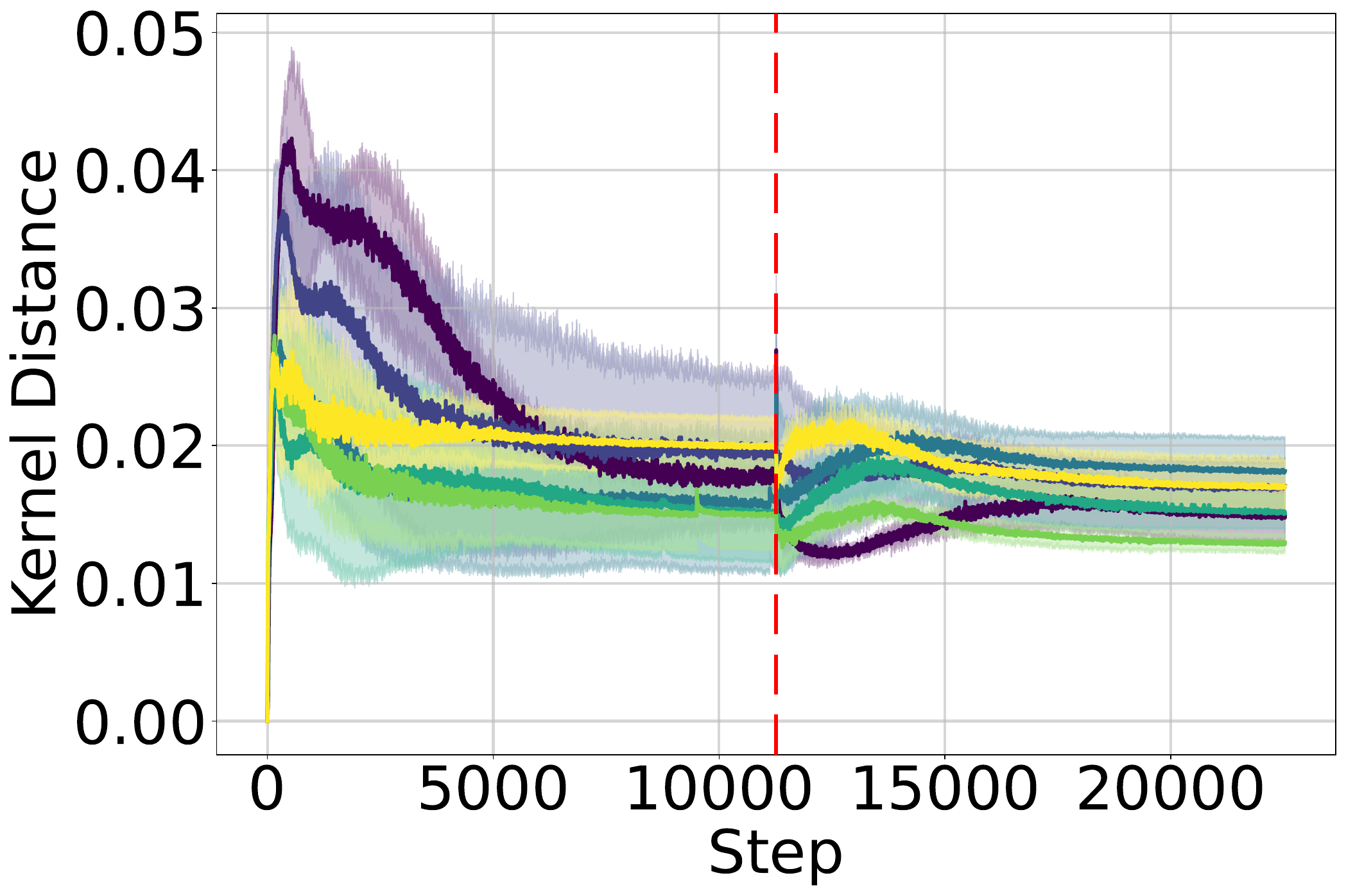}
    \hfill
    \includegraphics[width=0.20\textwidth]{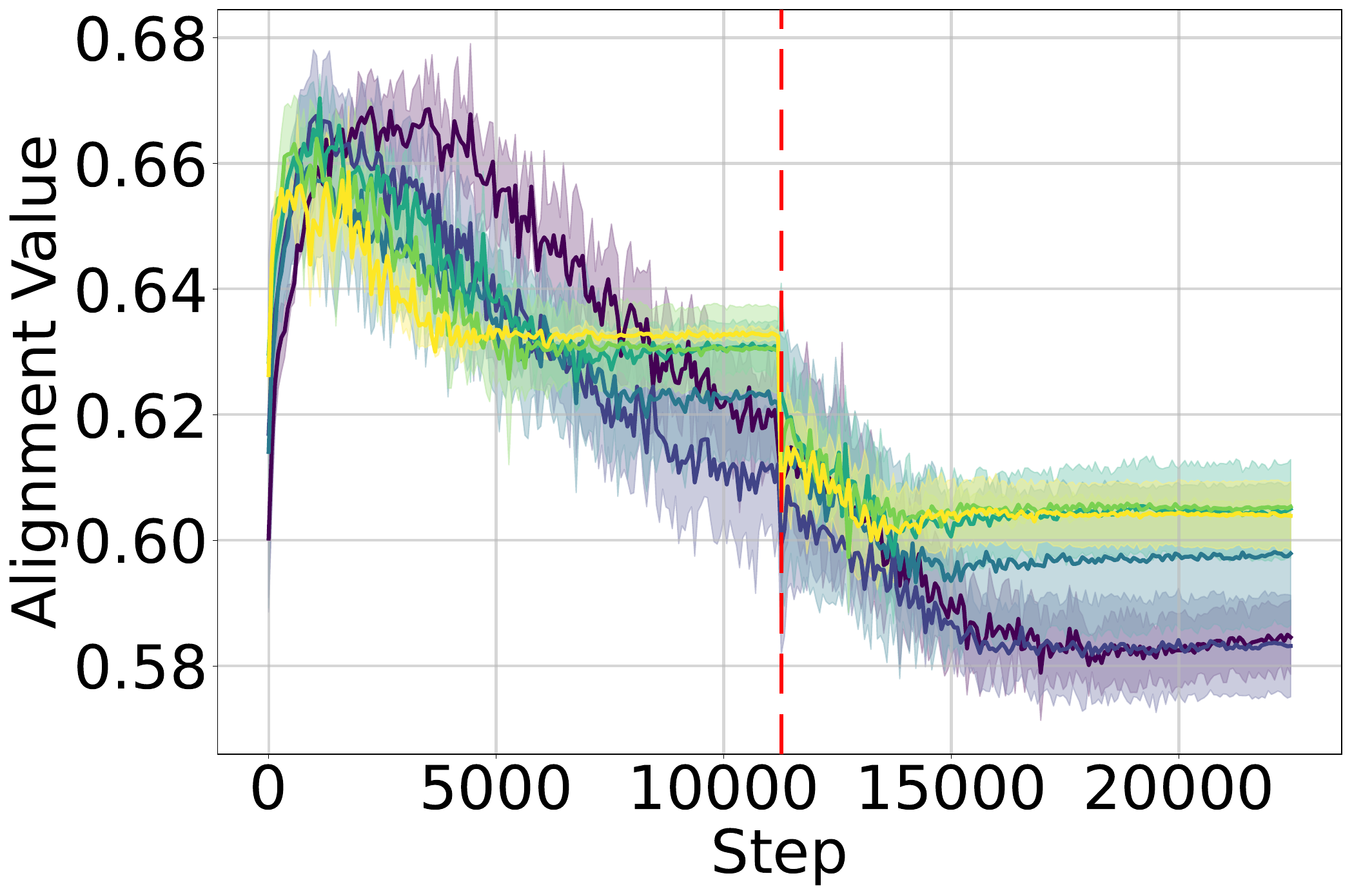}
    \hfill
    \includegraphics[width=0.20\textwidth]{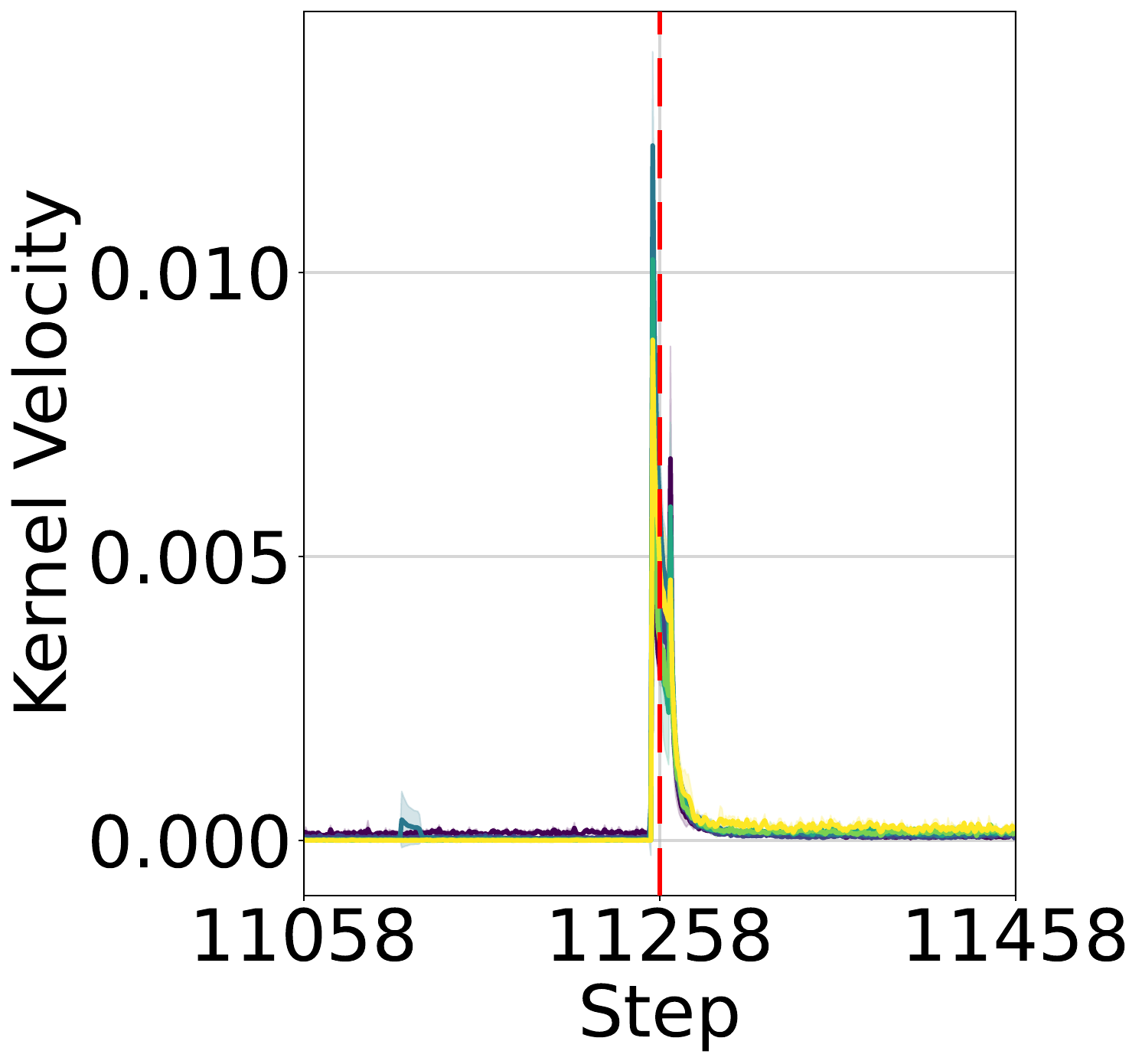}
    
    \vspace{2mm} % vertical space between rows
    
    % Second row: Width comparison (kaiming / no feature learning)
    \includegraphics[width=0.30\textwidth]{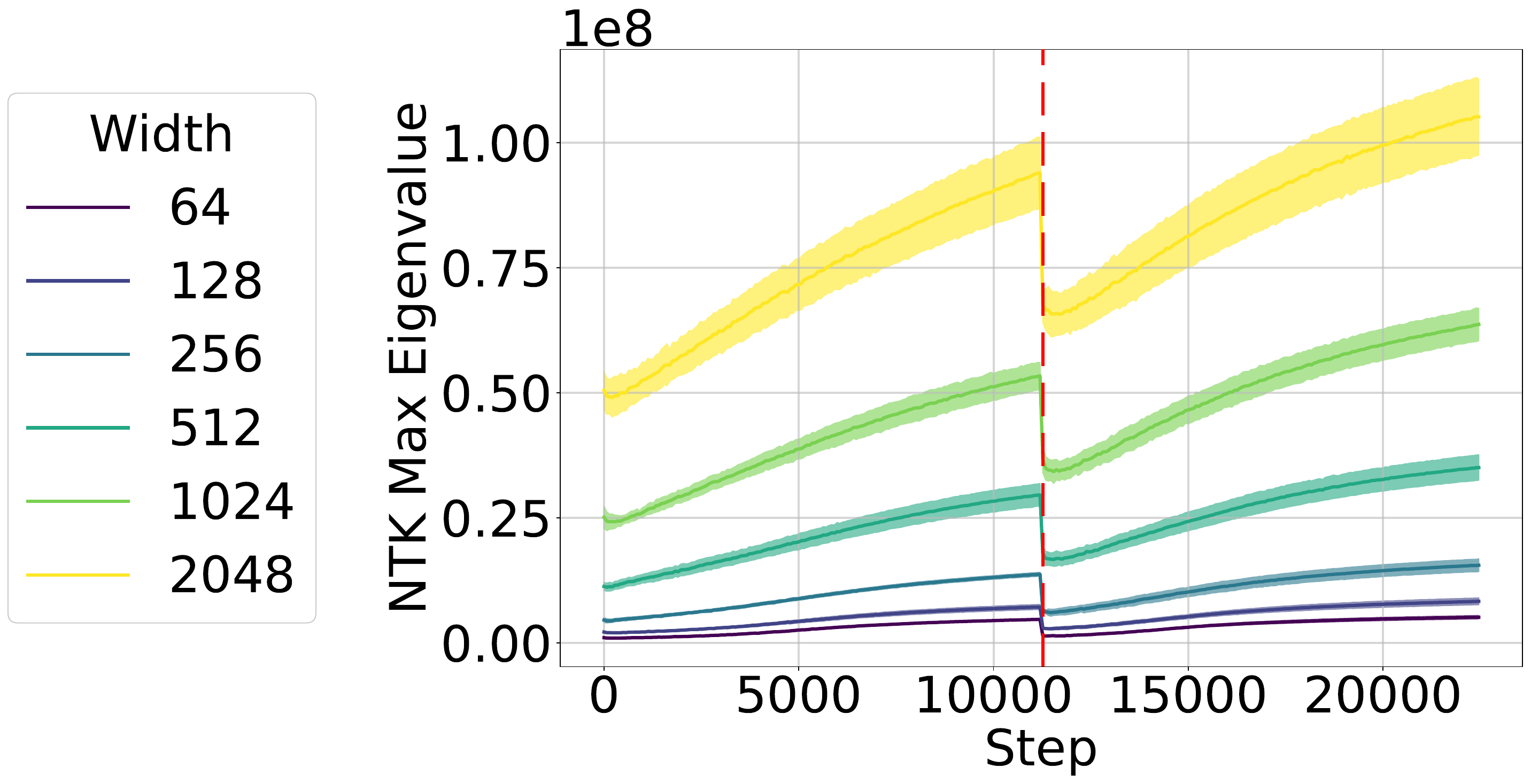}
    \hfill
    \includegraphics[width=0.20\textwidth]{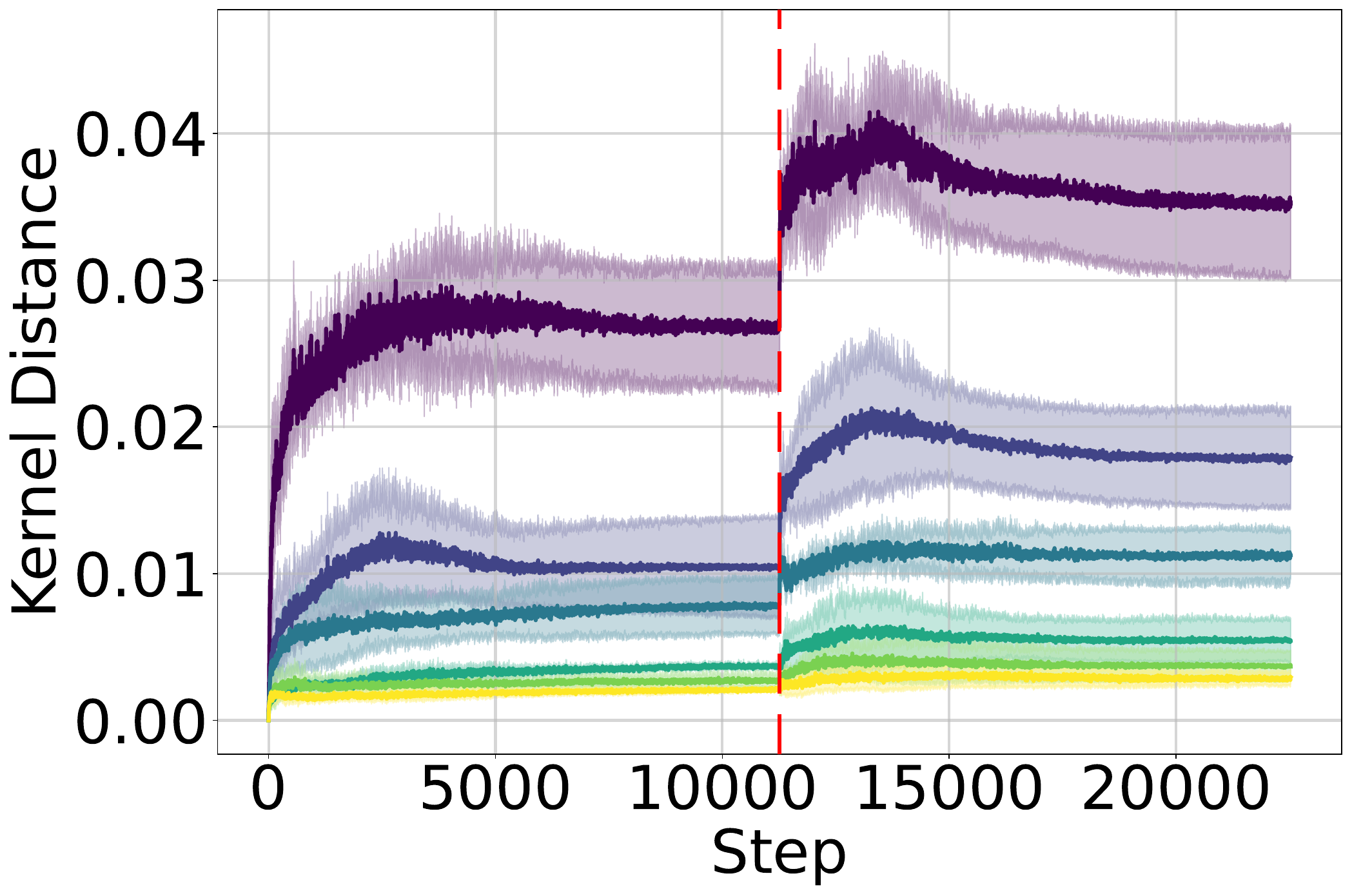}
    \hfill
    \includegraphics[width=0.20\textwidth]{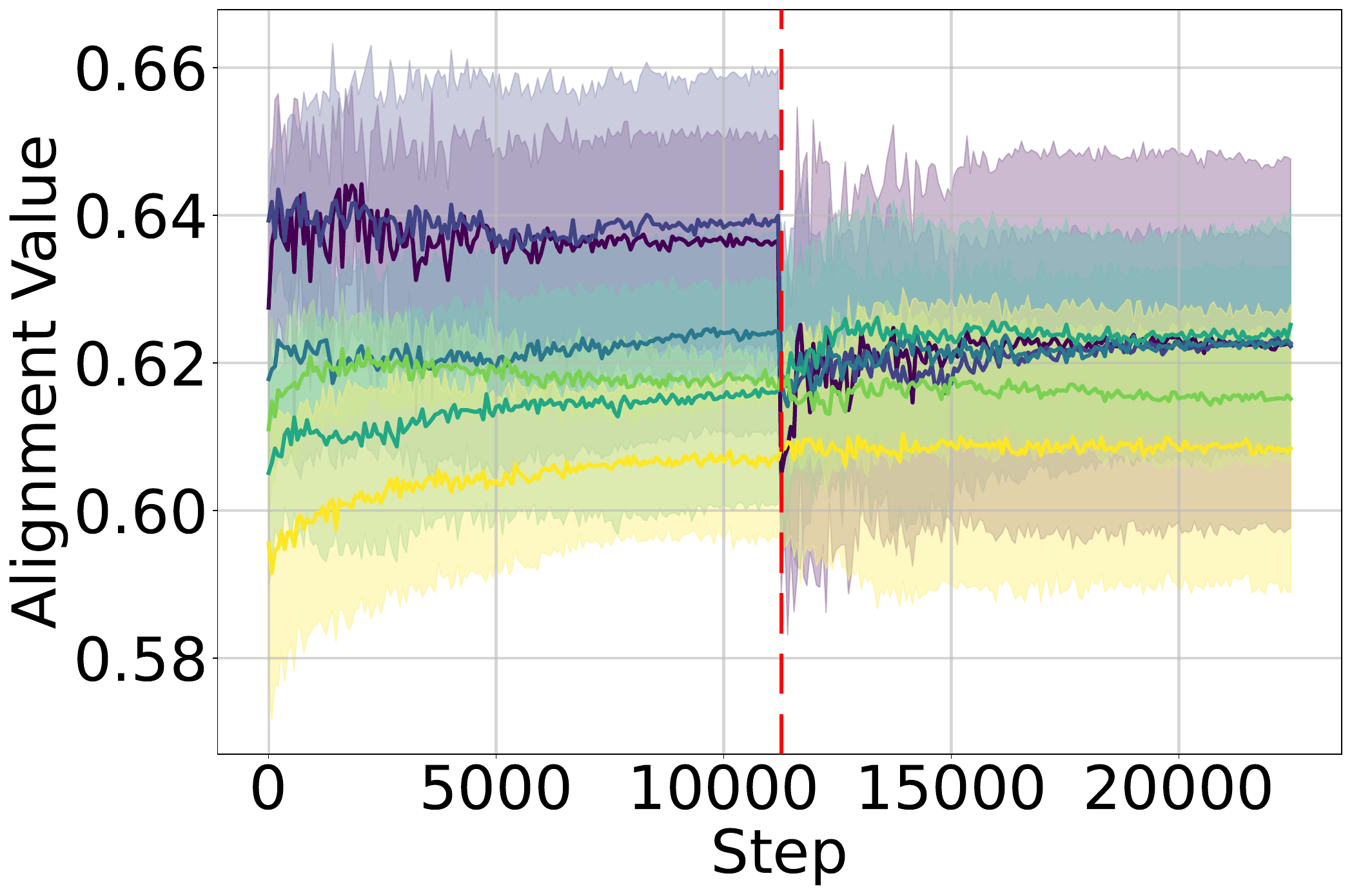}
    \hfill
    \includegraphics[width=0.20\textwidth]{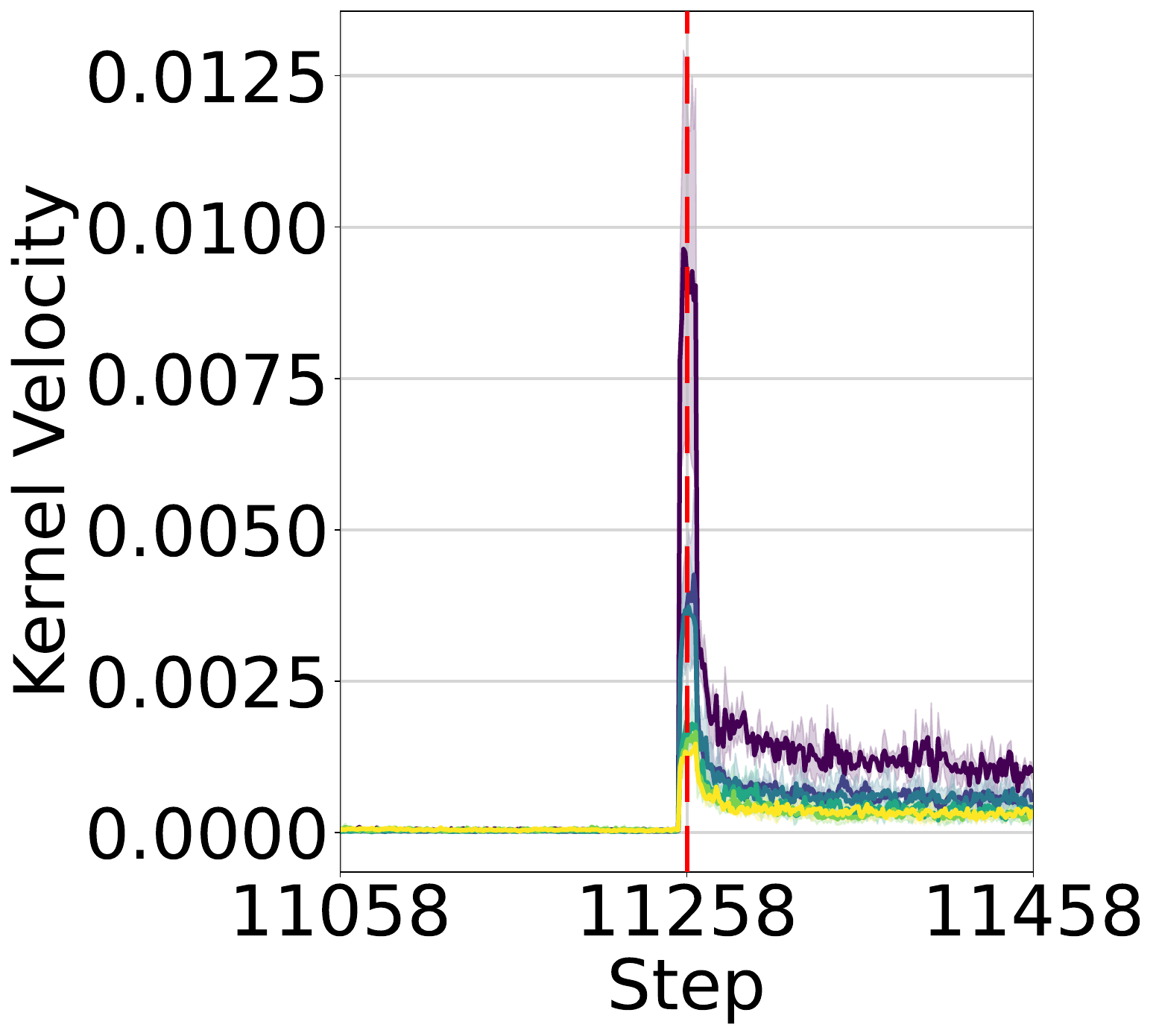}

    % \subfigure[]{
    %     \includegraphics[width=0.3\textwidth]{images/lr_comparison_ntk_max_eigenvalues_cnn_CIFAR10_inc5-5_e160_b32_w2048_sgd_s32.pdf}
    % }
    % \hfill
    % \subfigure[]{
    %     \includegraphics[width=0.15\textwidth]{images/lr_comparison_ntk_max_eigenvalues_cnn_CIFAR10_inc5-5_e160_b32_w2048_sgd_s32_task_switch.pdf}
    % }
    % \hfill
    % \subfigure[]{
    %     \includegraphics[width=0.22\textwidth]{images/lr_comparison_cka_cnn_CIFAR10_inc5-5_e160_b32_w2048_sgd_s32.pdf}
    % }
    % \hfill
    % \subfigure[]{
    %     \includegraphics[width=0.22\textwidth]{images/lr_comparison_alignment_cnn_CIFAR10_inc5-5_e160_b32_w2048_sgd_s32.pdf}
    % }

    %\zixuan{Added e as zoomed in subgraph as d.}
    \caption{Comparison of NTK Max Eigenvalue, Kernel Distance from initialization, Kernel Alignment and Kernel velocity across different widths with (Row 1) and without (Row 2) feature learning. The measurements are done on the first task before and after the task switch, with 5 classes in each task. One step corresponds to 10 iterations during training.}
    %\giulia{Mention on which data the kernel is calculated on: e.g. always the first task, or always the current task. 
    %\lyz{recovering? Maybe just the level of the drop. Cannot say too much about the trend after the drop}}
    %\lyz{(e) x-axis 'zoom-in'?}}
    \label{fig:compare_one_switch}
    %\vspace{-5mm}
\end{figure}

Scaling up neural networks—along with appropriate learning rate rescaling \citep{yang2020tensorprogramsiineural}—is known to induce the lazy regime, in which training occurs in a nearly linear function space and the network’s internal features remain effectively static. 
We confirm this behavior on the first task: increasing model width yields more static NTK, as indicated by reduced kernel distance and velocity during training (Figure~\ref{fig:compare_one_switch}-Column 2; Appendix Figure~\ref{fig:kaiming_5}). However, at the moment of task switch, we observe a clear and consistent spike in kernel velocity, signaling a (temporary) departure from the lazy regime. The network briefly enters a dynamic phase of feature adaptation before quickly returning to stability. We refer to this phenomenon as the \textbf{re-activation} of feature learning.

This reactivation is accompanied by a sharp drop in the NTK norm of the old task at the onset of the new task, followed by a gradual recovery. This creates a distinctive asymmetric V-shape or “check-mark” trajectory in the NTK norm that we observed consistently across all model widths. The timing of this drop aligns with the spike in velocity, suggesting a rapid reconfiguration of the network’s functional representation in response to the new task. Interestingly, we observe the same shape in the feature learning regime, which is closer to common practice. 
Similar patterns are observed in Kernel Alignment (Figure \ref{fig:compare_one_switch}-Column 3), indicating that the NTK rapidly changes direction at the task switch and begins evolving along a new direction.  We also find that these patterns persist for multiple task switches, as shown in Figure \ref{fig:comparison_multiple_switches} in the appendix.

\begin{figure}
    \centering
    \includegraphics[width=0.3\textwidth]{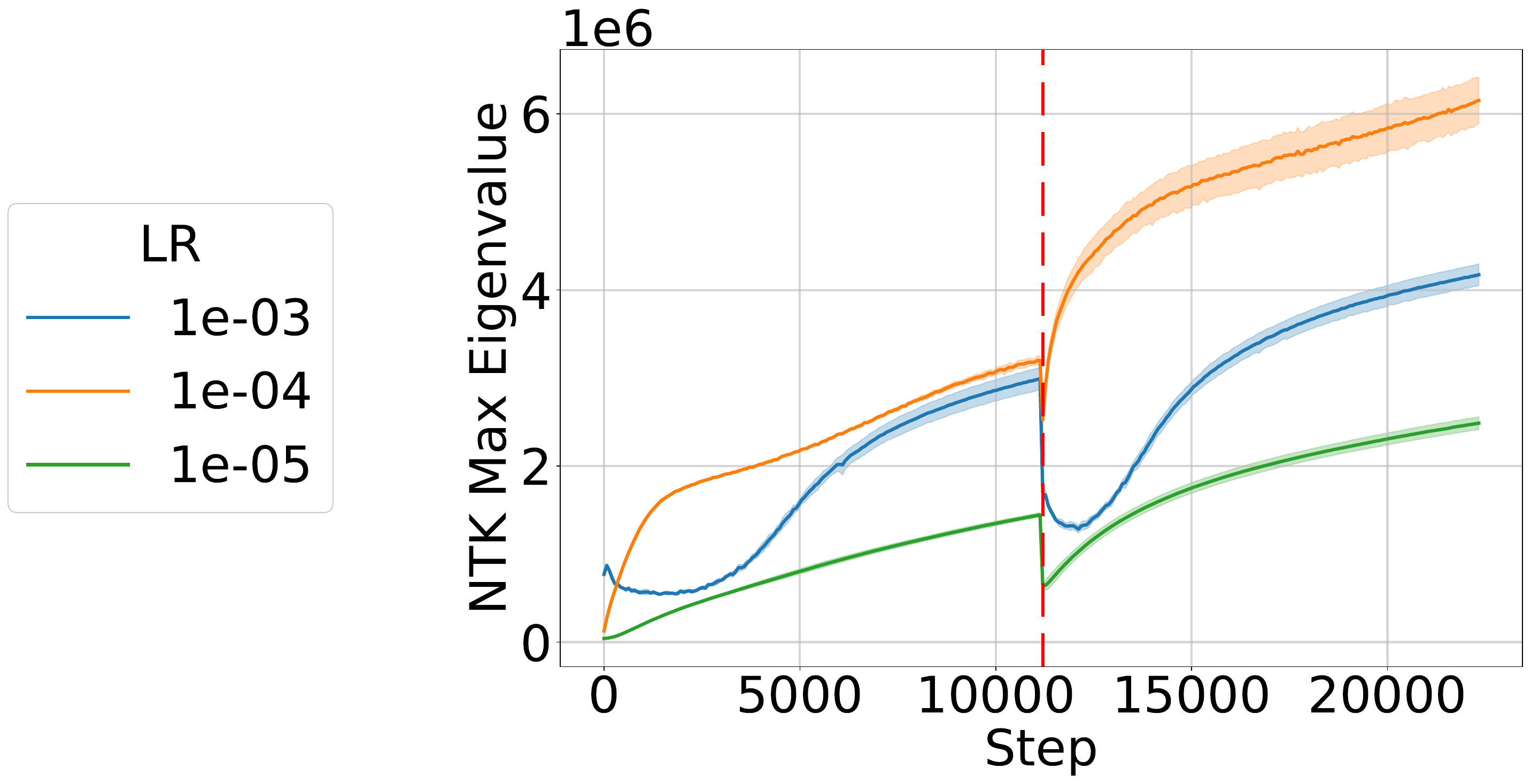}
    \hfill
    \includegraphics[width=0.15\textwidth]{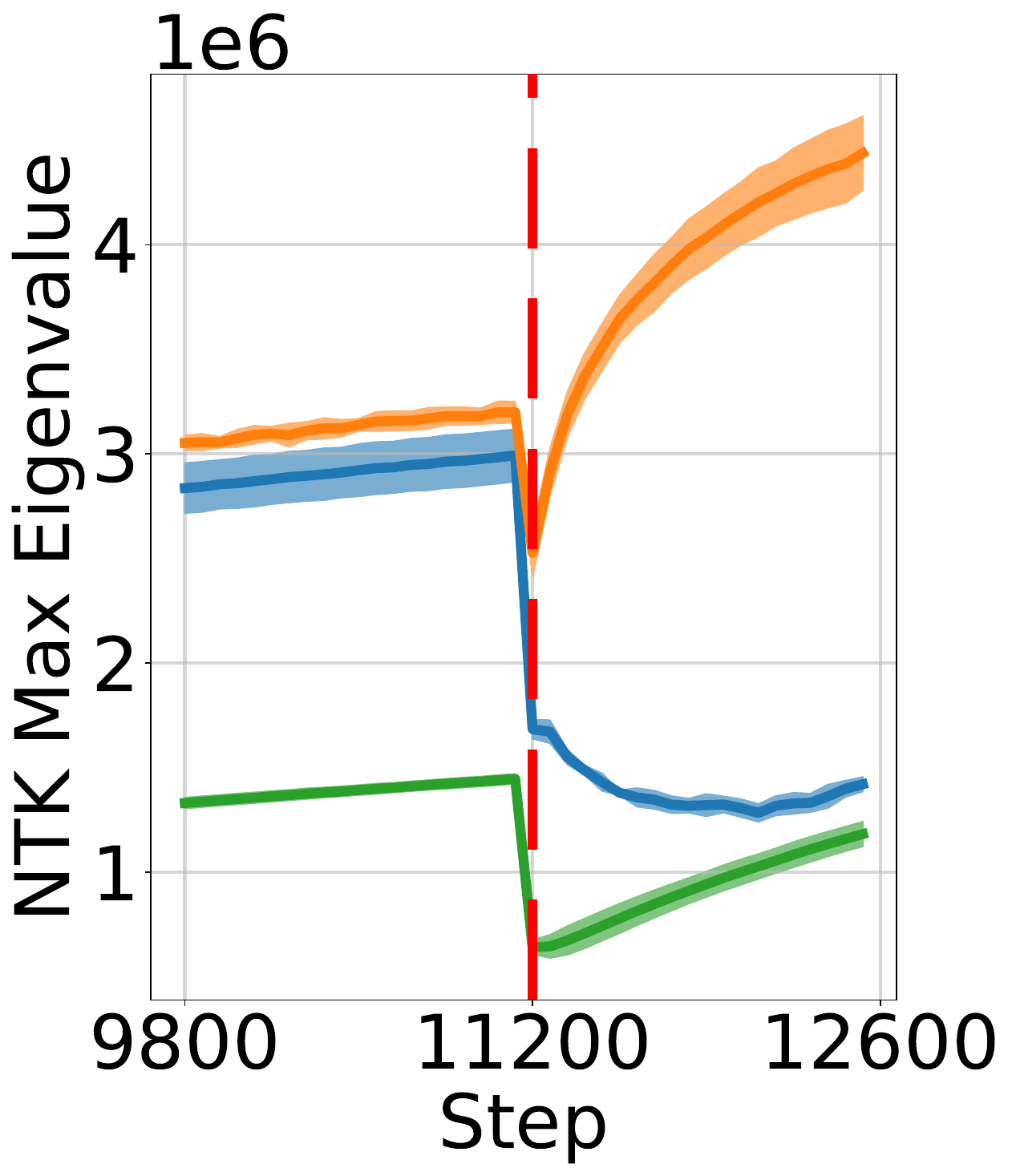}
    \hfill
    \includegraphics[width=0.22\textwidth]{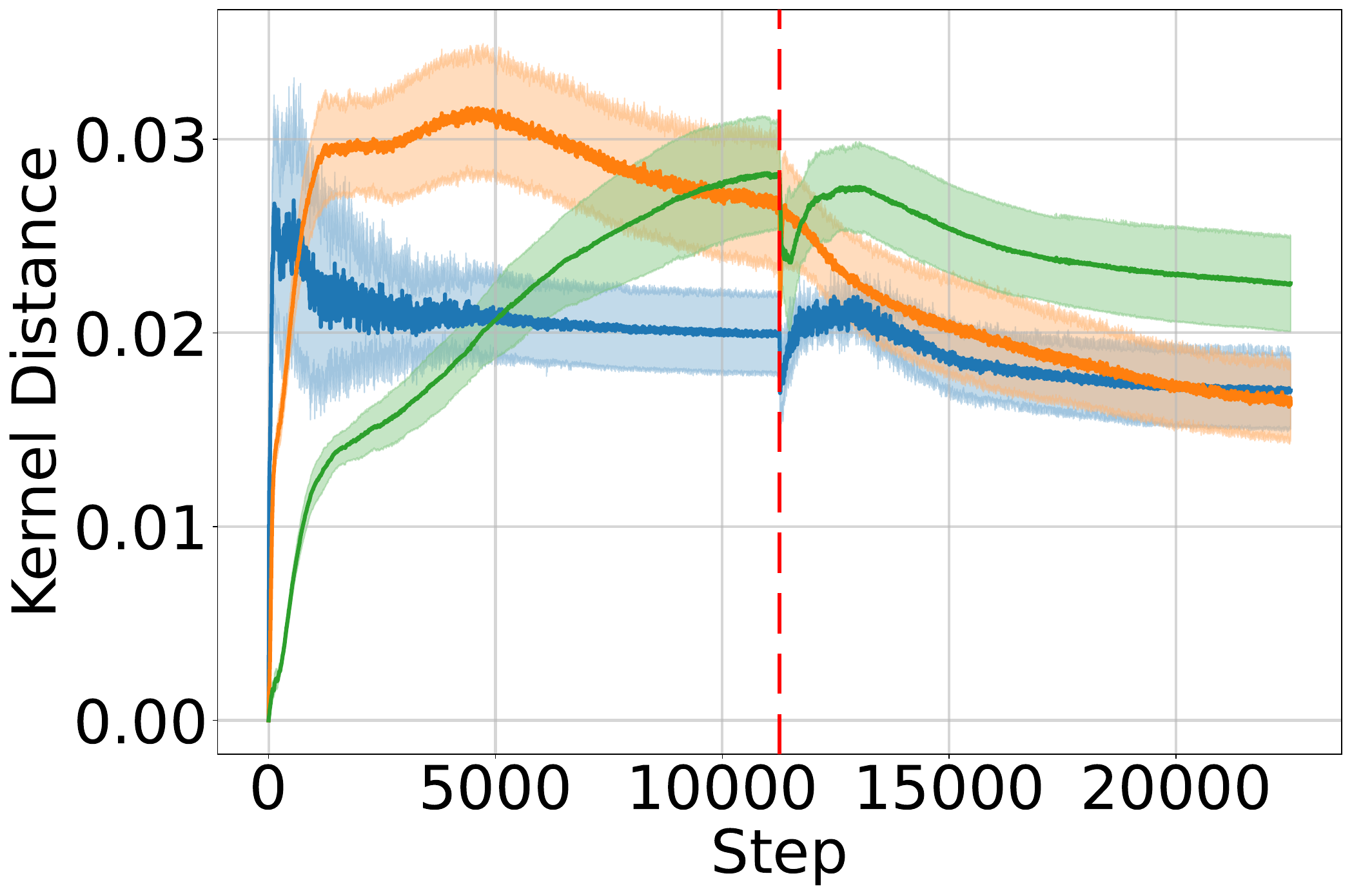}
    \hfill
    \includegraphics[width=0.22\textwidth]{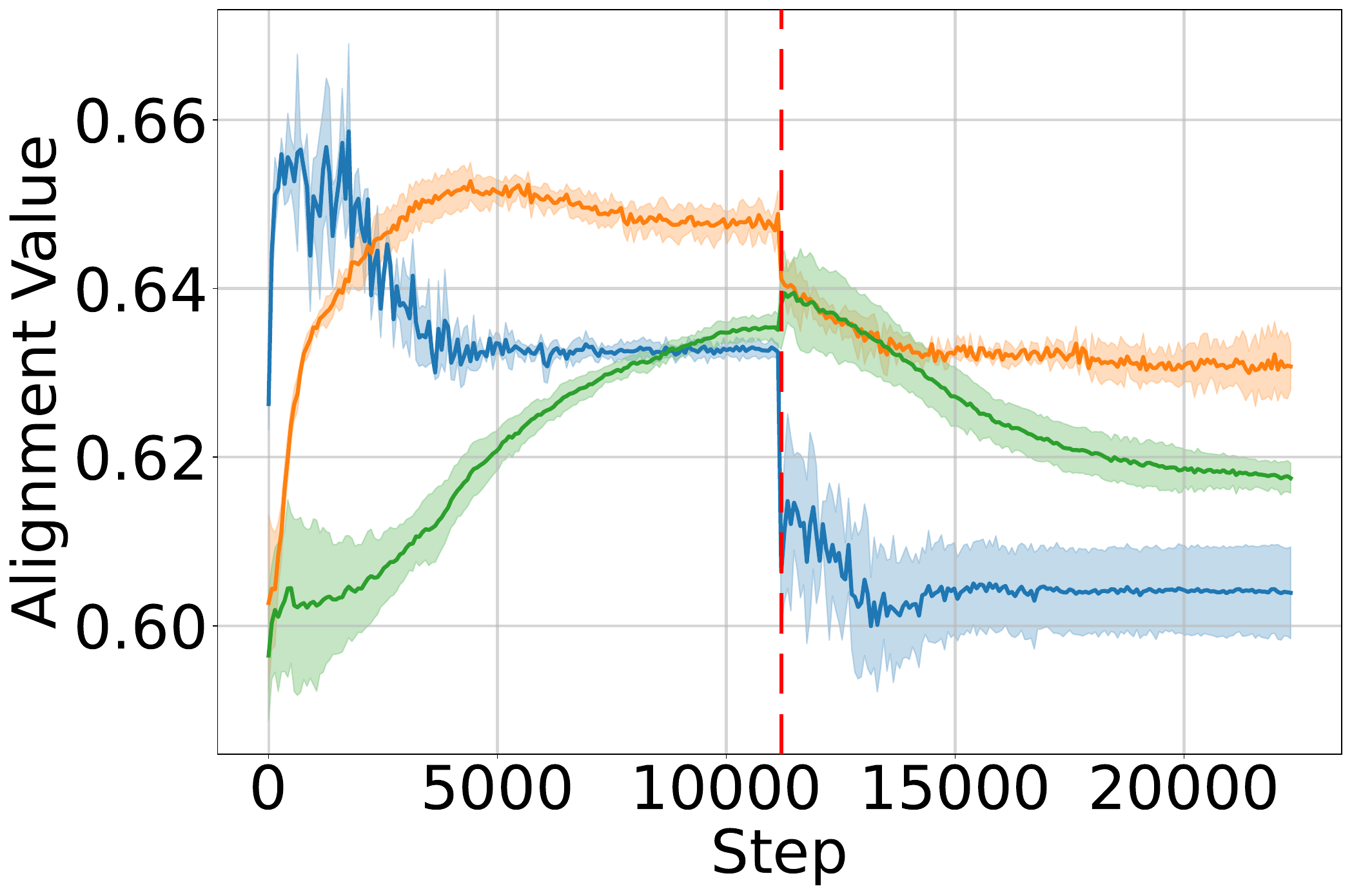}

    %\zixuan{Added e as zoomed in subgraph as d.}
    \caption{Comparison of NTK Max Eigenvalue, Kernel Distance from initialization, Kernel Alignment and Kernel velocity across different learning rates at fixed width ($N=2048$). }
    %\giulia{Mention on which data the kernel is calculated on: e.g. always the first task, or always the current task. 
    %\lyz{recovering? Maybe just the level of the drop. Cannot say too much about the trend after the drop}}
    %\lyz{(e) x-axis 'zoom-in'?}}
    \label{fig:compare_one_switch_lr}
    %\vspace{-5mm}
\end{figure}

We find that the behavior at task switch critically depends on the learning rate. 
%By changing the learning rate, we approximately categorize training phases at the task switch as, respectively, \textit{under-converged}, \textit{converged}, and \textit{over-converged}, corresponding to $lr=\{0.00001,0.0001,$ $0.001\}$, based on  the accuracy curves shown in Figure \ref{fig:lr_0.0001}a.
Notably, the checkmark shape is more visible for higher learning rates, while for lower learning rates,  it is collapsed on the first few steps after the task switch. Moreover, the network recovers from the disruption fastest when using the intermediate learning rate. This difference may stem from the training phase at the task switch: with the smallest learning rate, the network has not yet fully captured general patterns, whereas with the largest learning rate, it has already begun to overfit. Both situations result in a longer recovery time.

\paragraph{Remarks.} It is not obvious that the tangent kernel of the first task should evolve according to a structured pattern as observed. Firstly, prior to this work, it was not clear that the NTK would evolve at all under a lazy learning regime, and several theoretical studies made the unsubstantiated assumption that the lazy regime can be maintained under distribution shifts. Secondly, the robustness of these patterns across architectures and hyperparameters points to the existence of a shared underlying mechanism governing task transitions. This mechanism appears to be modulated by hyperparameters such as network width and learning rate, suggesting a unified framework for characterizing adaptation in continual learning.

% These patterns are consistent across different widths as shown in Figure \ref{fig:compare_one_switch}a. 
% Figure \ref{fig:compare_one_switch}b,e further highlight the re-activation of the learning dynamics at the task transition, with the \textit{converged} phase exhibiting the fastest return to effective feature learning. 

\subsection{Task Similarity Controls NTK Dynamics}

A task switch introduces a shift in the data distribution, which cause the reactivation of feature learning. However, "distribution shift" is a generic term that applies to many different scenarios. In particular, we consider two specific cases of distribution shift: the introduction of new classes, and the change of the relative frequencies of a set of known classes. All the following experiments are conducted in the feature learning regime. Further details can be found in Appendix \ref{sec:experiment_details}. 

In the first case, for each experiment $E_i$, the network is trained on distribution $\mathcal{D}_0$ in task 1 and $\mathcal{D}_i$ in task 2, where $\mathcal{D}_k$ denotes a uniform mixture over 10 classes $\{k, k+1, \ldots, k+9\}$. 
Thus the similarity between $\mathcal{D}_0$ and $\mathcal{D}_i$ can be measured as the overlap between the classes:

\[
\text{Similarity}(\mathcal{D}_0, \mathcal{D}_i) = \frac{|\mathcal{D}_0 \cap \mathcal{D}_i|}{|\mathcal{D}_0 \cup \mathcal{D}_i|}.
\]

By varying $i \in [0,1]$, we sweep the similarity between 1 (identical tasks) and 0 (no class overlap, a typical benchmark for continual learning).

In the second case, we define two disjoint class subsets $\tilde{\mathcal{D}}_0$ and $\tilde{\mathcal{D}}_1$, and interpolate between them with mixtures: $\tilde{\mathcal{D}}_\alpha = (1 - \alpha)\tilde{\mathcal{D}}_0 + \alpha\tilde{\mathcal{D}}_1$.
For each experiment $\tilde{E}_\alpha$ the network is trained on $\tilde{\mathcal{D}}_{0.1}$ in task 1 and $\tilde{\mathcal{D}}_\alpha$ in task 2, 
varying $\alpha \in [0.1,0.9]$. The similarity metric is linear:
\[\text{Similarity}(\tilde{\mathcal{D}}_\alpha,\tilde{\mathcal{D}}_\beta) = 1-|\alpha-\beta|.\]
 
%\vspace{-2mm}
% In this section, we investigate two further questions: (1) Does a greater distributional change cause larger dynamic changes? and (2) Does this phenomenon persist when the input shift does not introduce unseen distributions?
% We design two experiments:

% \paragraph{Experiment 1: Gradual Shift with New Class Introduced.} Each experiment $E_i$ trains on distribution $\mathcal{D}_0$ in task 1 and $\mathcal{D}_i$ in task 2, where each $\mathcal{D}_i$ is a uniform mixture over 10 classes $\{i, i+1, \ldots, i+9\}$. The similarity between $\mathcal{D}_0$ and $\mathcal{D}_i$ is defined as:
% \[
% \text{Similarity}(\mathcal{D}_0, \mathcal{D}_i) = \frac{|\mathcal{D}_0 \cap \mathcal{D}_i|}{|\mathcal{D}_0 \cup \mathcal{D}_i|}.
% \]
% \giulia{What is the range of this quantity? }

% \paragraph{Experiment 2: Gradual Shift within Fixed Class Support.} Each experiment $\tilde{E}_\alpha$ trains on $\tilde{\mathcal{D}}_{0.1}$ in task 1 and $\tilde{\mathcal{D}}_\alpha$ in task 2, where $\tilde{\mathcal{D}}_\alpha = (1 - \alpha)\tilde{\mathcal{D}}_0 + \alpha\tilde{\mathcal{D}}_1$ for two disjoint distributions $\tilde{\mathcal{D}}_0$, $\tilde{\mathcal{D}}_1$.

\begin{figure}[h!]
\vspace{-2mm}
    \centering
    % Row 1
    
        %\includegraphics[width=0.3\textwidth]{images/similarity/NTK_Velocity_w64.p}
        \includegraphics[width=0.33\textwidth]{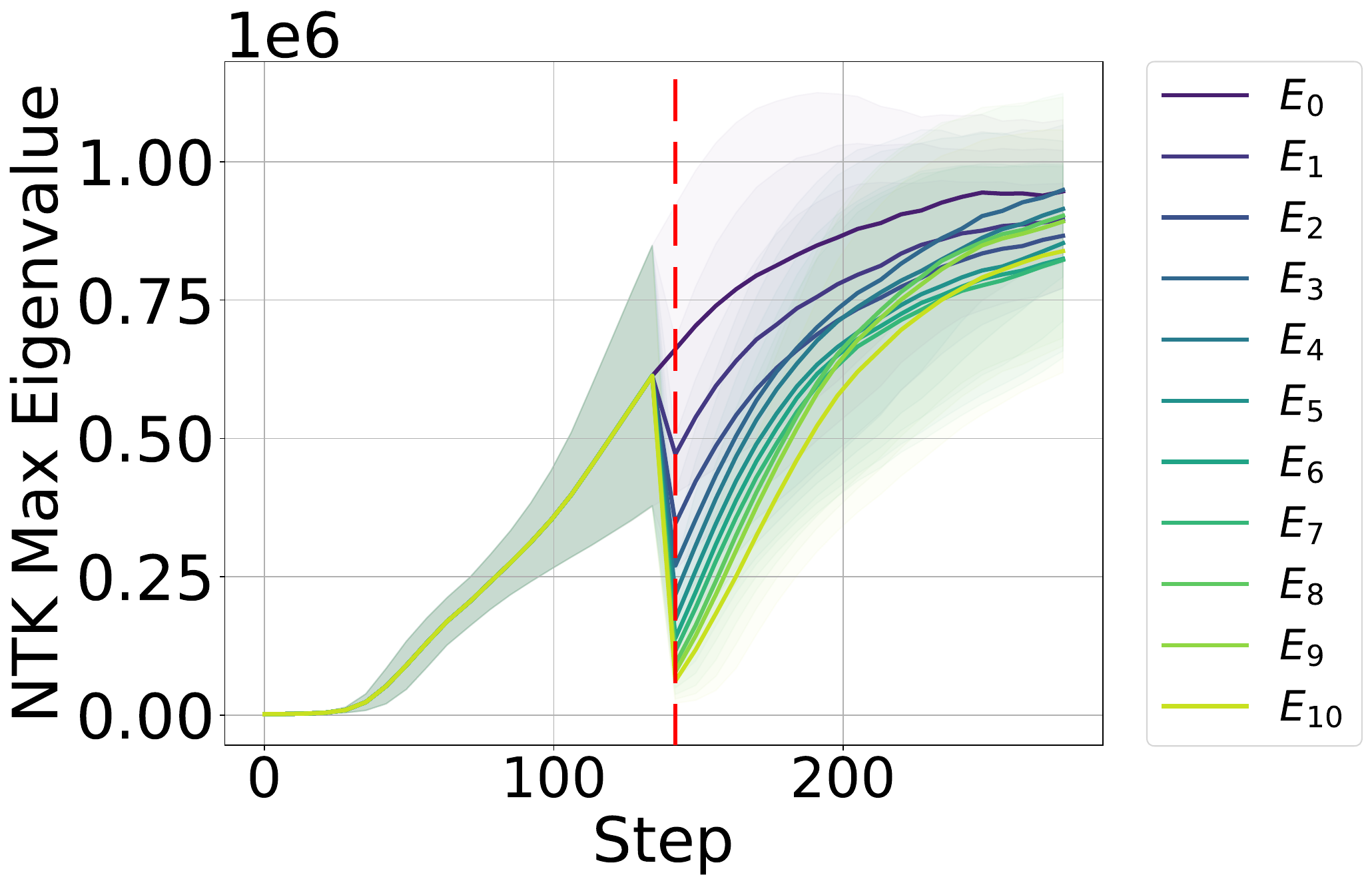}
        \includegraphics[width=0.3\textwidth]{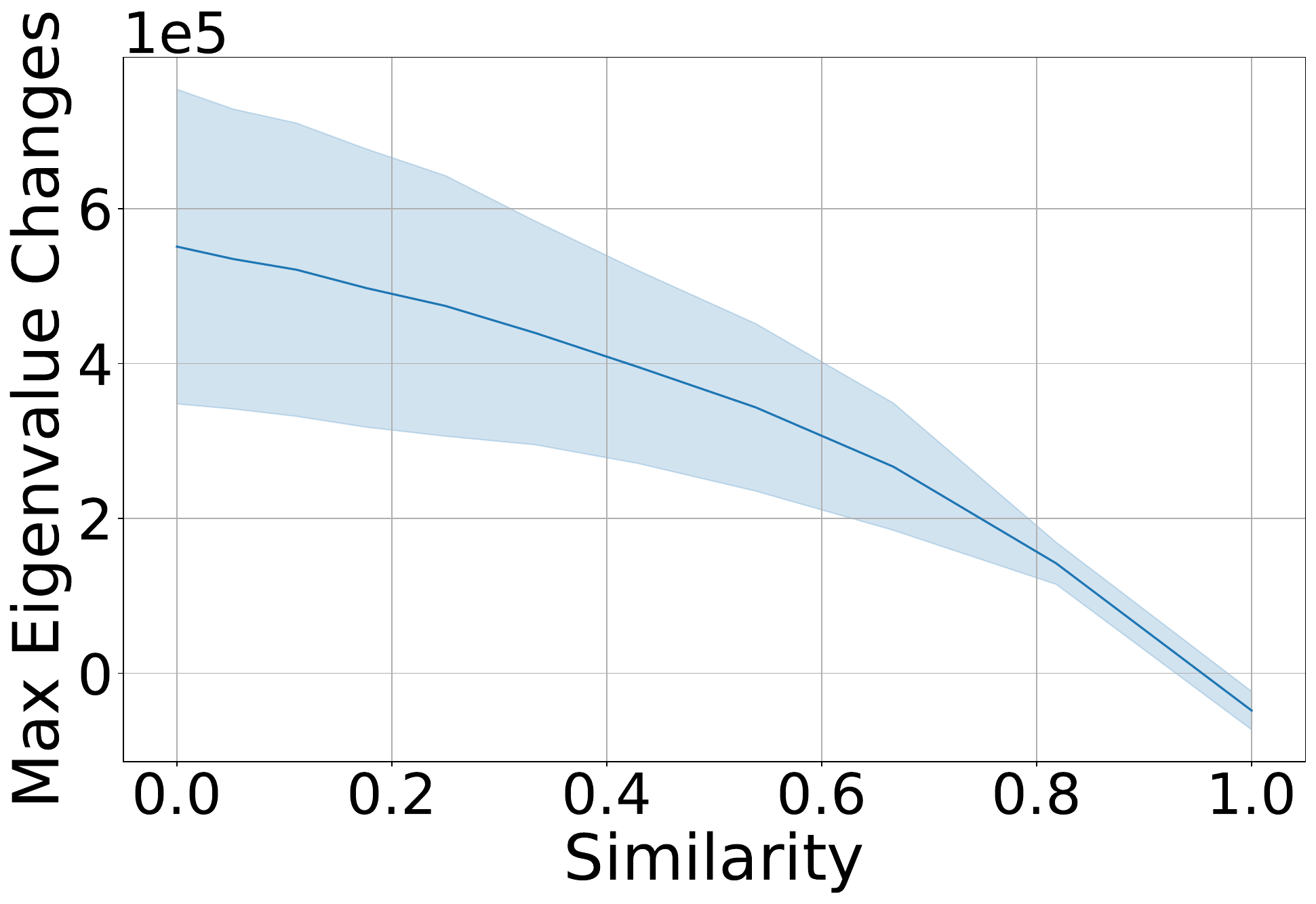}
      \includegraphics[width=0.32\textwidth]{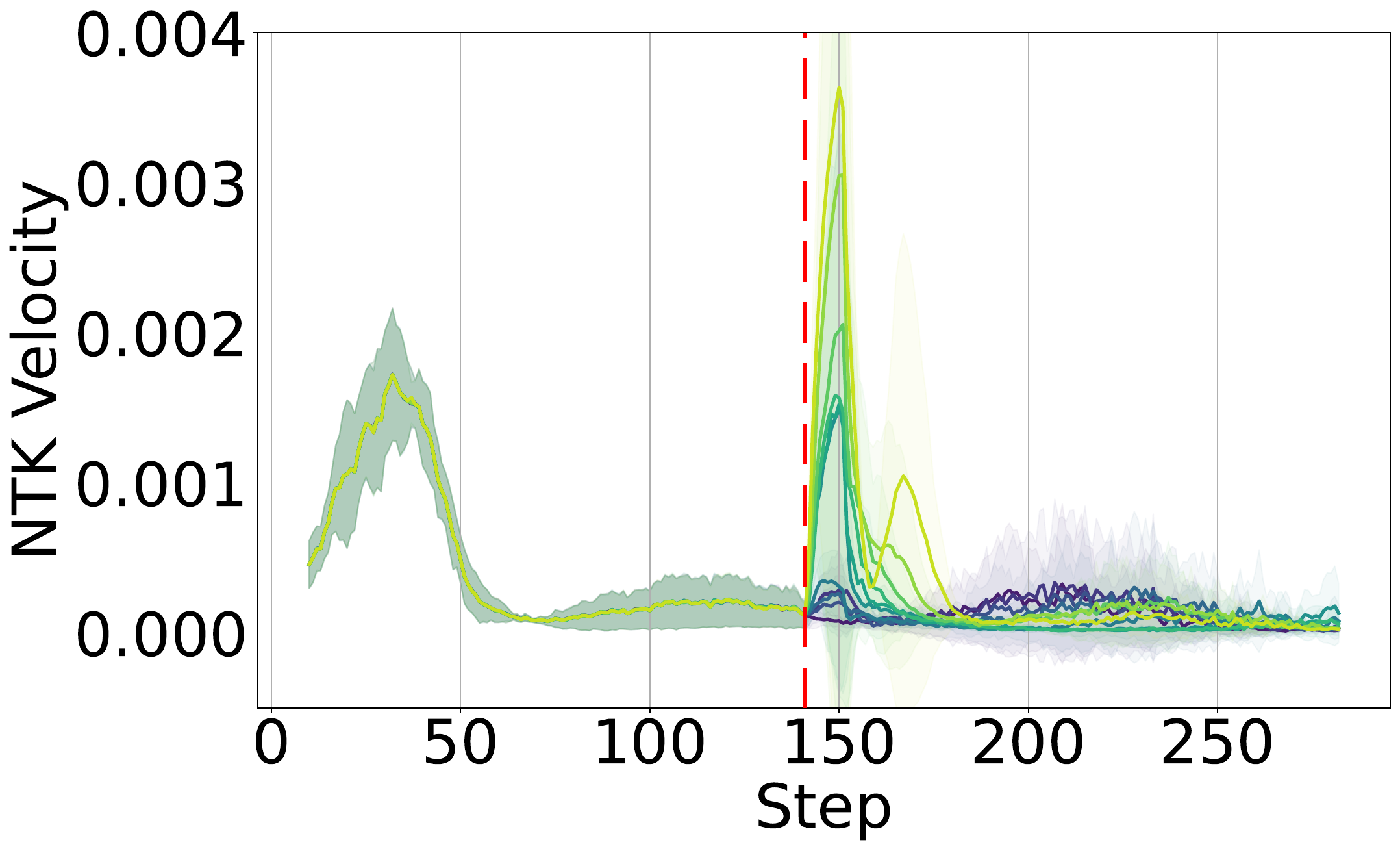}
    % Row 2
        %\includegraphics[width=0.254\textwidth]{images/similarity/NTK_Distance_from_Init_w64.pdf}
    %\subfigure[Zoomed-in]{ {\includegraphics[width=0.2\textwidth]{images/similarity/NTK_Distance_from_Init_w64_zoom.pdf}}}

    \caption{NTK Dynamics of the first task with concept-based distribution shift.} %\giulia{Same comments as above. + Can we have Steps instead of tasks in the x axis?}}
    \label{fig:task_similarity}
\end{figure}

When the new task introduces new concepts (experiments $E_0 \dots, E_{10}$), there is a direct relationship between the number of new classes and the amount of change in the NTK, as confirmed in the measurement of NTK norm, velocity and kernel distance (Figure \ref{fig:task_similarity}, Figure \ref{fig:ntk_distance_exp1}). We observe the characteristic check-mark shape in all but the $E_0$ case, where no distributional change occurs (Figure \ref{fig:task_similarity}-1). The drop in NTK norm becomes progressively smaller as class overlap increases, revealing a clear monotonic relationship between task similarity and the magnitude of NTK disruption.

Again, the NTK norm recovers gradually after the drop, consistent across all levels of similarity (Figure \ref{fig:task_similarity}-1). A similar trend is observed in the kernel distance (Figure \ref{fig:ntk_distance_exp1}), where larger distribution shifts cause more pronounced deviations from the previous NTK state. The trend is neatly ordered by task similarity, suggesting the existence of an underlying law governing the NTK spectral evolution, parametrized by the task similarity. Further, the kernel velocity (Figure \ref{fig:task_similarity}-3) confirms that most of the feature learning occurs immediately following the task switch, after which the network appears to settle back into a more stable regime within a few epochs.

We also note a diminishing return effect: the introduction of the first few new classes causes disproportionately large changes in NTK, while later additions have more incremental impact—suggesting a sublinear relationship between the number of new concepts and NTK disruption.
\vspace{1mm}

\begin{figure}[h]
    \centering
    % Row 2
        \includegraphics[width=0.33\textwidth]{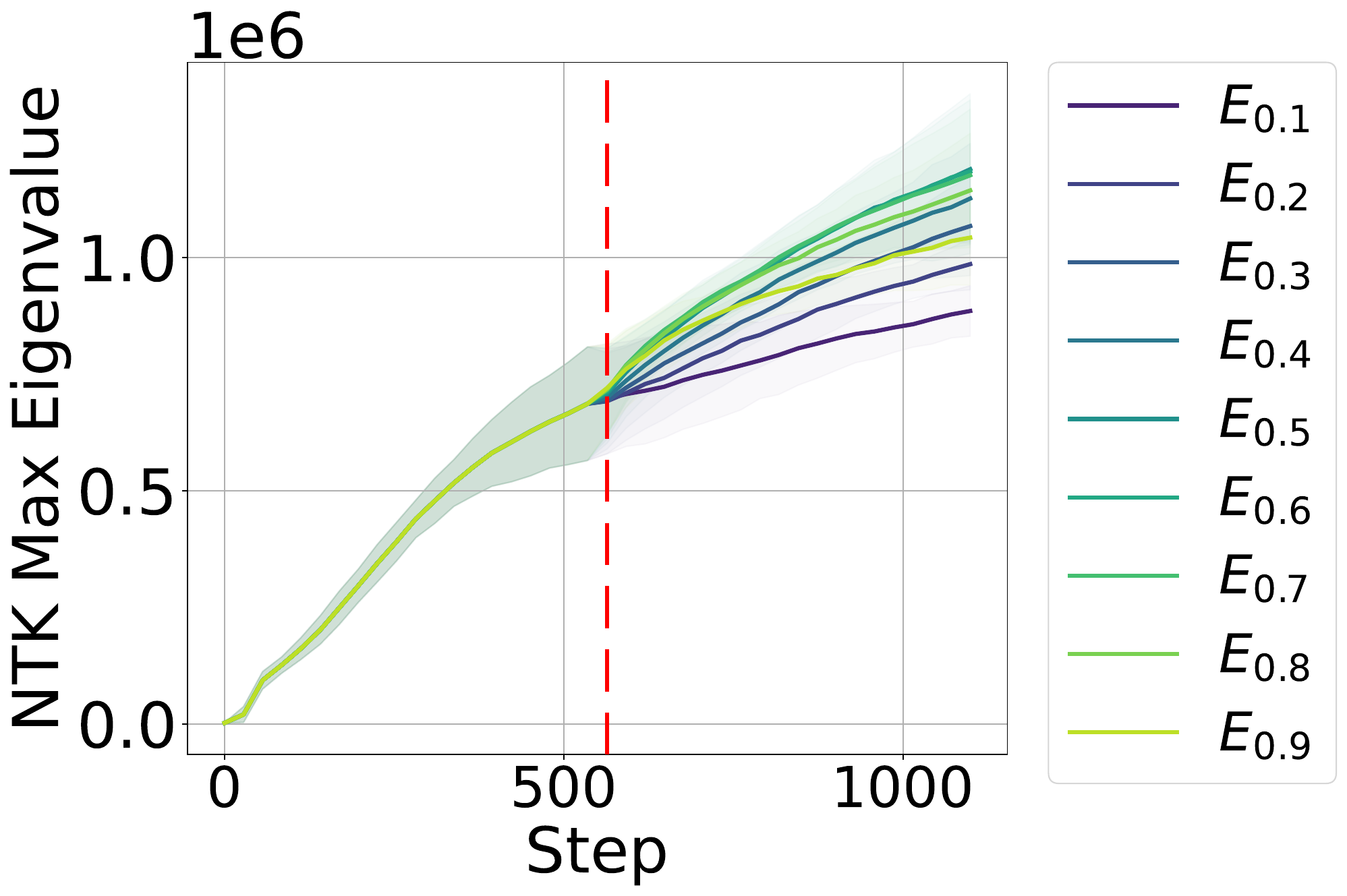}
        \includegraphics[width=0.3\textwidth]{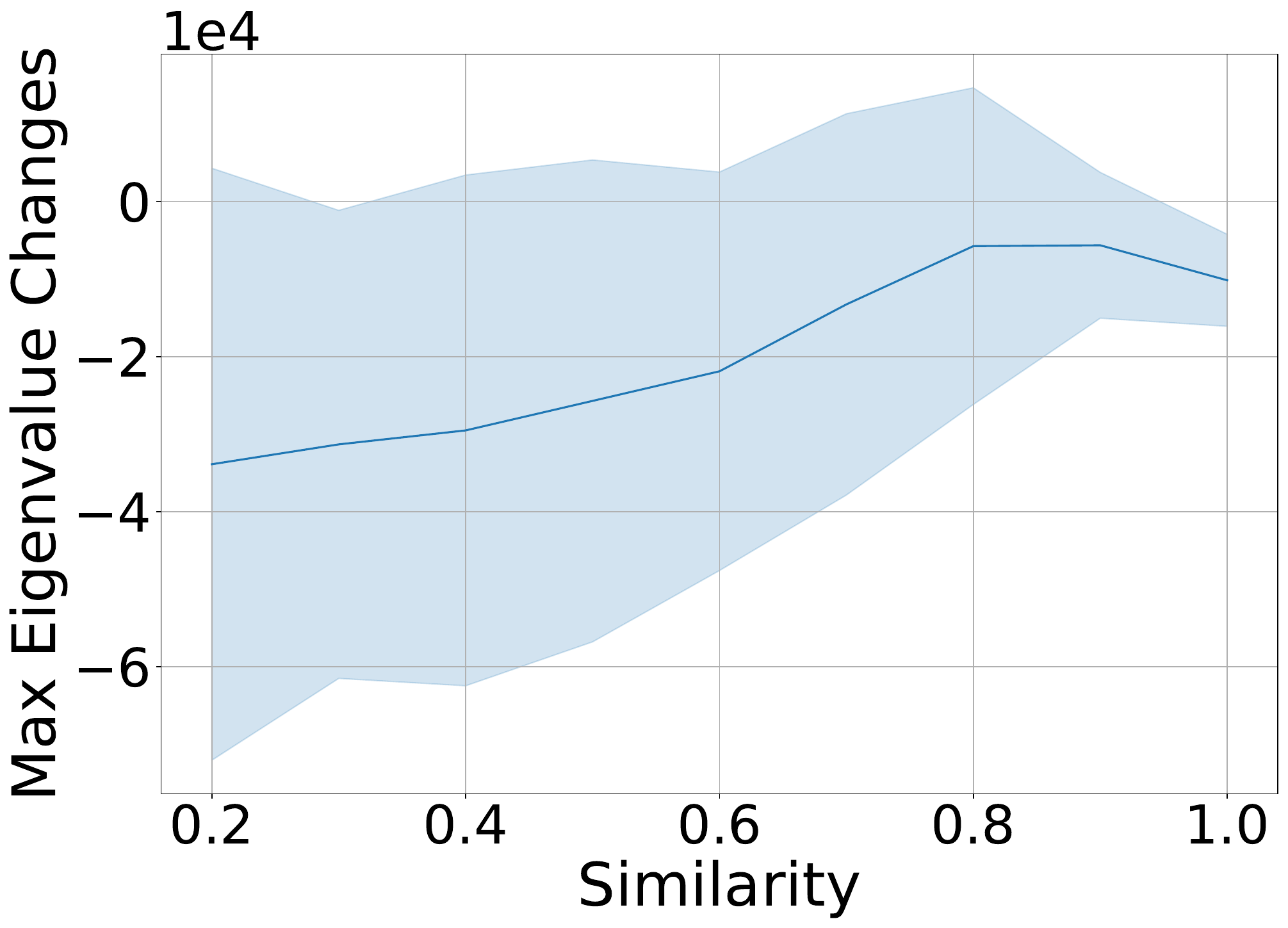}
        \includegraphics[width=0.32\textwidth]{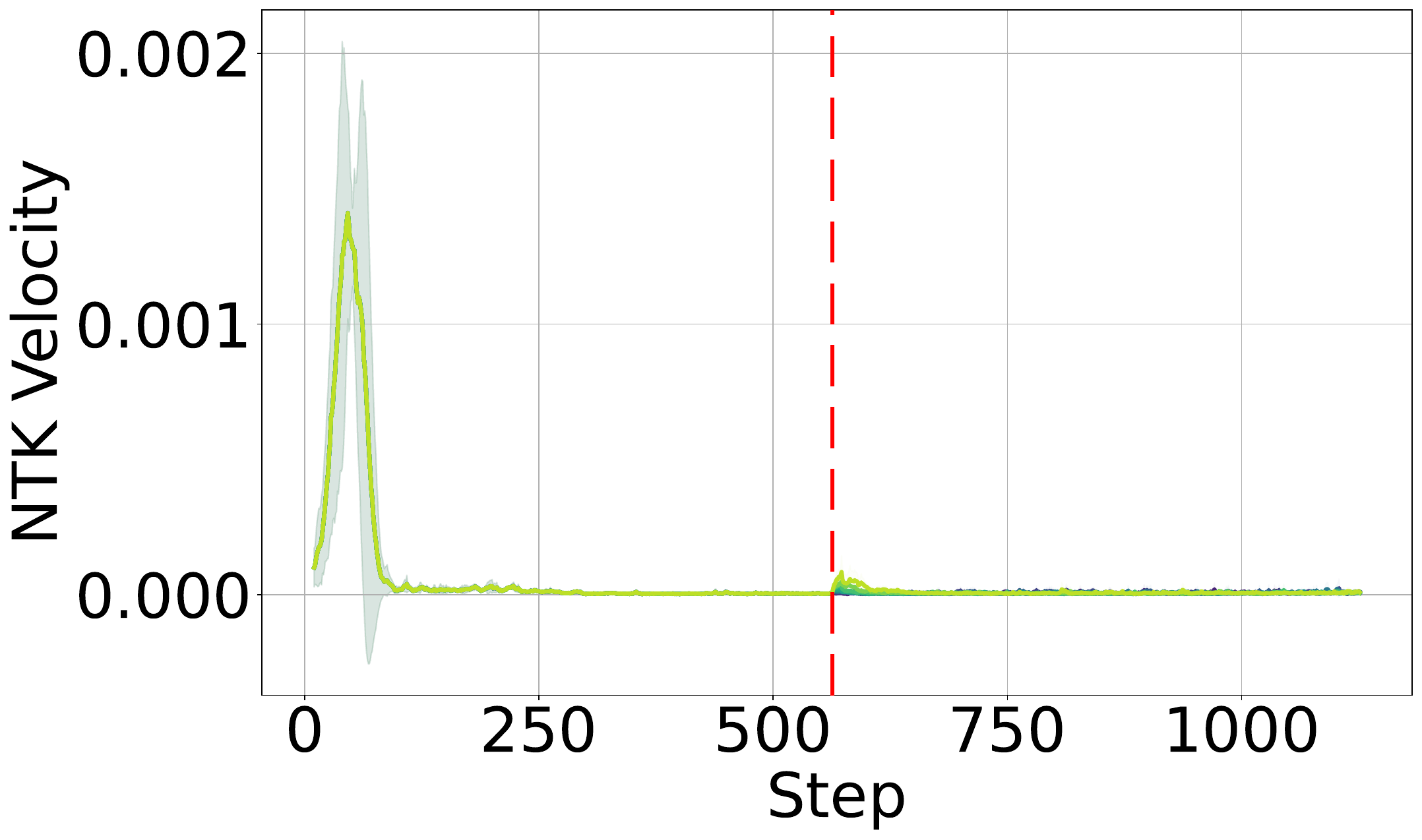}

    \caption{NTK Dynamics of the first task with frequency-based distribution shift.}
    \label{fig:task_similarity_alpha}
\end{figure}

The picture is very different if the new task does not introduce new classes, as in the experiments $E_\alpha$ with  $\alpha \in [0,1]$. 
Figure \ref{fig:task_similarity_alpha} shows that NTK changes in this case are significantly smaller than in Experiment 1. The NTK norm (Figure \ref{fig:task_similarity_alpha}-1) evolves smoothly without any sharp discontinuity at the task switch. Likewise, the kernel velocity (Figure \ref{fig:task_similarity_alpha}-3) remains low, indicating that feature reactivation does not occur in response to proportion shifts alone. Although some monotonic trends are still visible in the NTK eigenvalues (Figure \ref{fig:task_similarity_alpha}-2), their scale is minor.

Together, these results suggest that NTK dynamics encode rich information about task transitions, inviting a deeper theoretical understanding of reactivation phenomena in continual learning.

%\giulia{I think the results would be stronger if we had any of the following: 
%\begin{itemize}
%    \item Multiple architectures 
%    \item Measures of forgetting and learning accuracy (to align with the literature) together with summary statistics of the CKA or kernel velocity during learning + similarity? (in a table?)
%\end{itemize}
%}

\section{Discussion and Conclusion}

We highlight three key observations from our study:

\paragraph{Continual learning disrupts kernel stationarity.} Even in the lazy regime, task switches consistently trigger transient reactivation of feature learning. NTK metrics—including norm, velocity, and alignment—show sharp deviations at task transitions, indicating a temporary departure from static representations. This phenomenon is robust across widths and learning rates, challenging the assumption that wide networks behave as fixed-kernel learners in non-stationary settings.

\paragraph{Task similarity modulates reactivation.} The extent of reactivation depends strongly on the semantic relationship between tasks. When new tasks introduce disjoint class content, NTKs shift significantly. In contrast, changes in class proportions without new concepts lead to minimal NTK disruption. This suggests that semantic overlap could be used as a signal to anticipate representation change.

\paragraph{NTK theory is limited in non-stationary settings.} Classical NTK analyses rely on stationarity assumptions that fail in continual learning. Our results show that NTKs evolve in response to new data distributions, even at large widths. This motivates a theoretical shift: modeling NTK dynamics under distribution shift is essential for understanding continual adaptation and forgetting.

\medskip

In conclusion, our work reveals a structured departure from lazy dynamics at task boundaries—a phenomenon we term \emph{reactivation}. Future theory should account for these transitions, and practical algorithms might even benefit from explicitly managing them. We hope our findings motivate further research into the dynamics of neural networks in non-stationary environments.

% This work investigates the evolution of the empirical Neural Tangent Kernel (eNTK) in continual learning. Our results reveal significant changes in NTK dynamics as new tasks are introduced, suggesting a \emph{re-activation} from lazy training toward feature learning, triggered by the non-stationary nature of the task sequence. This phenomenon challenges the premises of existing theoretical studies and provide a new perspective to the study of catastrophic forgetting of past data.

% We further observe that tasks introducing previously unseen distributions induce more significant changes in the NTK, whereas tasks that shift within already-learned distributions tend to preserve the kernel structure. This highlights the sensitivity of learning dynamics to distributional novelty and emphasizes the importance of task similarity in continual learning.

% These findings call for a deeper theoretical understanding of learning dynamics under distribution shift, and for the development of a theoretical framework that extends beyond the stationary assumptions of classical NTK theory to accommodate sequential task settings. Moreover, how changes in NTK dynamics relate to the well-known challenges in continual learning, such as catastrophic forgetting and loss of plasticity, remains an open question. We hope that our findings lay the foundation for future research in this direction.

\newpage
\bibliography{sample}
\newpage 
\clearpage
\appendix

%\section{More}
\section{The NTK Framework and NTK Spectrum}
\label{sec:appendix_NTK}
\label{sec:theory}
%\giulia{Not sure if this structure is final, but this section is way too long, it takes up too much space. Can we reduce it to under half a page? The rest goes in the appendix.}
%\footnote{For simplicity in the exposition, we consider the target to be one-dimensional. In the Appendix we cover the more general multi-dimensional output case.}
Consider a dataset $\{\mathbf{x}_i\}_{i=1}^n$ with real targets $\{y_i\}_{i=1}^n$ and loss function $\ell$. Let $f_t:=f(\theta_t)$ be the neural network at time $t$, with parameters $\theta$. By gradient descent, in continuous time, the parameters evolve as:
\begin{equation}
    \partial_t\theta(t) = -\nabla_{\theta}\ell = - \frac{1}{n} \sum_{i=1}^n \dfrac{\partial\ell}{\partial f_t(\mathbf{x_i)}}\cdot\dfrac{\partial f_t(\mathbf{x_i)}}{\partial\theta_t} 
\end{equation}
Hence, by chain rule the network function space is determined by the \emph{Neural Tangent Kernel} $\Theta_t(\mathbf{x}_i,\mathbf{x}_j) = \nabla_\theta f_t(\mathbf{x}_i)^\top \nabla_\theta f_t(\mathbf{x}_j)$:
\begin{equation}
\label{eq:f_update}
    \partial_t{f_t(\mathbf{x})} = \dfrac{\partial f_t}{\partial \theta_t}\cdot \partial_t\theta_t
    = -\frac{1}{n}\sum_{i=1}^n\Theta_t(\mathbf{x},\mathbf{x}_i)\dfrac{\partial\ell}{\partial f_t(\mathbf{x_i)}}
\end{equation}
\vspace{-8mm}
% Here $\Theta_t$ is the empirical Neural Tangent Kernel, its (i,j)-th entry is defined as:
% \begin{equation*}
%  \Theta_t(\mathbf{x}_i,\mathbf{x}_j) = \nabla_\theta f_t(\mathbf{x}_i)^\top \nabla_\theta f_t(\mathbf{x}_j)
% \end{equation*} 

\subsection{Eigenvalues of NTK}
\label{sec:ntk_eigenvalues}
Define the function output on dataset as $\mathbf{f}=[f(\mathbf{x_1}),f(\mathbf{x_2}),...,f(\mathbf{x_n})]^\top$ and define the residuals as $\mathbf{e} = \mathbf{f} - \mathbf{y}$. For squared loss,  the evolution of $\mathbf{f}$ and $\mathbf{e}$ at each time step is:
\begin{equation}
    \label{eq:ve_update}
    \mathbf{e}_{t+1} = (\mathbf{I} - \eta \Theta_t)\mathbf{e}_t,
\end{equation} 
We diagonalize the NTK as $\Theta = Q \Lambda Q^\top$ and project $\mathbf{e}$ onto the eigenbasis $Q$:
\begin{equation}
\label{eq:eigen_update}
   \tilde{\mathbf{e}}_{t+1} = (\mathbf{I} - \eta \Lambda)\tilde{\mathbf{e}}_{t}, \quad
   \tilde{{e}}_{t+1}^i =  (1 - \eta \lambda_i) \tilde{{e}}_{t}^i  \;\;
   \text{ for each eigenmode }i
\end{equation}
Thus, the error in each eigenmode decays at a rate determined by the corresponding eigenvalue \( \lambda_i \), indicating that the NTK governs the learning speed of each mode based on its eigenvalue. An higher NTK norm thus corresponds to faster convergence in some eigenmodes.
\vspace{-2mm}
\subsection{Centered Kernel Alignment}
\label{appendix:cka_def}
Centered Kernel Alignment (CKA) is a statistical similarity measure used to compare the structure of two kernels, typically in the context of neural network analysis.

The (linear) CKA between $A$ and $B$ is defined as:

\[
\mathrm{CKA}(A, B) = \frac{\langle A, B \rangle_F}{\|A\|_F \|B\|_F}
= \frac{\operatorname{Tr}(A\top B)}{\sqrt{\operatorname{Tr}(A^\top A)} \sqrt{\operatorname{Tr}(B^\top B)}}
\]
Here, $\langle A, B \rangle_F = \operatorname{Tr}(A^\top B)$ denotes the Frobenius inner product, and $\|A\|_F = \sqrt{\operatorname{Tr}(A^\top A)}$ is the Frobenius norm. A high CKA value (close to 1) indicates that the two kernels exhibit similar internal structure.

%\giulia{content missing here?}
\vspace{-2mm}

\subsection{Computation of NTK}
%\giulia{Let's discuss here details of the kernel computation, and the stability of kernel norm with batch size. }
%\lyz{NTK computation details? }
Neural Tangent Kernel (NTK) matrices are computed based on \citet{yang2020tensorprogramsiineural}. The gradients of the trained models are cleared and a single sample is fed through the network while recording the gradient. This process is repeated for multiple samples and the NTK matrix is computed by taking the inner product between all the gradients.

\begin{figure}[h]
    \centering
    % Row 1
    \subfigure[Max eigenvalues ($n = 10$)]{
        \includegraphics[width=0.31\textwidth]{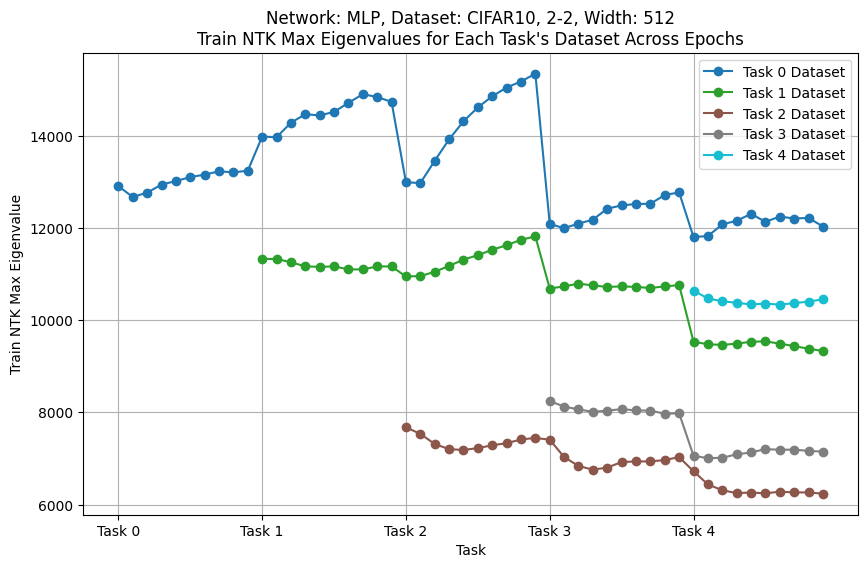}
    }
    \subfigure[Max eigenvalues ($n = 20$)]{
        \includegraphics[width=0.31\textwidth]{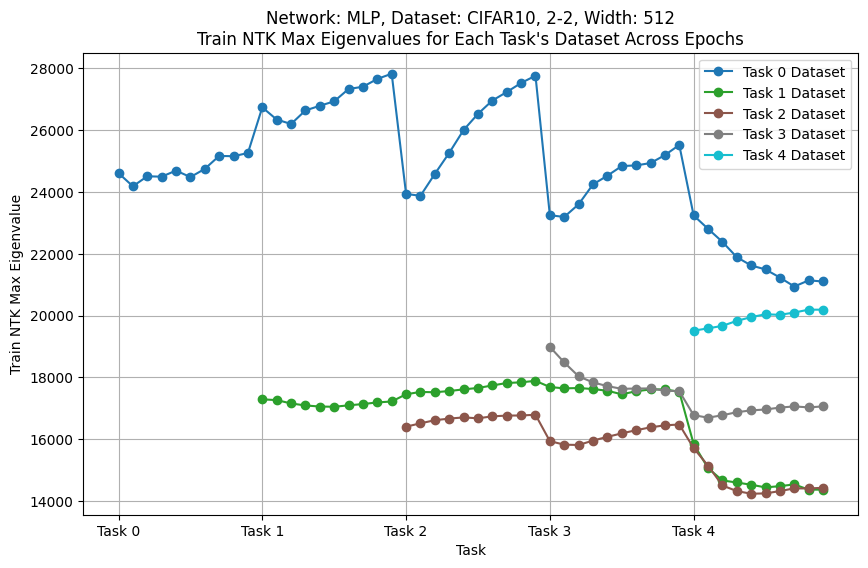}
    }
    \subfigure[Max eigenvalues ($n = 100$)]{
        \includegraphics[width=0.31\textwidth]{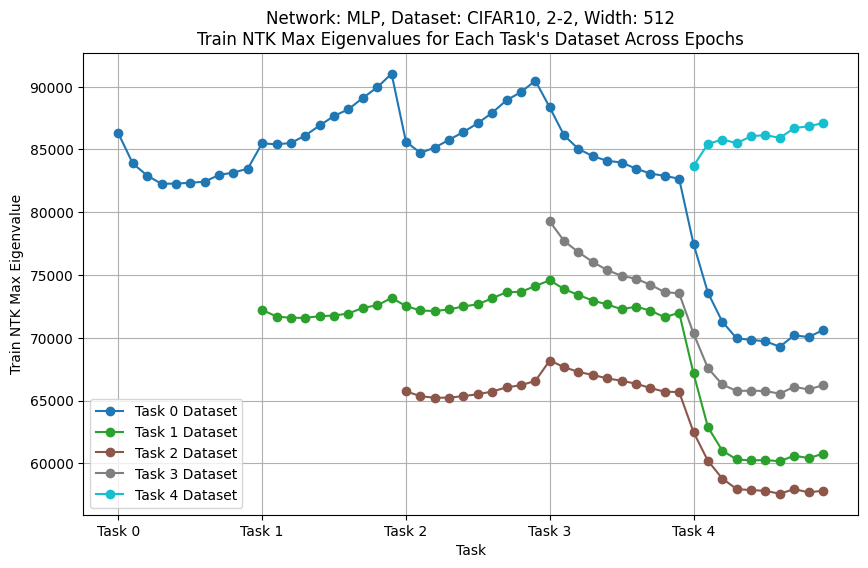}
    }
 % Row 2
    \subfigure[NTK Norm ($n = 10$)]{
        \includegraphics[width=0.31\textwidth]{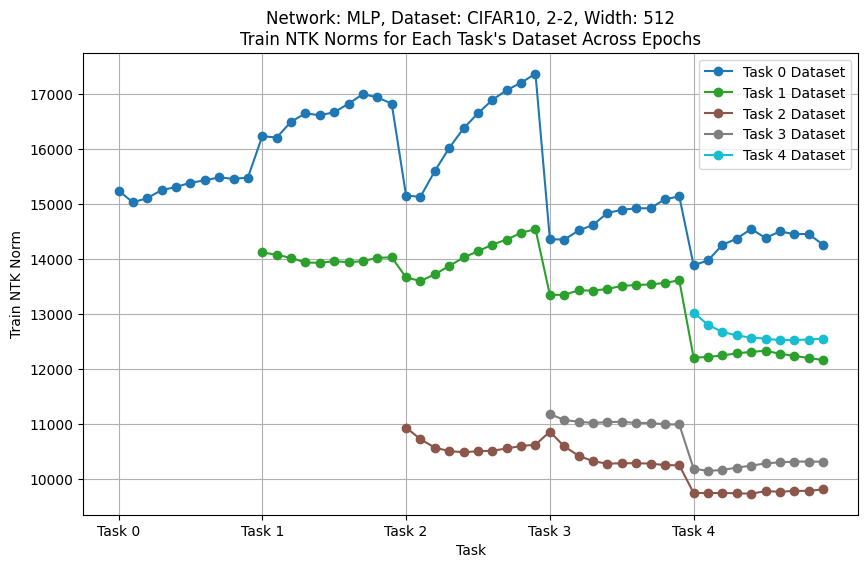}
    }
    \subfigure[NTK Norm ($n = 20$)]{
        \includegraphics[width=0.31\textwidth]{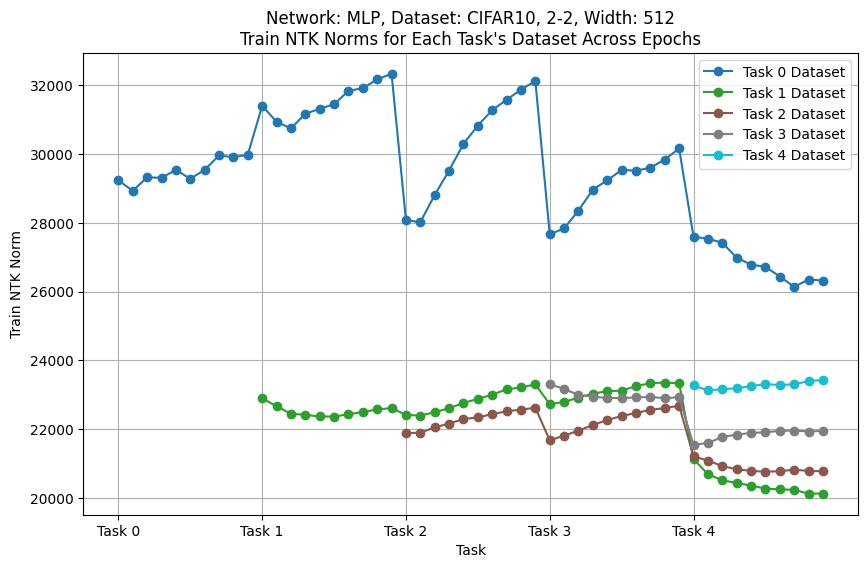}
    }
    \subfigure[NTK Norm ($n = 100$)]{
        \includegraphics[width=0.31\textwidth]{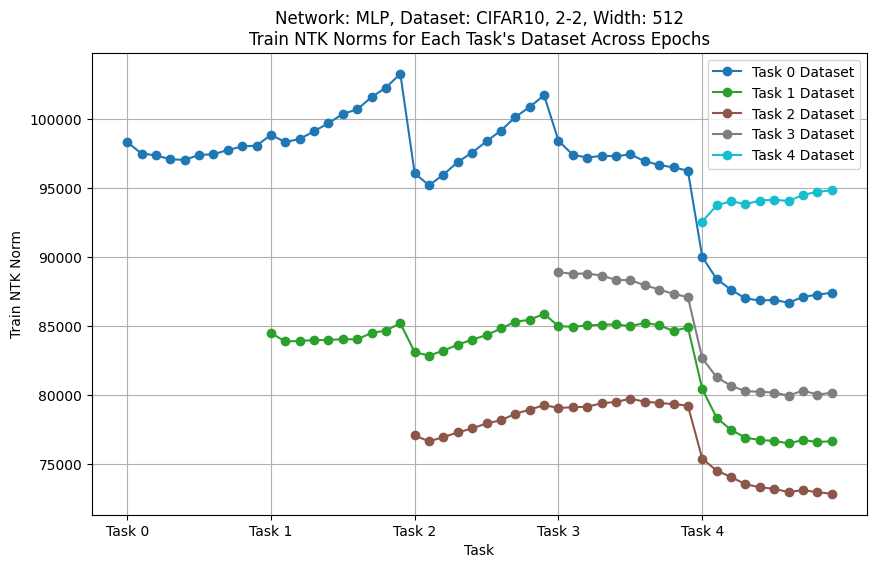}
    }

    \caption{NTK dynamics computed with different sample sizes ($n$).}

    \label{fig:sample_size}
\end{figure}

We empirically evaluated the impact of sample size on the computation of the empirical NTK as a sanity check, alongside ablation studies. Specifically, we tested three different sample sizes ($10$, $20$, and $100$). The results (Fig. \ref{fig:sample_size}) indicate that while the NTK norm increases linearly with sample size, the overall trend of the NTK norm remains consistent across all sample sizes.

For all our main experiments, we use a batch size of 32 random samples to compute the empirical NTK. This choice is sufficient to reliably capture the trend of the NTK throughout the training process.

\section{Experiment Details}
\label{sec:experiment_details}

\subsection{Network parametrization.}
\label{subsec:net-param}
The NTK parametrization induces lazy learning dynamics as the width of the network increases to infinity \citep{jacot2018neural, yang_feature_2022}. For a network of width $N$, the weights are initialized according to $W_{ij}^{(l)}\sim \mathcal{N}(0,1)$ and rescaled by defining the training parameters to be $\omega_{ij}^l = \frac{1}{\sqrt{N}} W_{ij}^{(l)}$. 
To mimic the NTK parametrization, we adopt the Kaiming Normal initialization and scale the learning rate with respect to width with $1/N$. The learning rate re-scaling has been shown to be essential in order to produce a valid limit as the width increases, and avoid the blow-up in the network dynamics \citep{yang_feature_2022}.
%In NTK scheme, weights are initialized with variance $1/width$, i.e. $W_{ij}^{(l)}\sim \mathcal{N}(0,1/width)$ and $b^{(l)}\sim \mathcal{N}(0,1)$, which makes the network evolve according to a fixed NTK kernel \cite{Jacot_NEURIPS2018_NTK}. 
Kaiming Normal initialization draws weights from a zero-mean normal distribution with $W_{ij}\sim\mathcal{N}(0,2/n_{in})$, where $n_{in}$ is the number of input connections to the neuron, proportional to $N$ in our settings \cite{he2015delving}. Therefore, Kaiming Normal initialization with scaled learning rate $1/N$ is very similar to NTK parametrization with only factor-wise difference. 

Unless stated otherwise, all the experiments are performed under feature learning regime, with Kaiming uniform initialization. 

\subsection{Task Shifts Experiments}
\label{subsec: task_shifts}
To analyze learning dynamics in a continual learning setting, we train a simple Convolutional Neural Network (CNN) consisting of three convolutional layers, three pooling layers, and a fully connected layer with ReLU activation functions for image classification on CIFAR and ImageNet. All experiments use the SGD optimizer and cross-entropy loss. 

The classification problem on CIFAR-10 is split into two tasks, each containing five classes. We explore various settings, including CNN widths (64, 128, 256, 512, 1024, 2048), learning rates (1e-3, 1e-4, 1e-5), and training epochs (10, 20, 40, 80, 160). Here, the width of the CNN refers to the number of channels in the convolutional layers.

\subsection{Task Similarity Experiments}
\label{appendix:exp_similarity}
%\lyz{to be expanded}
We train a simple CNN using the same configuration described in Section~\ref{subsec: task_shifts}. Each experiment consists of two sequential tasks with varying degrees of similarity, where no data from the first task is reused in the second. The model with width 32 and 64 is trained for 20 epochs per task using stochastic gradient descent (SGD) with a constant learning rate of 1e-3.
\paragraph{Experiment 1: Gradual Shift with New Class Introduced.}  
We define a family of input distributions $\mathcal{D}_i = \{i, i+1, \ldots, i+9\}$, where each $\mathcal{D}_i$ is a uniform mixture over 10 consecutive CIFAR-100 classes. In experiment $E_i$, we construct a two-task continual learning scenario: \textbf{Task 1} trains on $\mathcal{D}_0$ and \textbf{Task 2} trains on $\mathcal{D}_i$ for $i = 0, \ldots, 10$. The similarity between $\mathcal{D}_0$ and $\mathcal{D}_i$ is defined as:
\[
\text{Similarity}(\mathcal{D}_0, \mathcal{D}_i) = \frac{|\mathcal{D}_0 \cap \mathcal{D}_i|}{|\mathcal{D}_0 \cup \mathcal{D}_i|}.
\]

\paragraph{Experiment 2: Gradual Shift within Fixed Class Support.}  
Define two disjoint sets of classes on CIFAR 10: $\tilde{\mathcal{D}}_0 = \{0,1,2,3,4\}$ and $\tilde{\mathcal{D}}_1 = \{5,6,7,8,9\}$, and construct a family of mixed distributions:
\[
\tilde{\mathcal{D}}_\alpha = (1 - \alpha)\tilde{\mathcal{D}}_0 + \alpha\tilde{\mathcal{D}}_1, \quad \alpha \in \{0.1, 0.2, \ldots, 0.9\}.
\]
Here for each experiment $\tilde{E}_\alpha$, \textbf{Task 1} is fixed to learn from $\tilde{\mathcal{D}}_{0.1}$, and \textbf{Task 2} learns from $\tilde{\mathcal{D}}_\alpha$. 

\begin{figure}[h]
    \centering
   \includegraphics[width=0.4\textwidth]{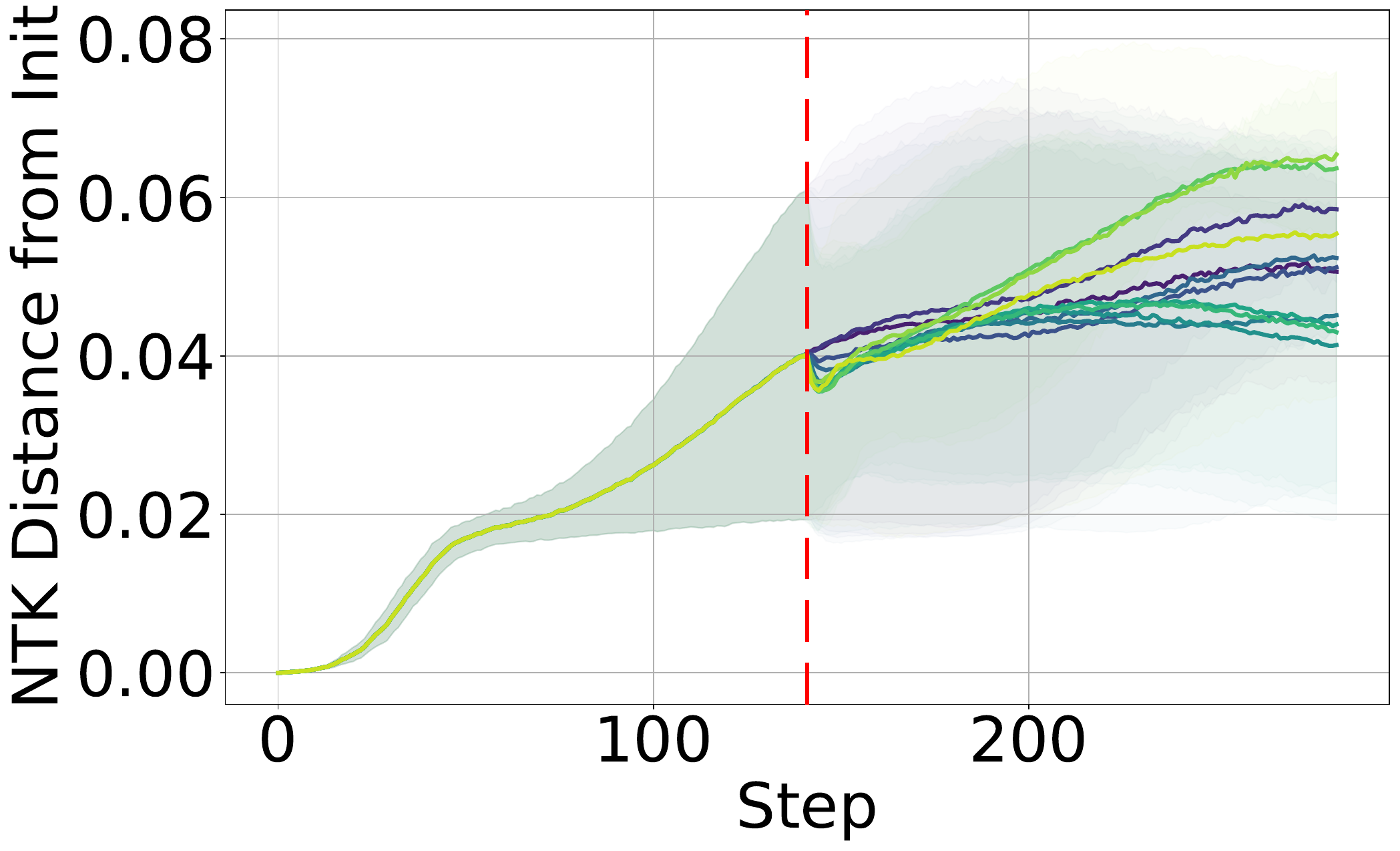}
    \includegraphics[width=0.35\textwidth]{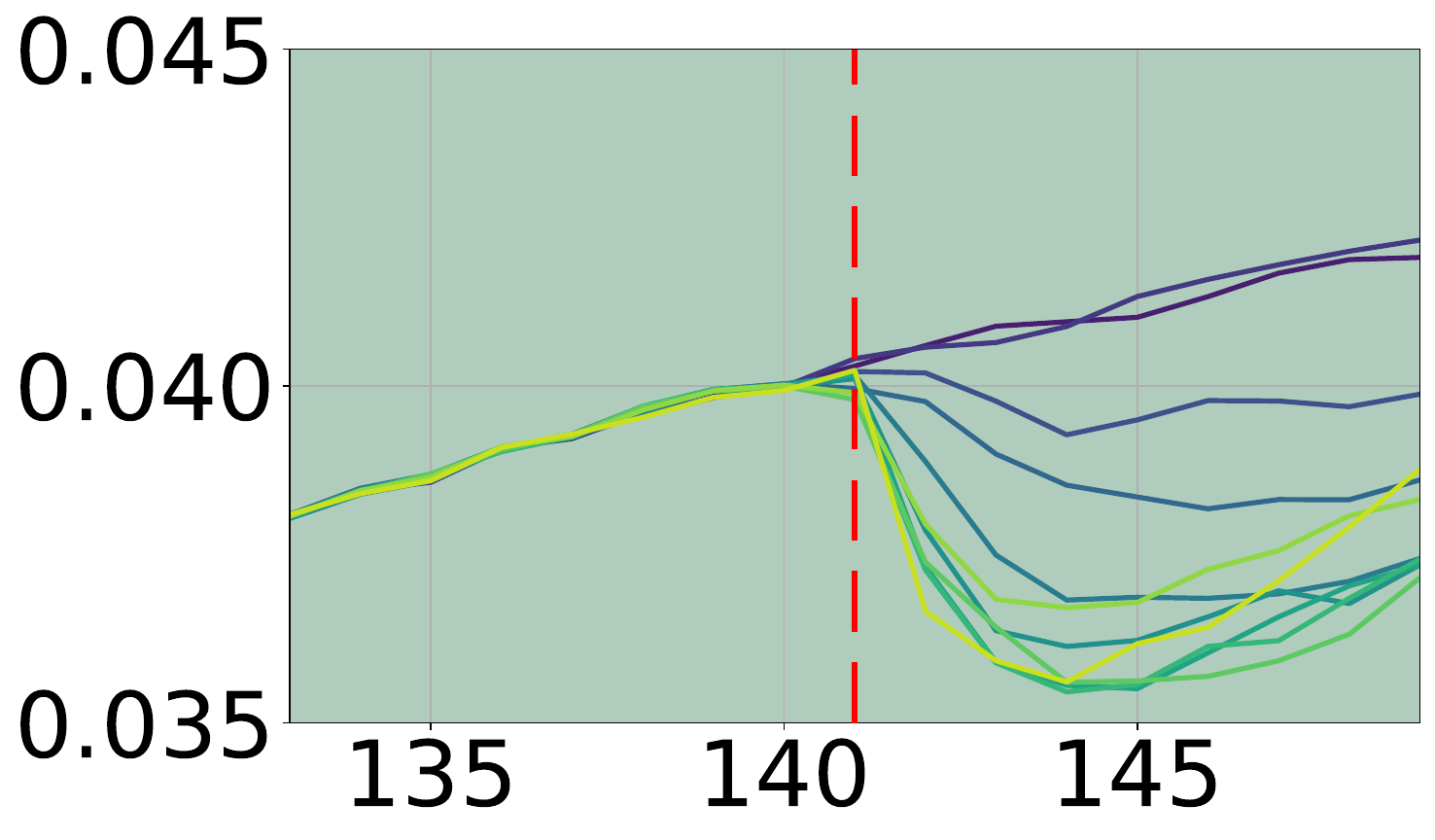}
    \caption{Kernel distance in Experiment 1 and a zoomed-in view around the task switch.}
    \label{fig:ntk_distance_exp1}
\end{figure}

\section{Complementary Results}
\label{sec:complementary-results}

\subsection{Comprehensive Metrics evaluation for CIFAR10 (Two Tasks)}

\begin{figure}[h!]
    \centering
    % 第一排
    \subfigure[Accuracy]{
        \includegraphics[height=0.21\textwidth]{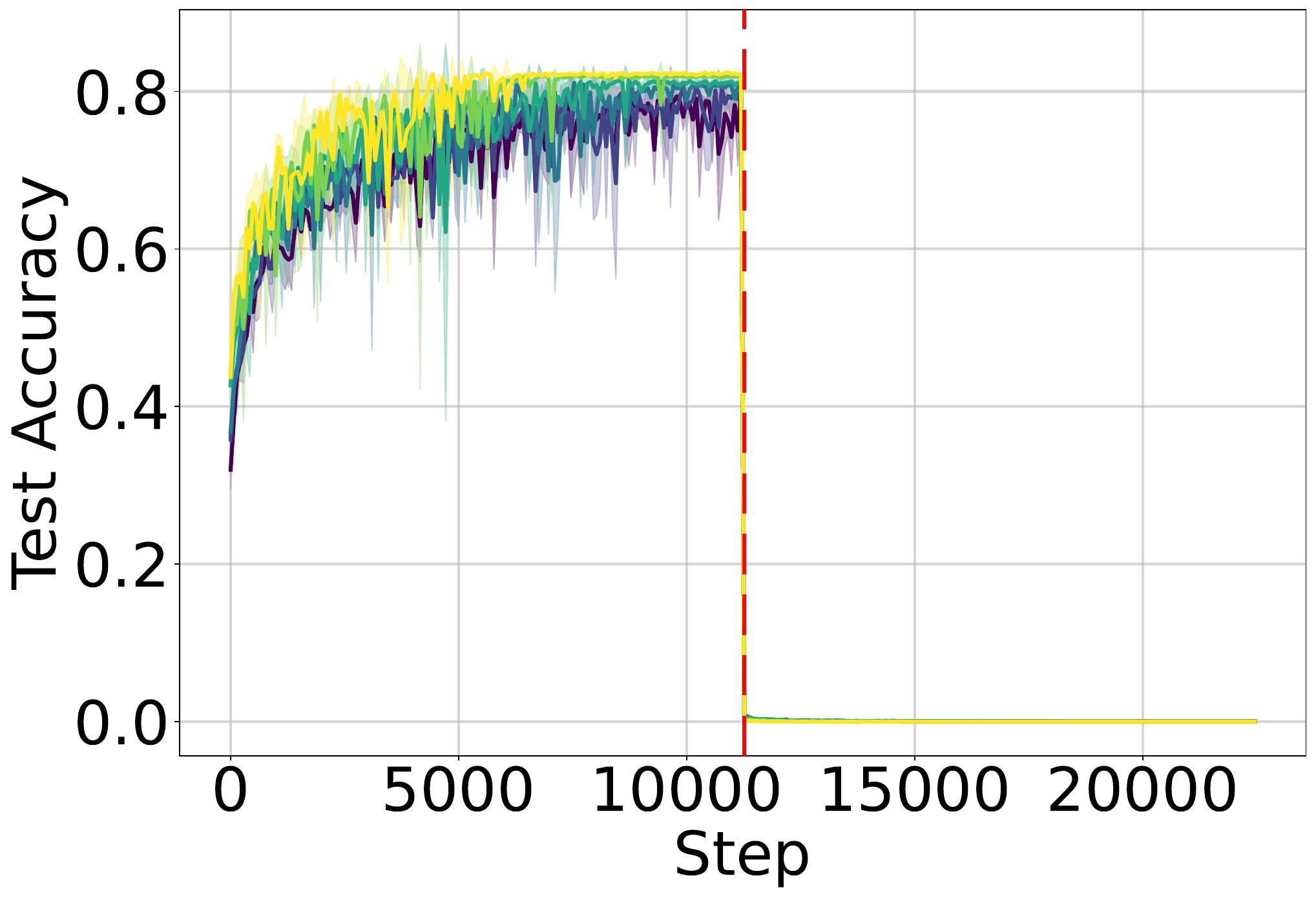}
    }
    \hfill
    \subfigure[Alignment]{
        \includegraphics[height=0.21\textwidth]{images/lr_0.001/width_comparison_alignment_cnn_CIFAR10_inc5-5_e160_b32_lr0.001_sgd_s32.pdf}
    }
    \hfill
    \subfigure[Kernel Distance]{
        \includegraphics[height=0.21\textwidth]{images/lr_0.001/width_comparison_cka_cnn_CIFAR10_inc5-5_e160_b32_lr0.001_sgd_s32.pdf}
    }

    % 第二排
    \subfigure[Max Eigenvalue]{
        \includegraphics[height=0.2\textwidth]{images/lr_0.001/width_comparison_ntk_max_eigenvalues_cnn_CIFAR10_inc5-5_e160_b32_lr0.001_sgd_s32.pdf}
    }
    \hspace{0.01\textwidth}
    \subfigure[Velocity (dt=10)]{
        \includegraphics[height=0.2\textwidth]{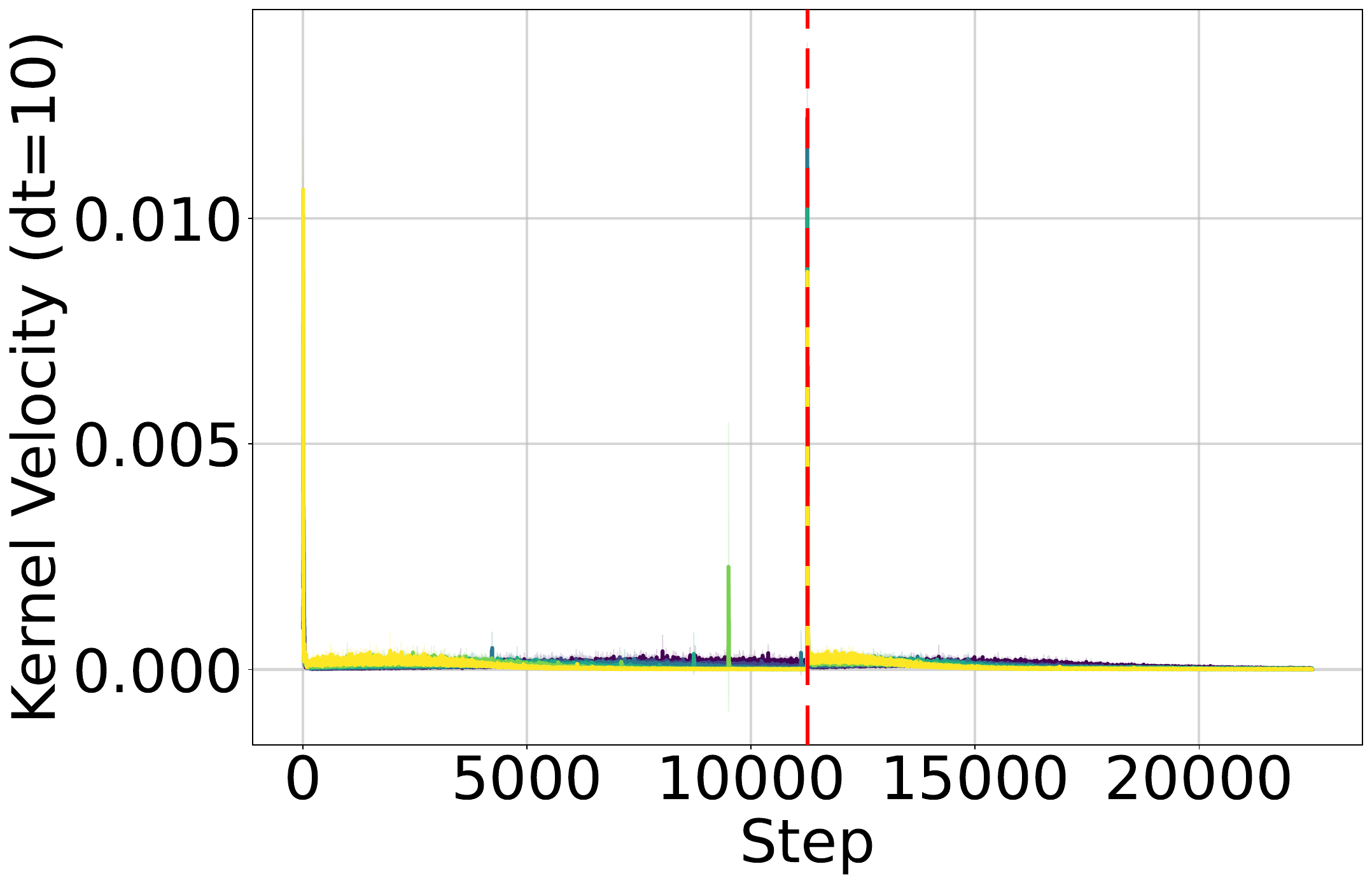}
    }
    \hspace{0.01\textwidth}
    \subfigure[Velocity (dt=10) Zoomed-in]{
        \includegraphics[height=0.2\textwidth]{images/lr_0.001/width_comparison_velocity_dt10_cnn_CIFAR10_inc5-5_e160_b32_lr0.001_sgd_s32_zoom_in.pdf}
    }

    \caption{Comparison of different metrics across network widths for CNN trained on CIFAR10 with learning rate 0.001. The number of epochs per task is set to 160. (a) Test accuracy, (b) Alignment, (c) Kernel distance, (d) Maximum eigenvalue of NTK, (e) Kernel velocity with dt=10, (f) Kernel velocity (zoomed-in).}
    \label{fig:lr_0.001}
\end{figure}

\begin{figure}[h!]
\vspace{-5mm}
    \centering
    % 第一排
    \subfigure[Accuracy]{
        \includegraphics[height=0.21\textwidth]{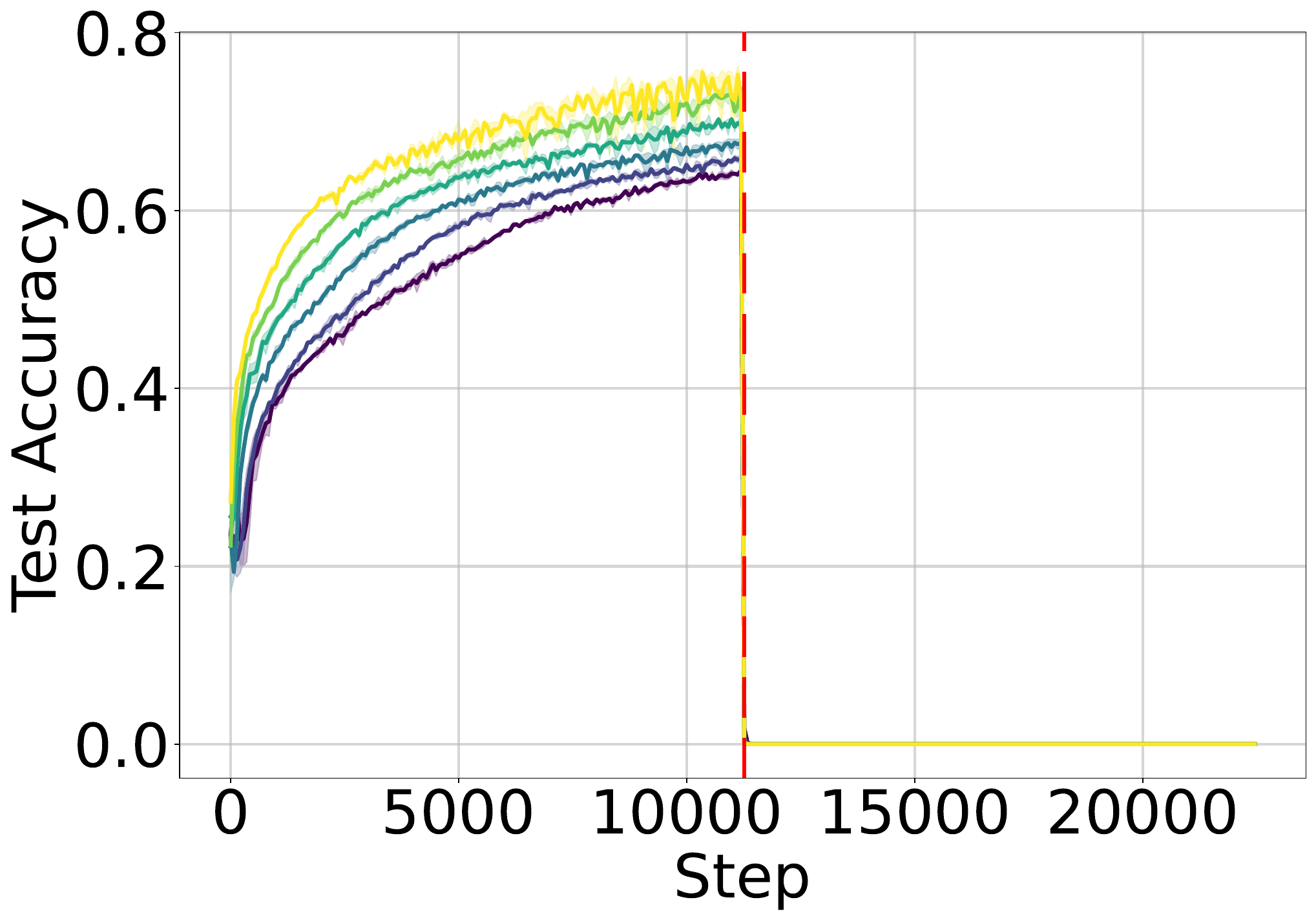}
    }
    \hfill
    \subfigure[Alignment]{
        \includegraphics[height=0.21\textwidth]{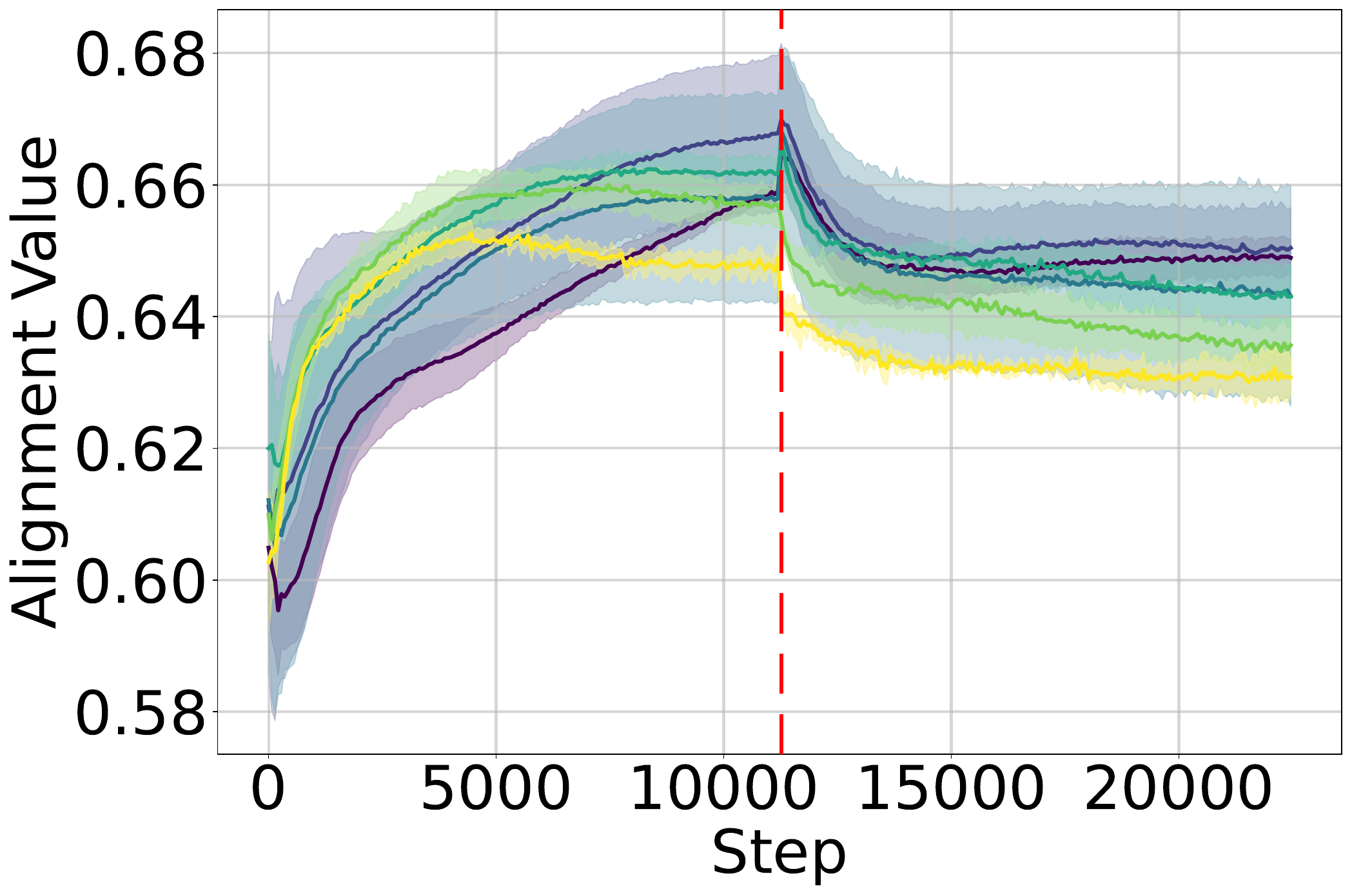}
    }
    \hfill
    \subfigure[Kernel Distance]{
        \includegraphics[height=0.21\textwidth]{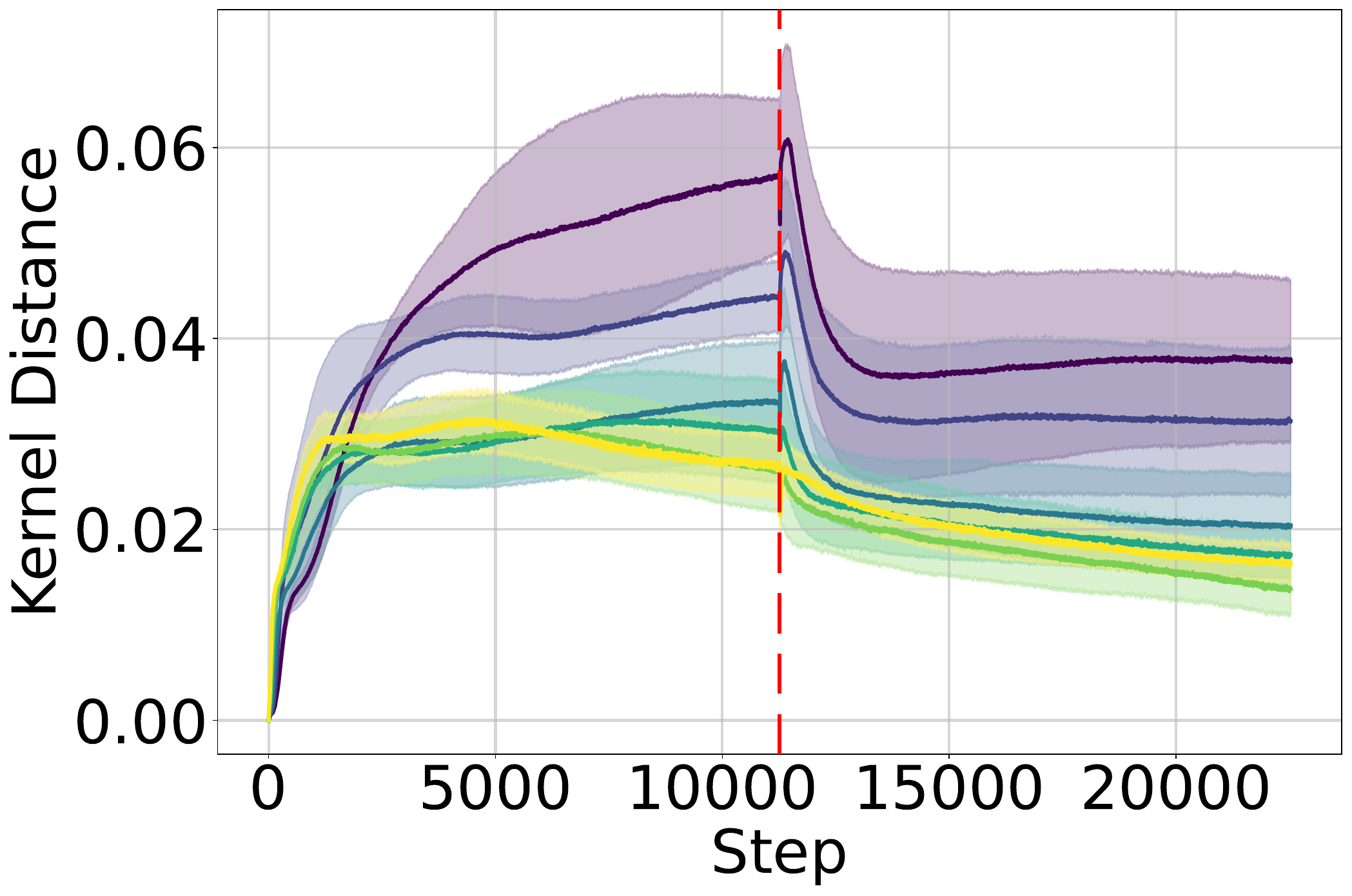}
    }

    % 第二排
    \subfigure[Max Eigenvalue]{
        \includegraphics[height=0.2\textwidth]{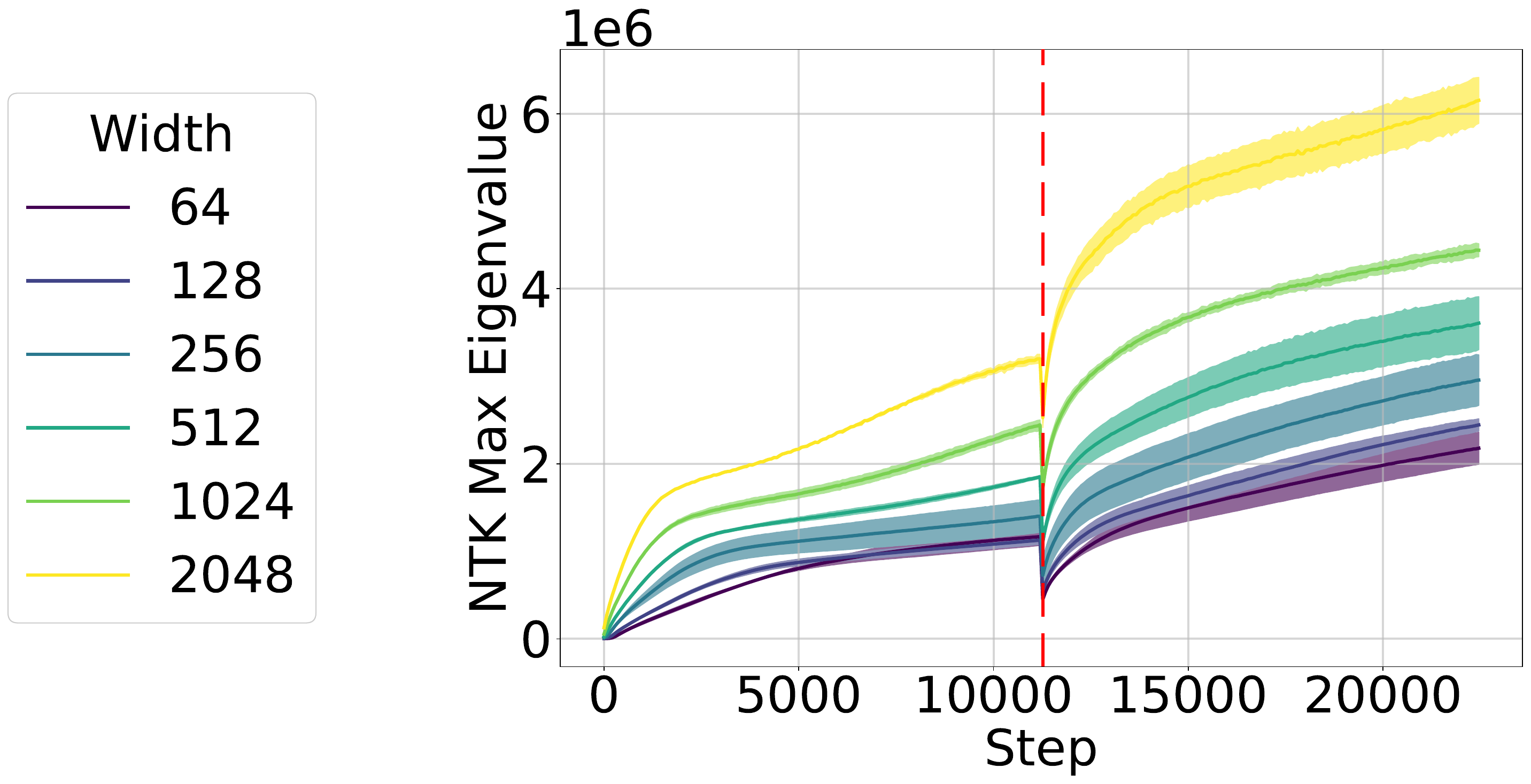}
    }
    \hspace{0.01\textwidth}
    \subfigure[Velocity (dt=10)]{
        \includegraphics[height=0.19\textwidth]{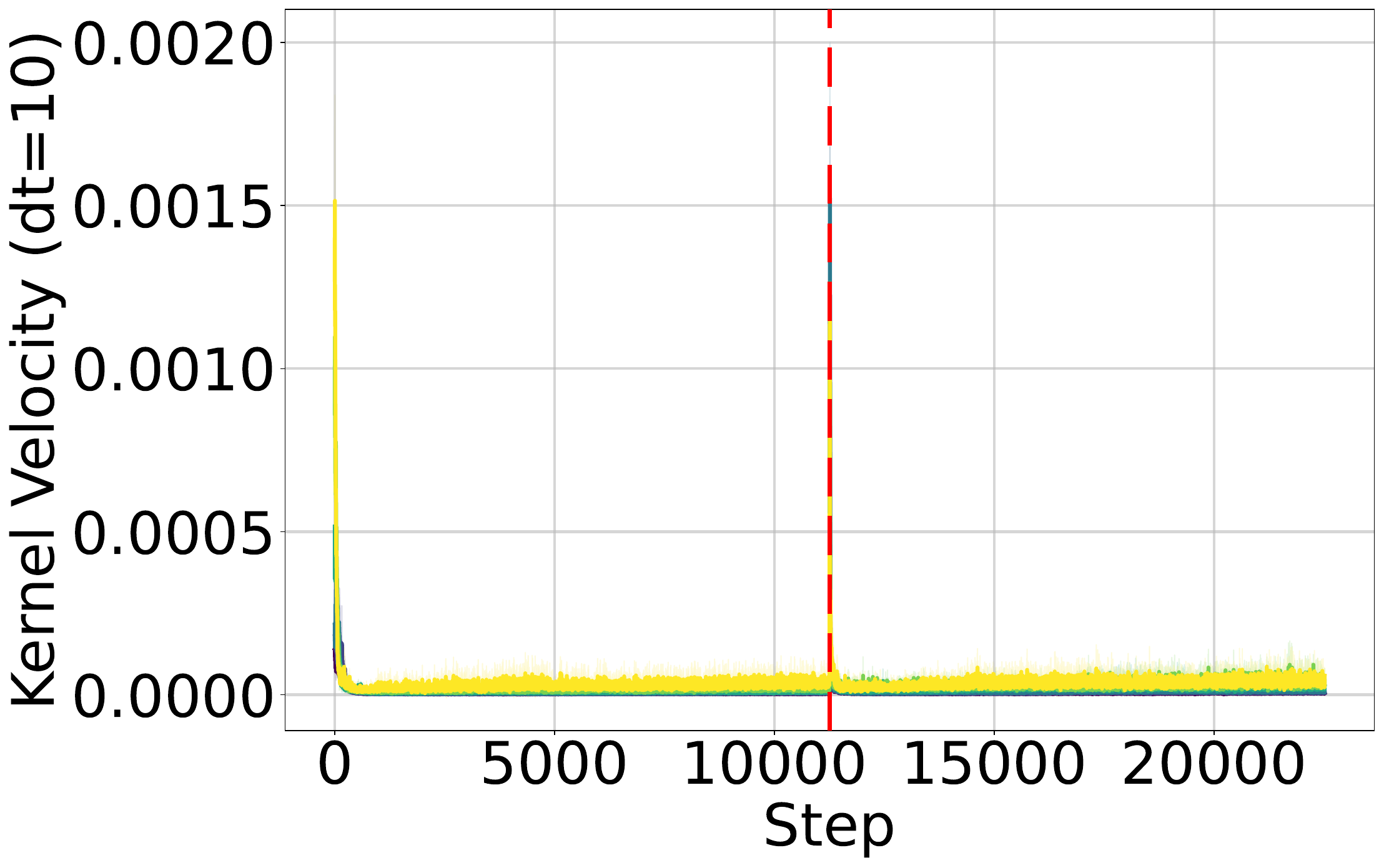}
    }
    \hspace{0.01\textwidth}
    \subfigure[Velocity (dt=10) Zoomed-in]{
        \includegraphics[height=0.2\textwidth]{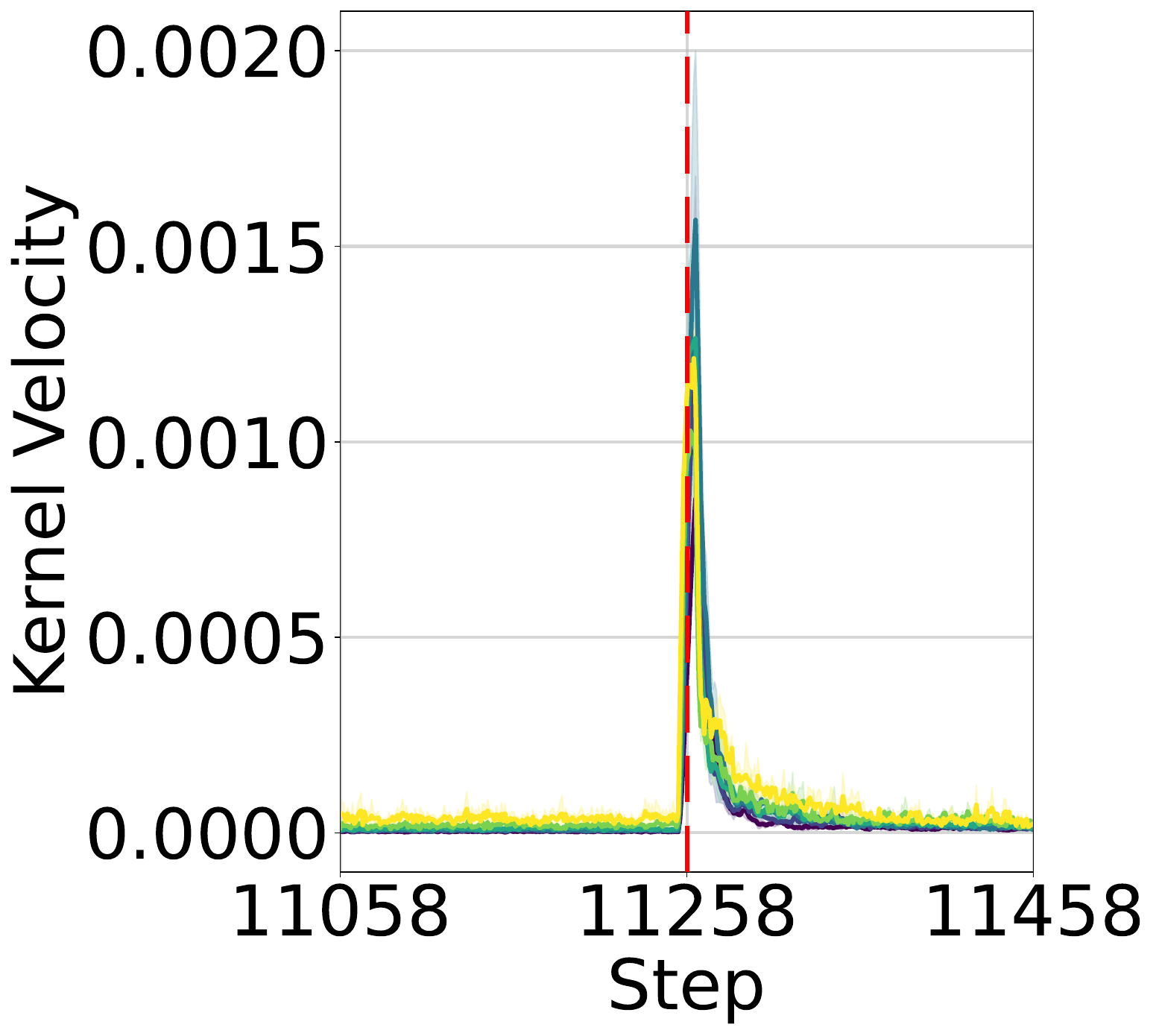}
    }

    \caption{Comparison of different metrics across network widths for CNN trained on CIFAR10 with learning rate 0.0001. The number of epochs per task is set to 160. (a) Test accuracy, (b) Alignment, (c) Kernel distance, (d) Maximum eigenvalue of NTK, (e) Kernel velocity with dt=10, (f) Kernel velocity (zoomed-in).}
    \label{fig:lr_0.0001}
\end{figure}

\begin{figure}[h!]
    \centering
    % 第一排
    \subfigure[Accuracy]{
        \includegraphics[height=0.21\textwidth]{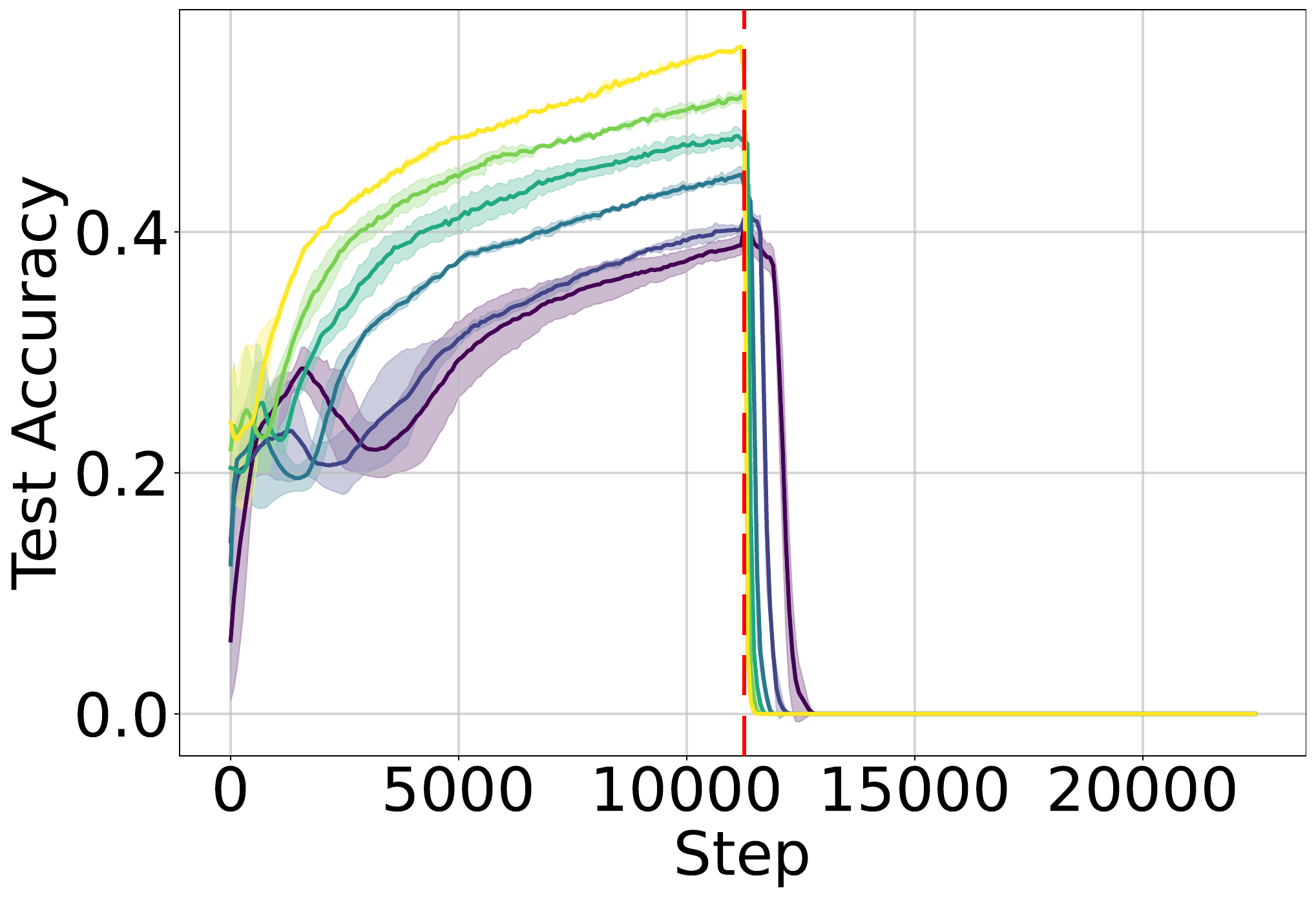}
    }
    \hfill
    \subfigure[Alignment]{
        \includegraphics[height=0.22\textwidth]{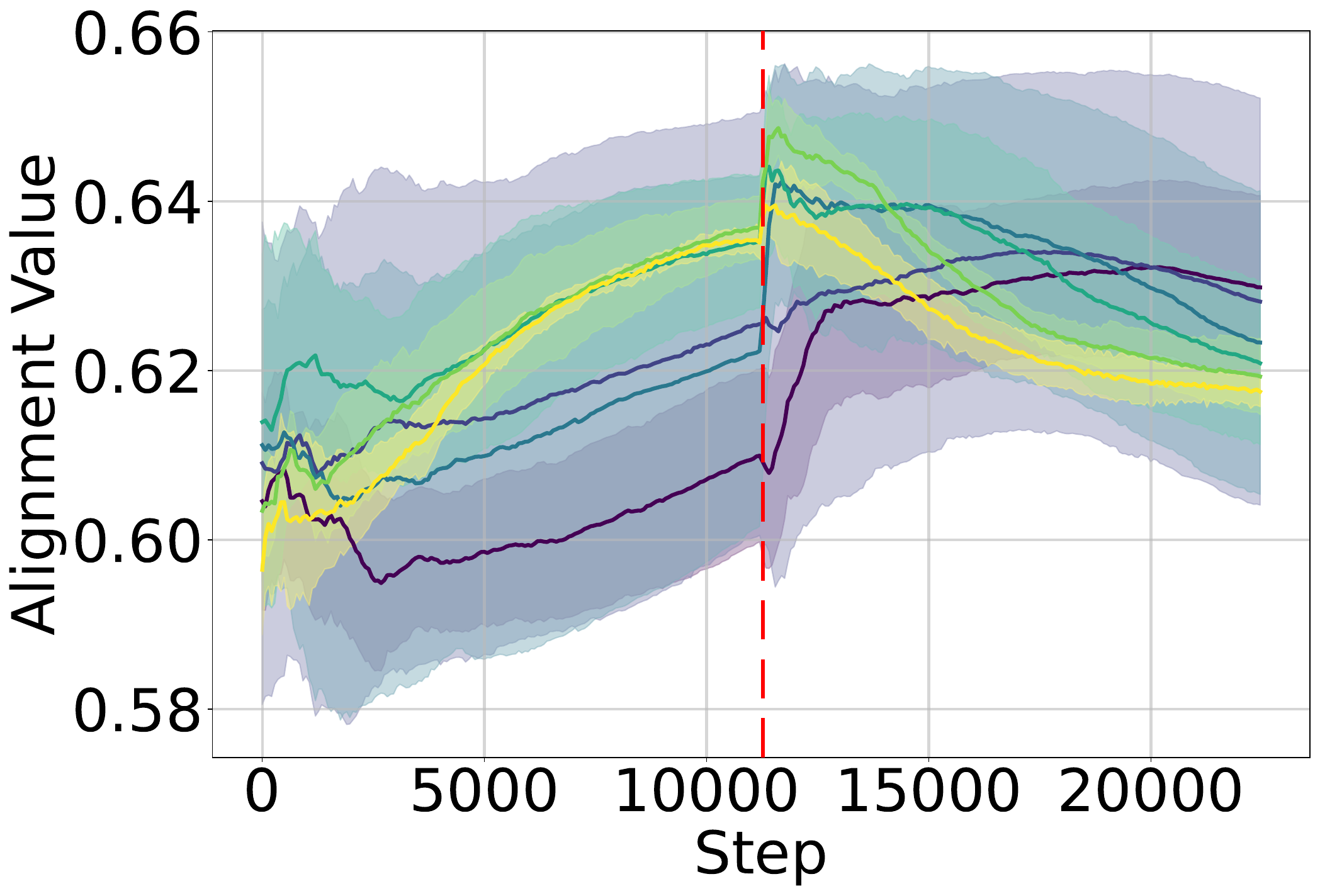}
    }
    \hfill
    \subfigure[Kernel Distance]{
        \includegraphics[height=0.21\textwidth]{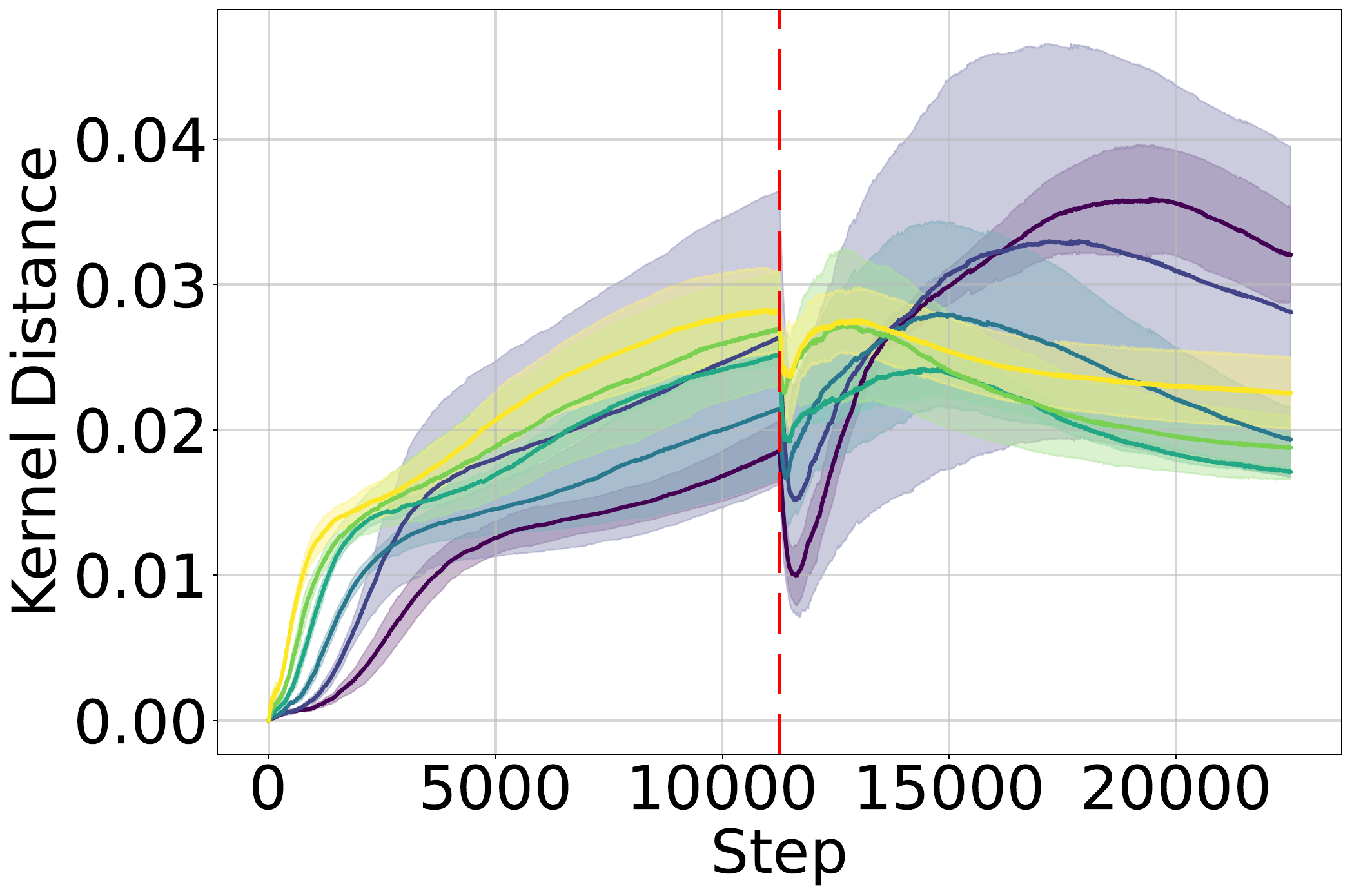}
    }

    % 第二排
    \subfigure[Max Eigenvalue]{
        \includegraphics[height=0.19\textwidth]{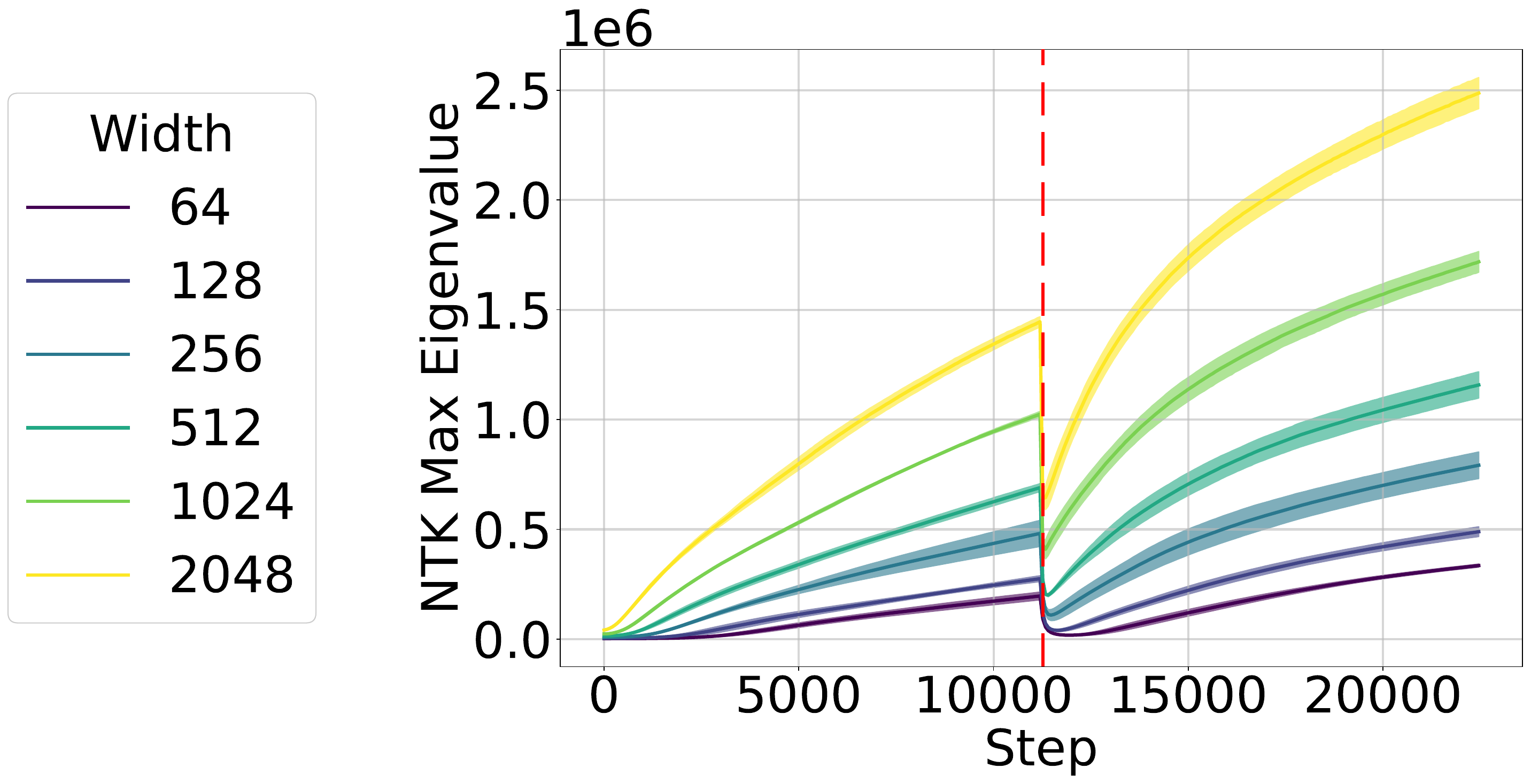}
    }
    \hspace{0.01\textwidth}
    \subfigure[Velocity (dt=10)]{
        \includegraphics[height=0.19\textwidth]{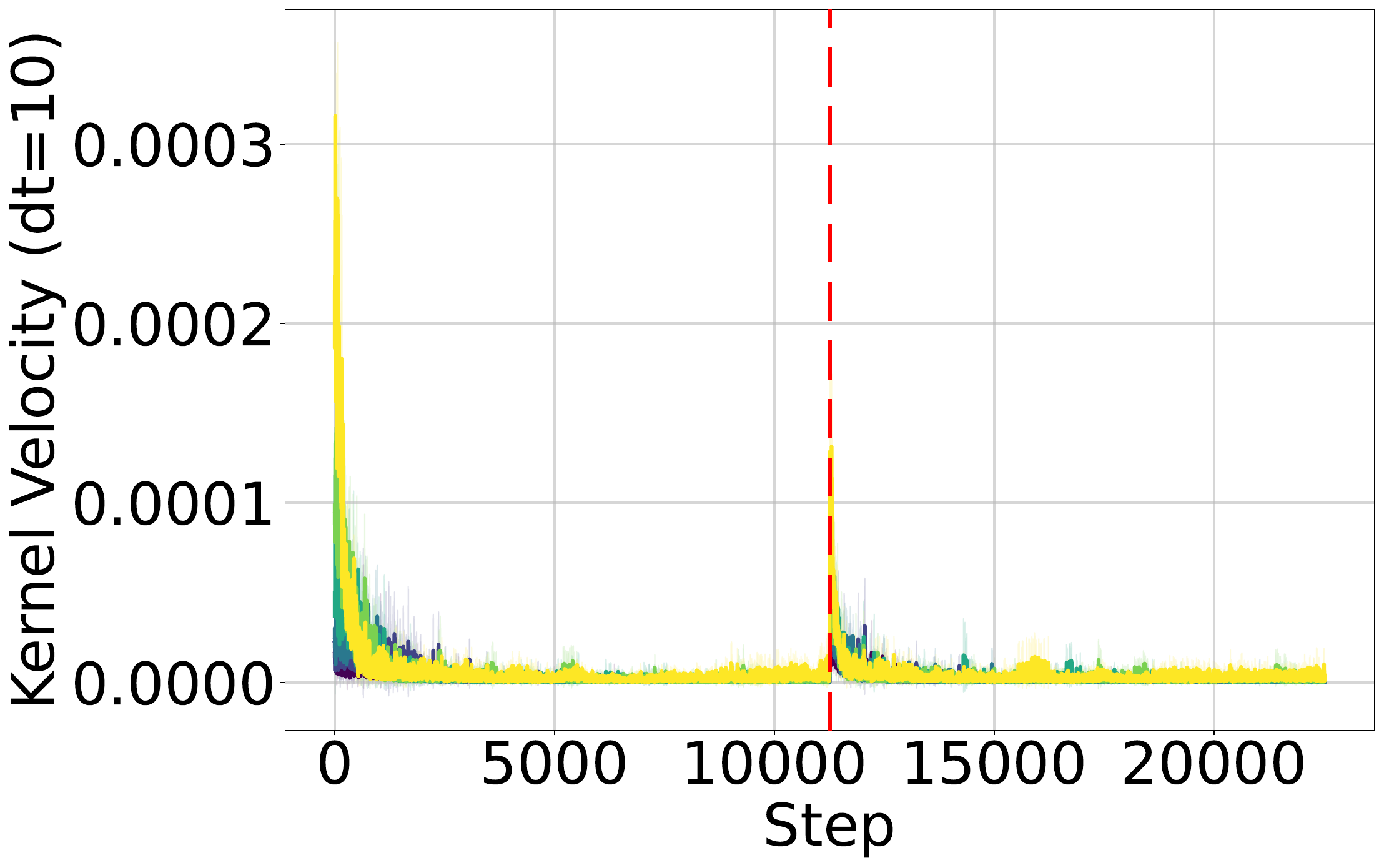}
    }
    \hspace{0.01\textwidth}
    \subfigure[Velocity (dt=10) Zoomed-in]{
        \includegraphics[height=0.2\textwidth]{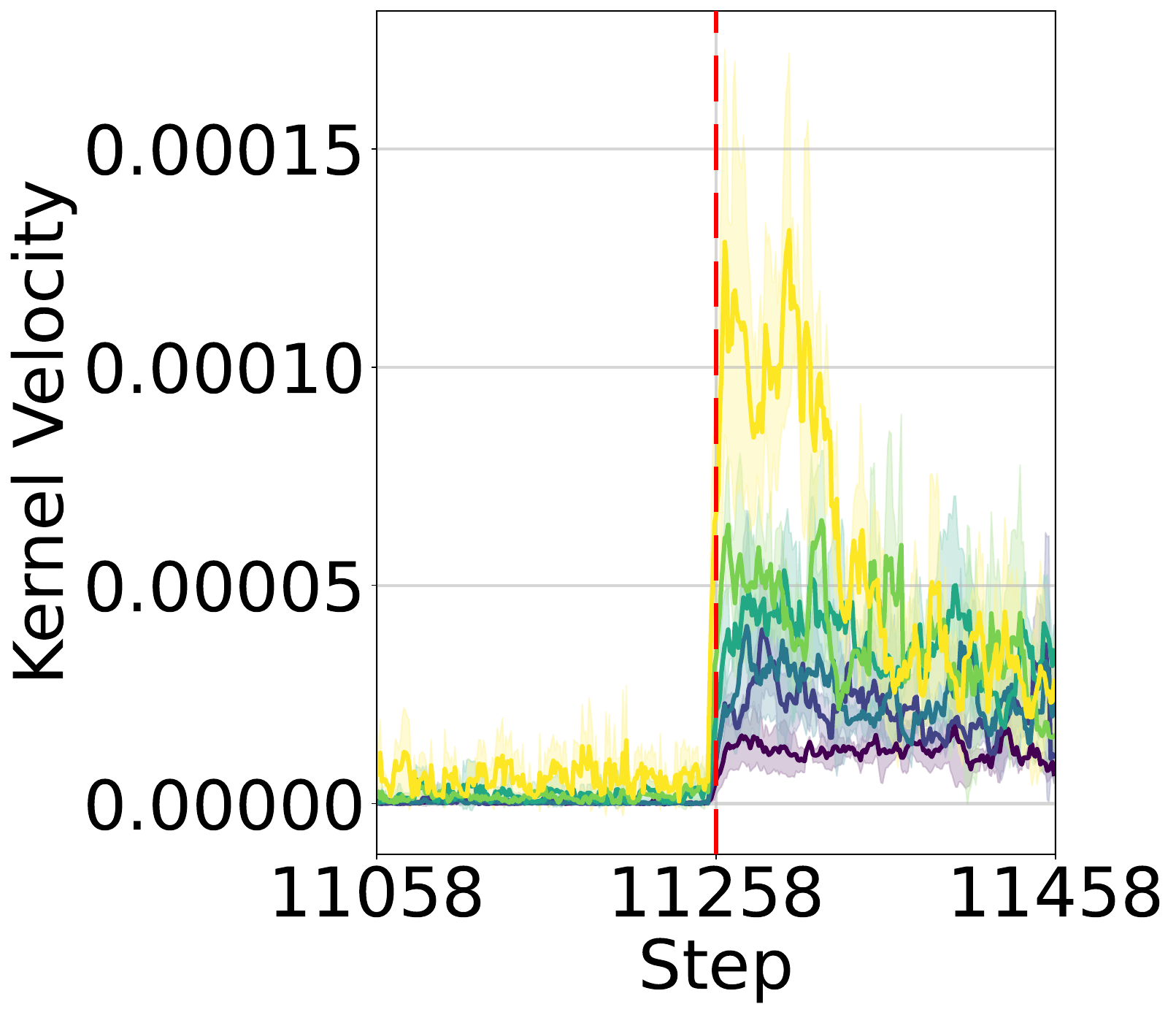}
    }

    \caption{Comparison of different metrics across network widths for CNN trained on CIFAR10 with learning rate 0.00001. The number of epochs per task is set to 160. (a) Test accuracy, (b) Alignment, (c) Kernel distance, (d) Maximum eigenvalue of NTK, (e) Kernel velocity with dt=10, (f) Kernel velocity (zoomed-in).}
    \label{fig:lr_1e5}
\end{figure}

In Figures \ref{fig:lr_0.001}, \ref{fig:lr_0.0001}, and \ref{fig:lr_1e5}, we show various metrics calculated during the CIFAR10 experiment as described in \ref{subsec: task_shifts}. 

\subsection{Extension to Multiple Task Switches}
\begin{figure}[h!]
    \centering
    % First row Width comparision
    \subfigure[Accuracy]{
        \includegraphics[width=0.31\textwidth]{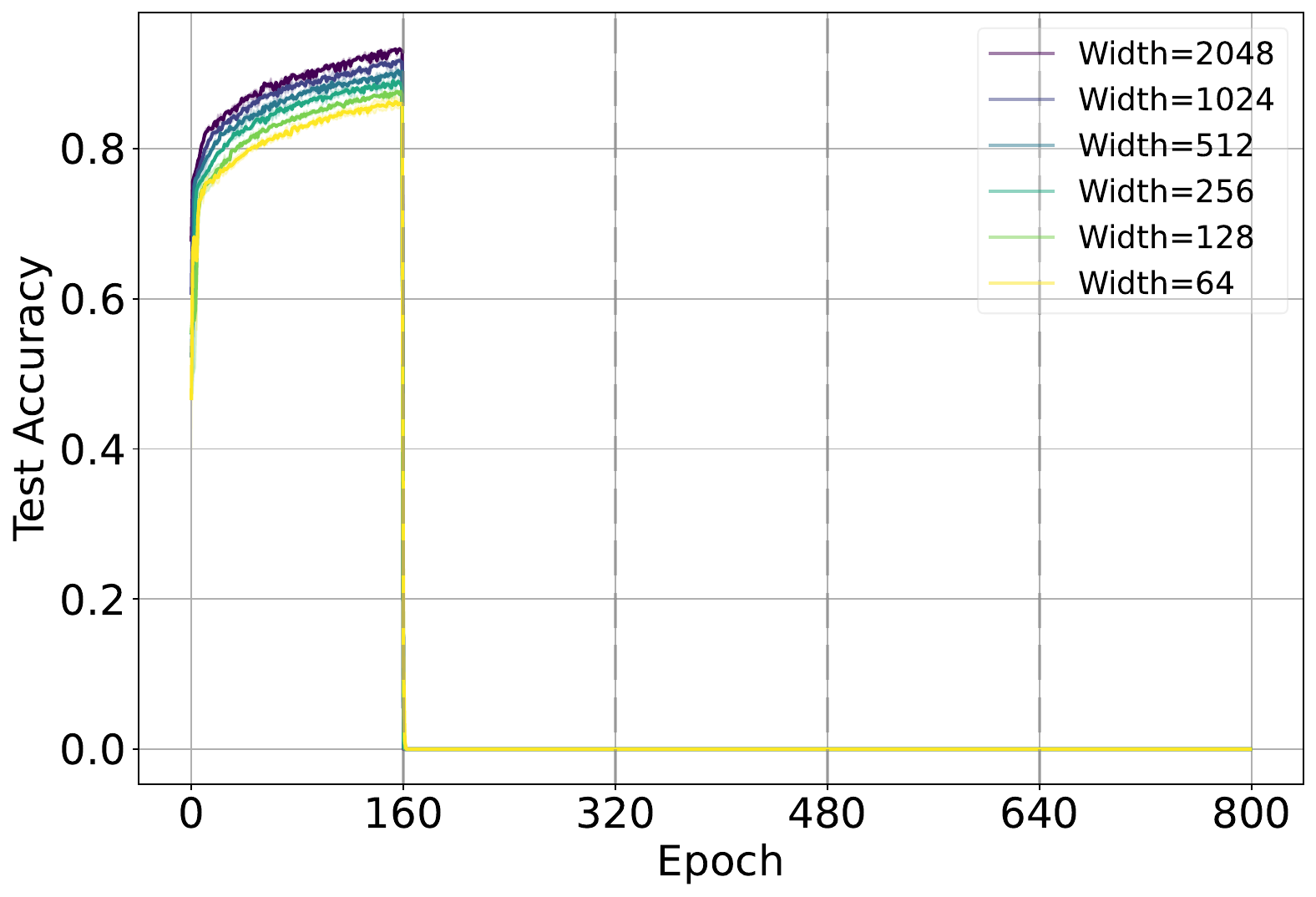}
    }
    \hfill
    \subfigure[Alignment]{
        \includegraphics[width=0.31\textwidth]{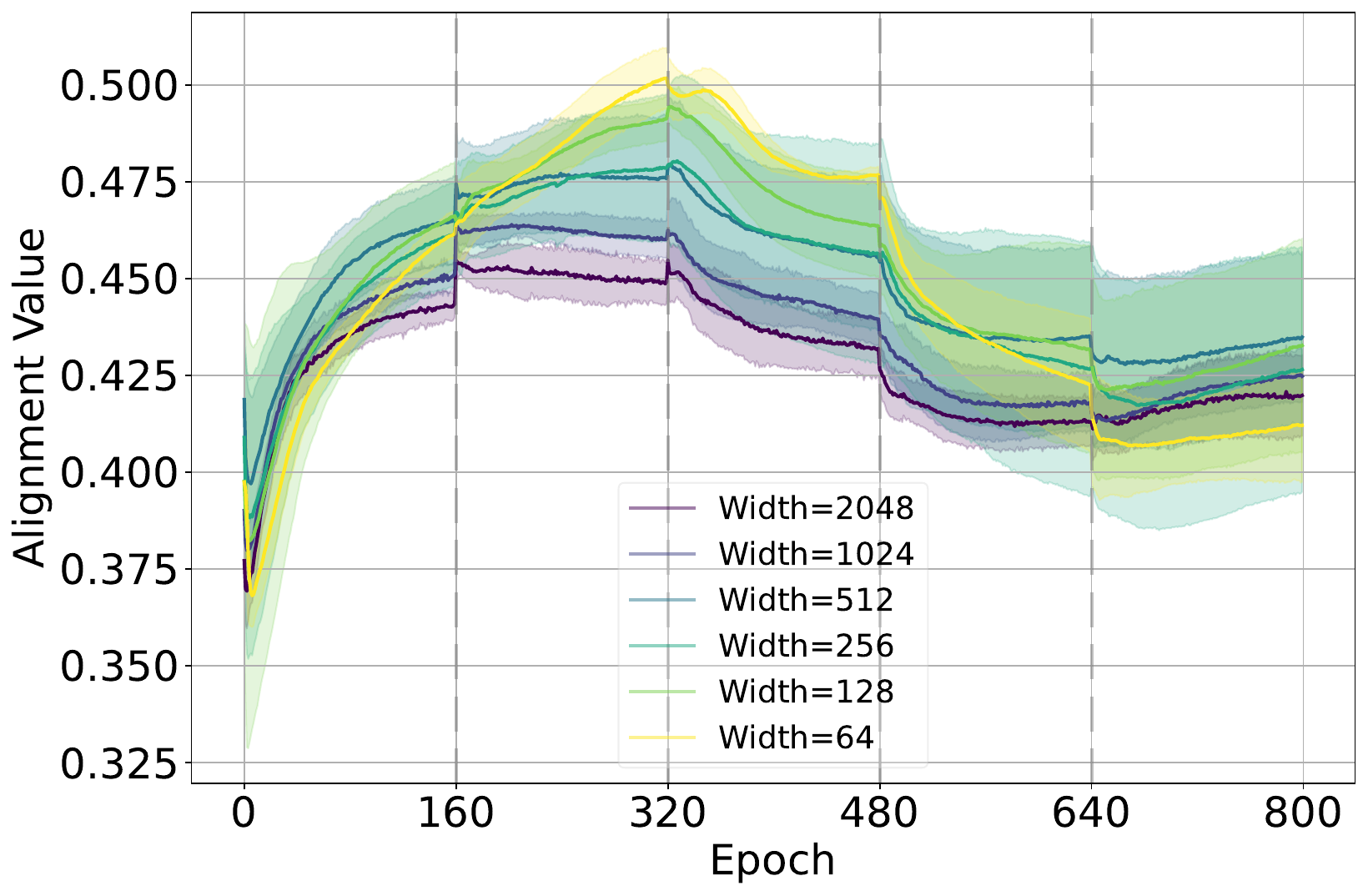}
    }
    \hfill
    \subfigure[Kernel Distance]{
        \includegraphics[width=0.31\textwidth]{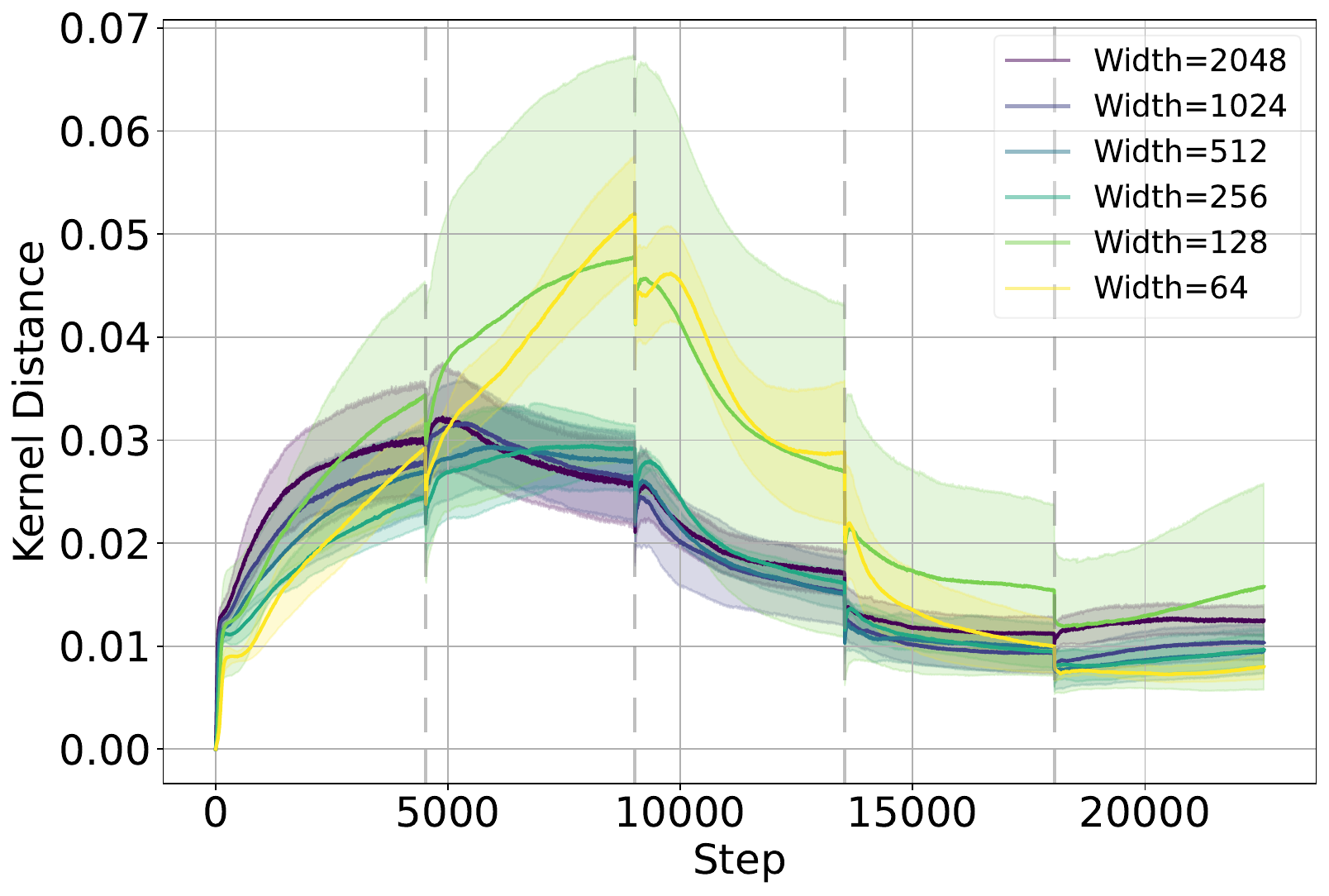}
    }

    % Second row
    \subfigure[Max Eigenvalue]{
        \includegraphics[width=0.42\textwidth]{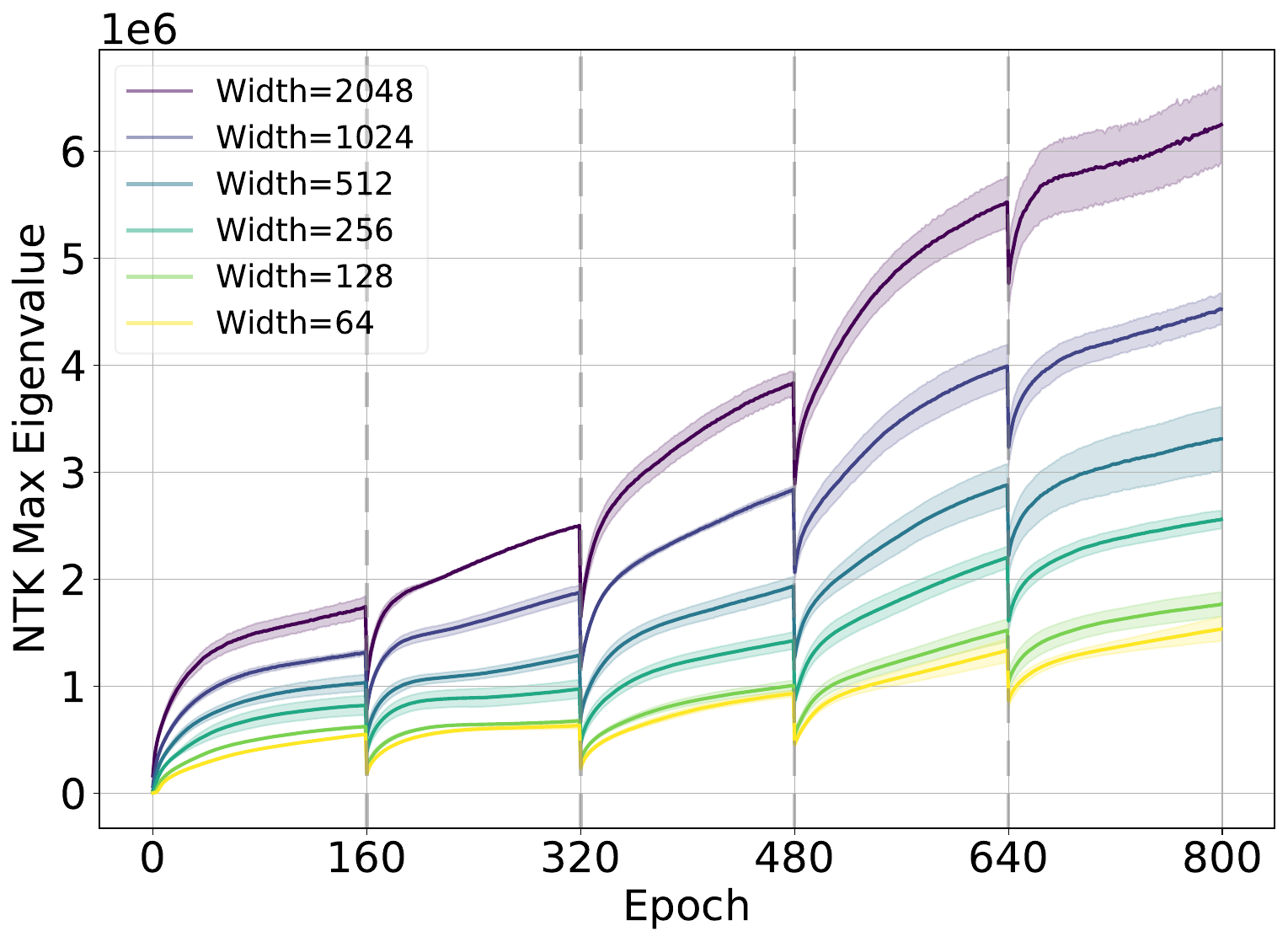}
    }
    \hfill
    \subfigure[Velocity (dt=10)]{
        \includegraphics[width=0.46\textwidth]{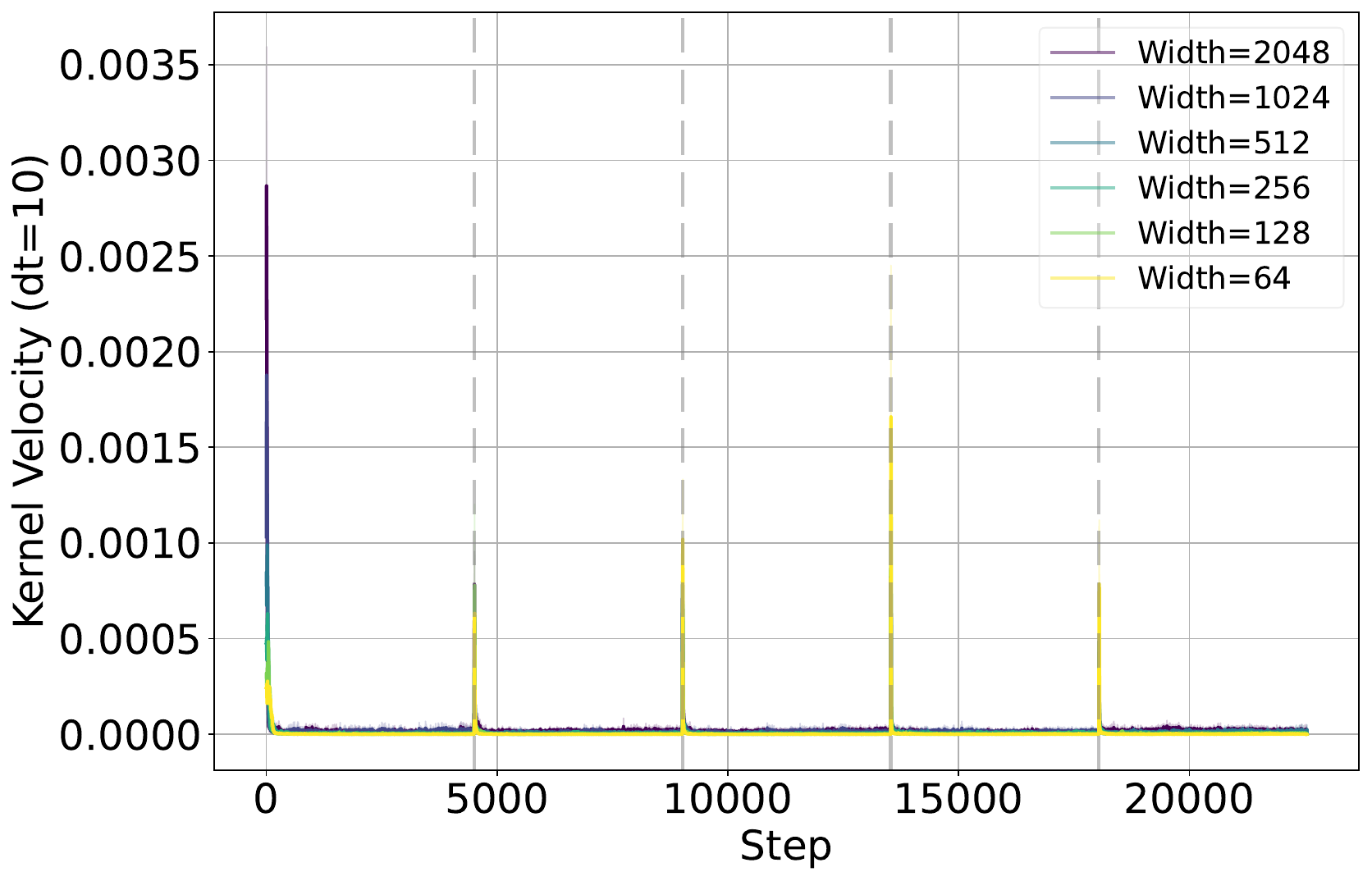}
    }
    % \subfigure[Velocity (dt=10) Zoomed-in]{
    %     \includegraphics[width=0.20\textwidth]{images/width_comparison_velocity_dt10_cnn_CIFAR10_inc2-2_e160_b32_lr0.0001_sgd_s32_zoom_in.pdf}
    
    \hfill
    \caption{Comparison of different metrics across network widths for CNN trained on CIFAR10 during multiple task switches. The number of epochs per task is set to 160. (a) Test accuracy, (b) Alignment, (c) Kernel distance, (d) Maximum eigenvalue of NTK, (e) Kernel
velocity with dt=10.}
    \label{fig:comparison_multiple_switches}
\end{figure}

To investigate whether the patterns persist during different task switches, we also perform experiments on 5 sequential tasks with 2 classes in each task on CIFAR-10 shown in Figure \ref{fig:comparison_multiple_switches}. 

\subsection{Experiments on ImageNet100}
\begin{figure}
    \centering
    % First row
    \subfigure[width 250]{
        \includegraphics[width=0.31\textwidth]{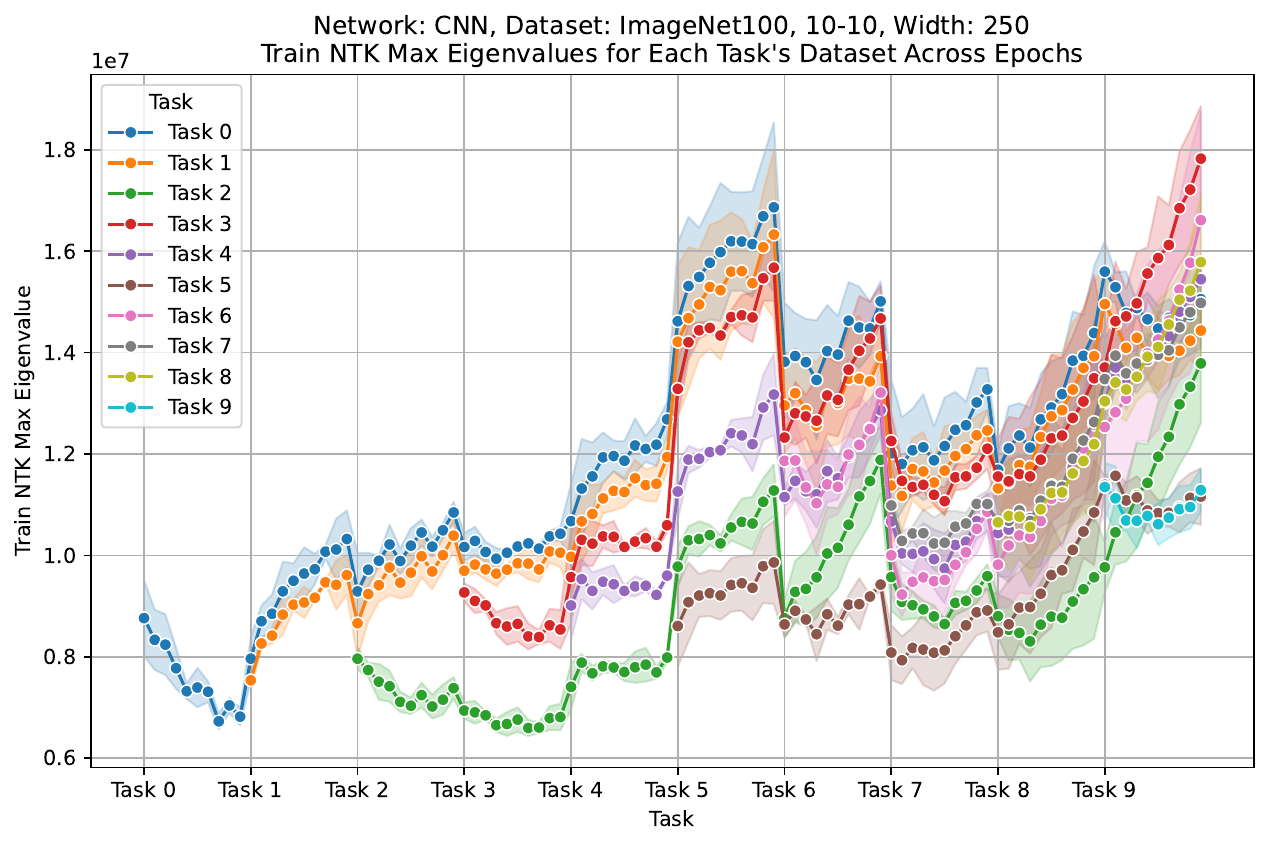}
    }
    \hfill
    \subfigure[width 500]{
        \includegraphics[width=0.31\textwidth]{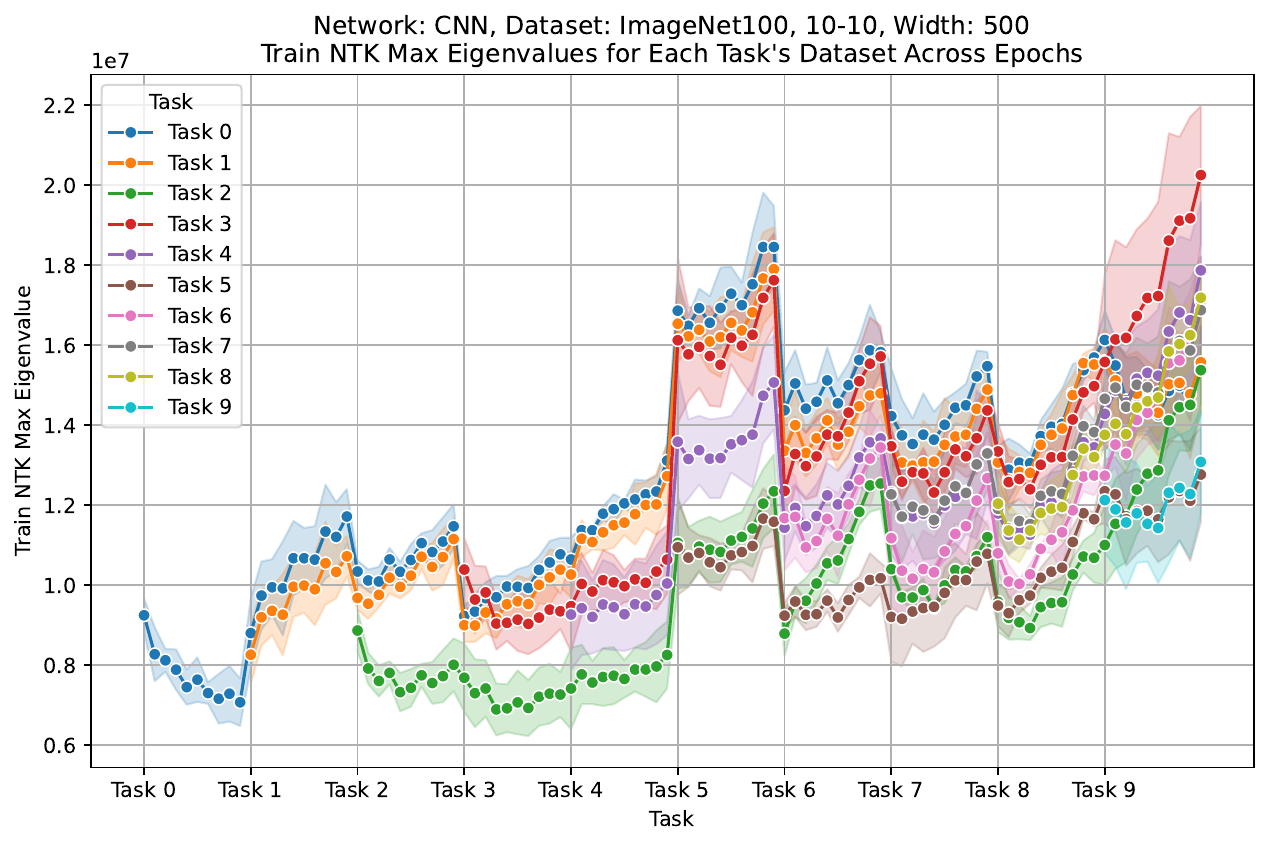}
    }
    \hfill
    \subfigure[width 1000]{
        \includegraphics[width=0.31\textwidth]{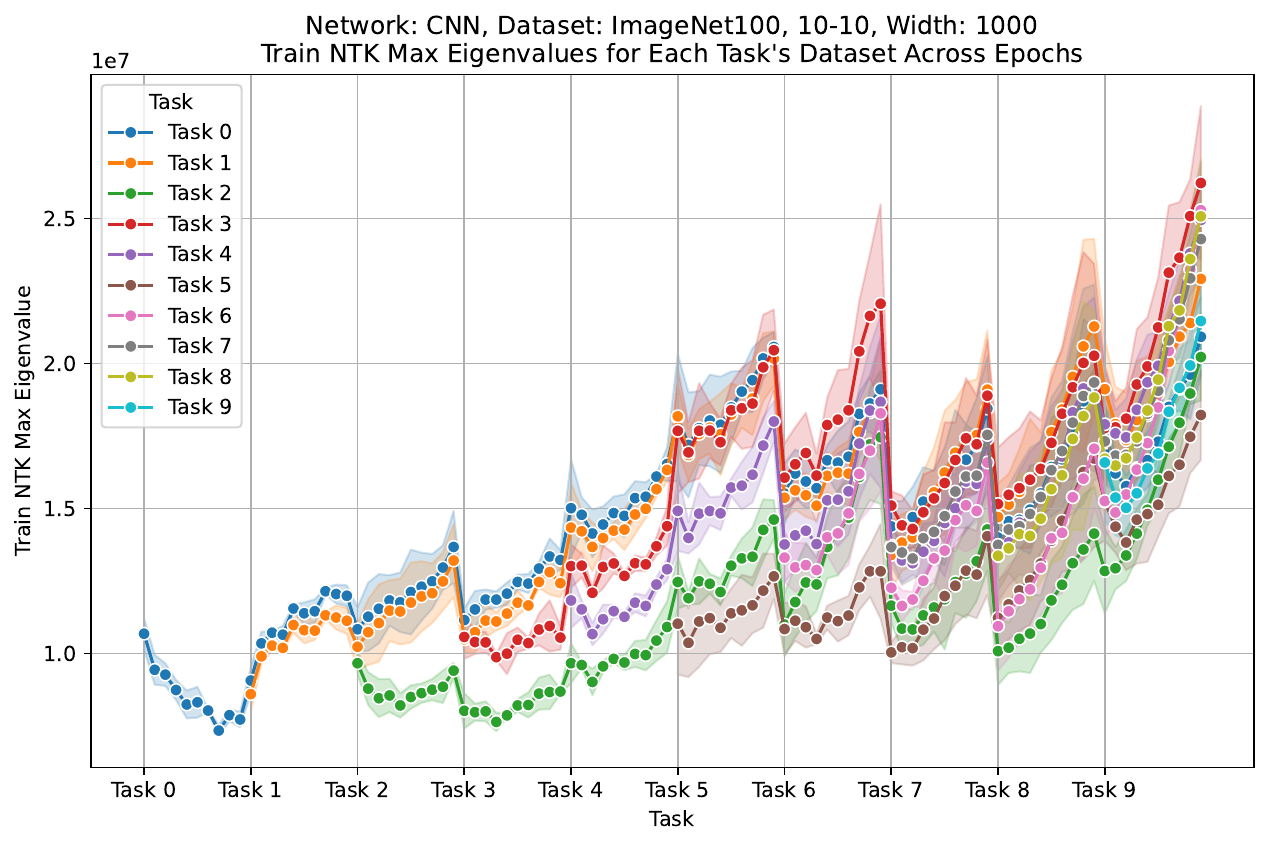}
    }

    % Second row
    \subfigure[10 epochs]{
        \includegraphics[width=0.31\textwidth]{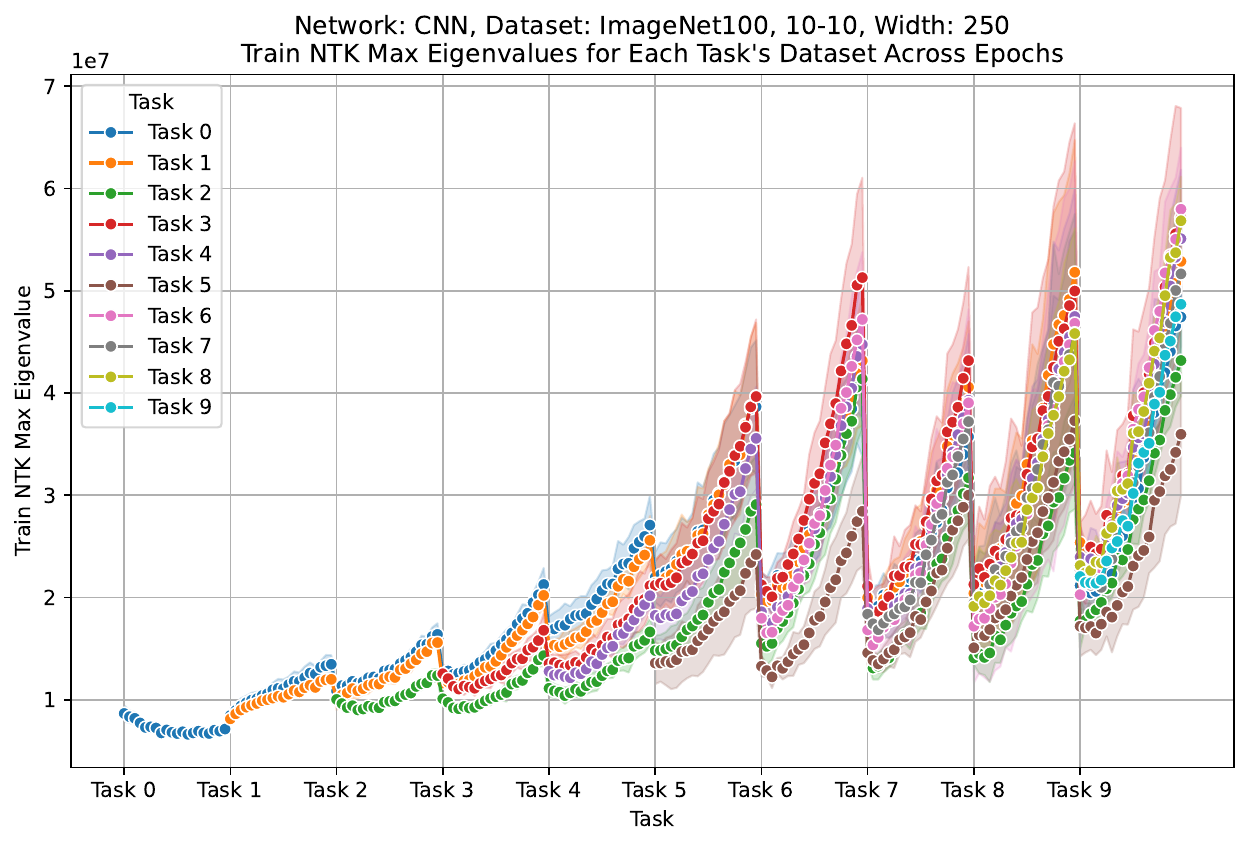}
    }
    \hfill
    \subfigure[20 epochs]{
        \includegraphics[width=0.31\textwidth]{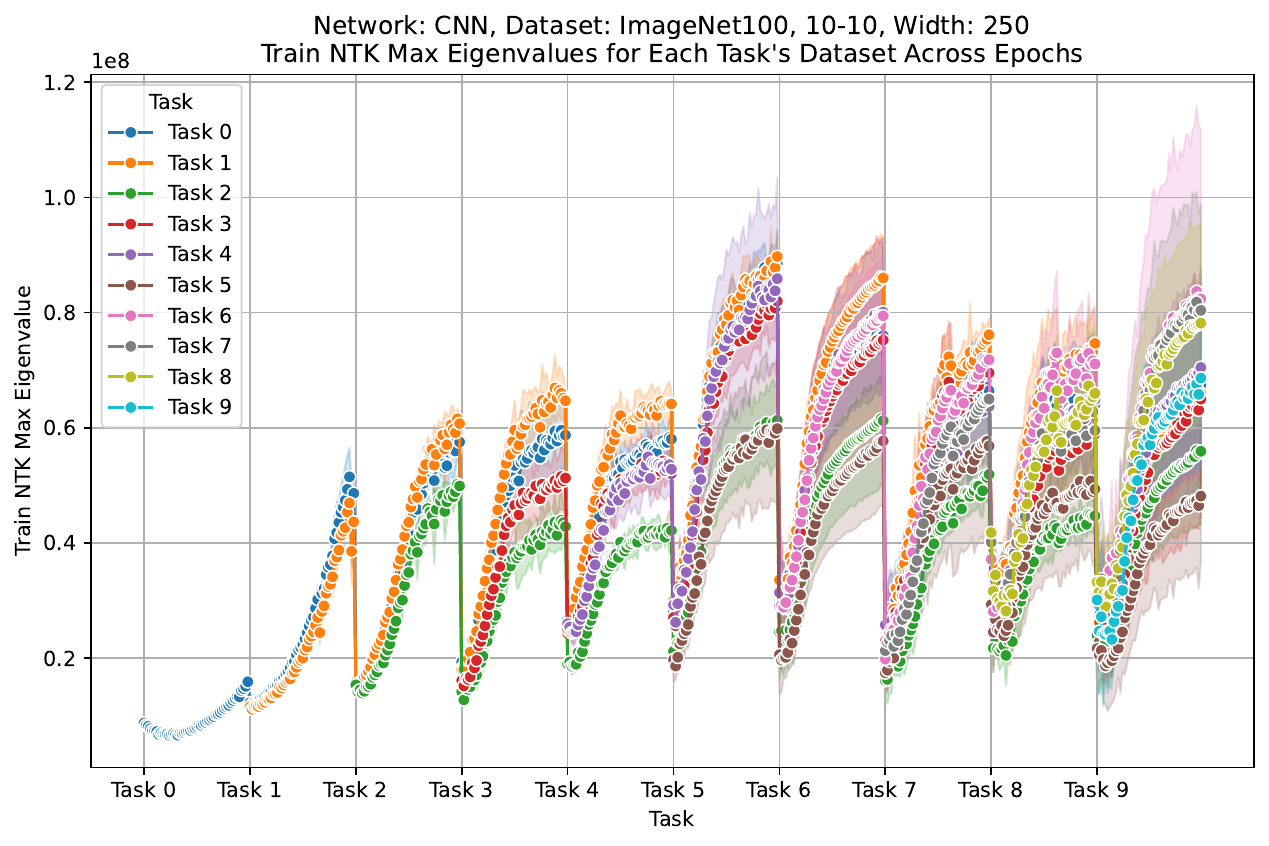}
    }
    \hfill
    \subfigure[100 epochs]{
        \includegraphics[width=0.31\textwidth]{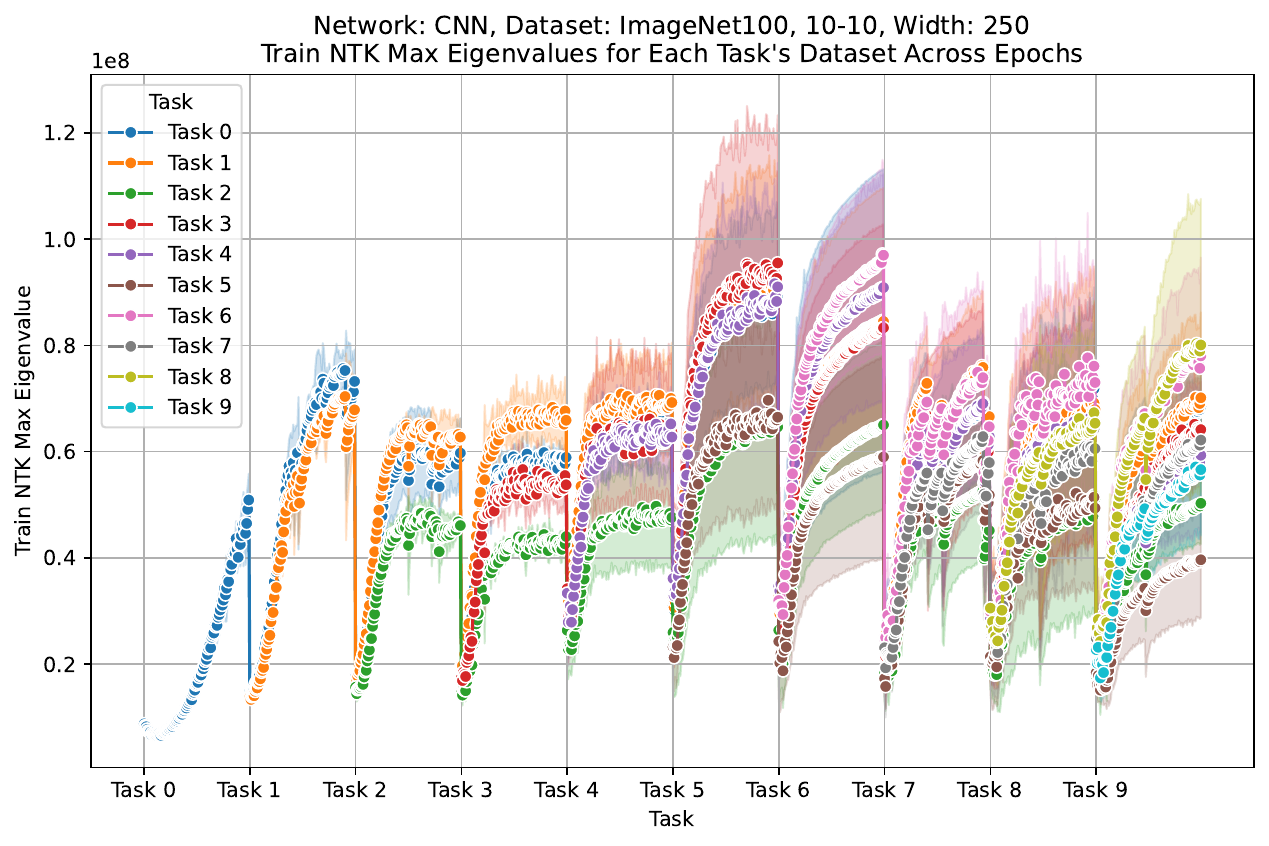}
    }

    \caption{ The effect of width and number of epochs per task on NTK spectrum on ImageNet100. The first row, left to right width $250$, $500$ and $1000$ respectively, the second row, left to right epoch number per task $20$, $50$ and $100$ respectively.}\label{fig:imagenet}
\end{figure}

%\lyz{[add trend]}
To further support our conclusions, we analyzed the evolution of the NTK spectrum on a larger dataset, ImageNet100. In Figure \ref{fig:imagenet}, we compare the effects of varying network width and the number of epochs per task on the NTK spectrum. Similarly to the experiments on CIFAR 10, our results show a significant change in NTK dynamics upon the arrival of new tasks. This change does get smoother as the network width expands, however it does not disapear. Also, no matter whether the old task is trained to full convergence (evaluated by the change in NTK), the re-activation upon task switch persists. 

All experiments use a learning rate of $1 \times 10^{-3}$ and SGD as the optimizer. Comparisons are made relative to the base setting: width $250$ and $10$ epochs per task.  

\subsection{Additional lazy regime results}

% \begin{figure}[h!]
%     \centering
    
%     % First row
%     \subfigure[Accuracy]{
%         \includegraphics[width=0.31\textwidth]{images/kaiming/width_comparison_accuracy_cnn_CIFAR10_inc5-5_e160_b32_kaiming_sgd_s32.pdf}
%     }
%     \subfigure[Velocity (dt=10)]{
%         \includegraphics[width=0.31\textwidth]{images/kaiming/width_comparison_velocity_dt10_cnn_CIFAR10_inc5-5_e160_b32_kaiming_sgd_s32.pdf}
%     }

%     \caption{Additional metrics across network widths for CNN trained on CIFAR10 The number of epochs per task is set to 160. (a) Test accuracy, (b)) Kernel velocity with dt=10. \giulia{Plot the widths in the reverse order, such that we see the larger widths on top. Also, which parametrization is this?}\zixuan{Do you mean drawing the color corresponding to the smaller width on the bottom layer? Re-uploaded.}}
%     \label{fig:kaiming_5}
% \end{figure}

% In order to guarantee stability as we scale the model width we adopt the NTK parametrization using Kaiming Normal initialization and scale the learning rate with respect to width with $0.1/width$ to ensure stable training \cite{park2019effectnetworkwidthstochastic}.

\begin{figure}[h!]
    \centering
    % 第一排
    \subfigure[Accuracy]{
        \includegraphics[height=0.21\textwidth]{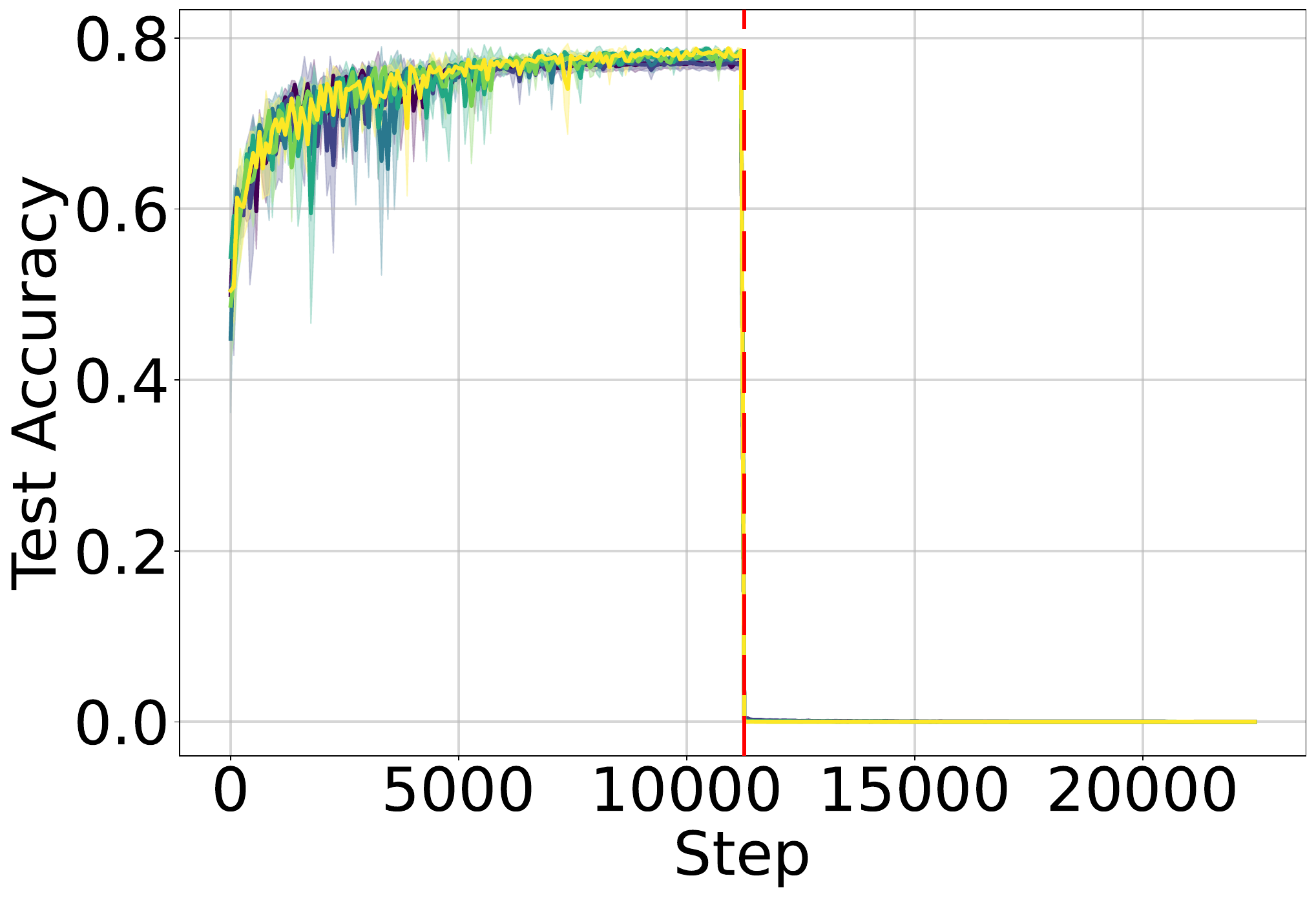}
    }
    \hfill
    \subfigure[Alignment]{
        \includegraphics[height=0.2\textwidth]{images/kaiming/width_comparison_alignment_cnn_CIFAR10_inc5-5_e160_b32_kaiming_sgd_s32.pdf}
    }
    \hfill
    \subfigure[Kernel Distance]{
        \includegraphics[height=0.21\textwidth]{images/kaiming/width_comparison_cka_cnn_CIFAR10_inc5-5_e160_b32_kaiming_sgd_s32.pdf}
    }

    % 第二排
    \subfigure[Max Eigenvalue]{
        \includegraphics[height=0.19\textwidth]{images/kaiming/width_comparison_ntk_max_eigenvalues_cnn_CIFAR10_inc5-5_e160_b32_kaiming_sgd_s32.pdf}
    }
    \hspace{0.01\textwidth}
    \subfigure[Velocity (dt=10)]{
        \includegraphics[height=0.19\textwidth]{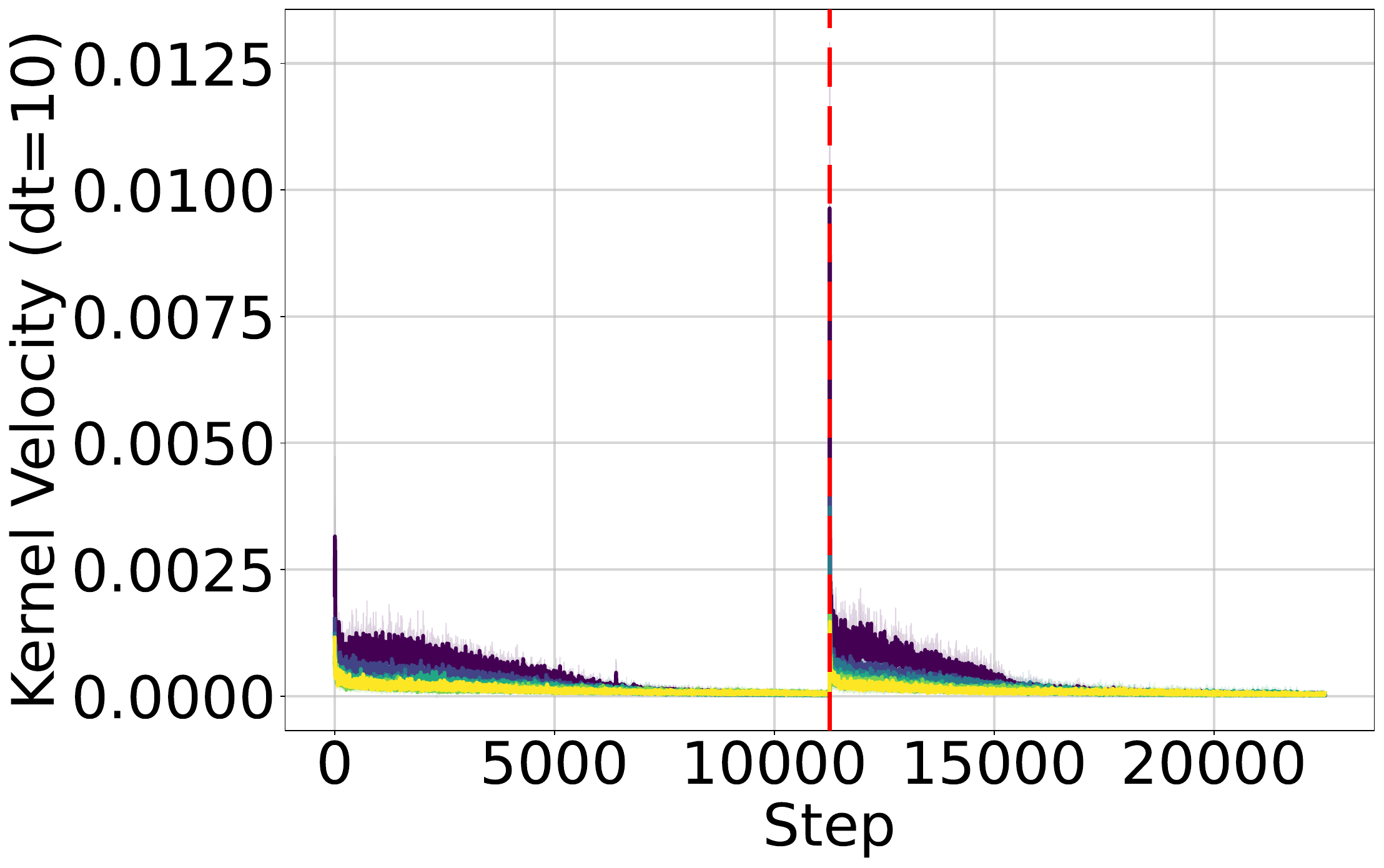}
    }
    \hspace{0.01\textwidth}
    \subfigure[Velocity (dt=10) Zoomed-in]{
        \includegraphics[height=0.2\textwidth]{images/kaiming/width_comparison_velocity_dt10_cnn_CIFAR10_inc5-5_e160_b32_kaiming_sgd_s32_zoom_in.pdf}
    }

    \caption{Comparison of different metrics across network widths for CNN trained on CIFAR10 with NTK parametrization. The number of epochs per task is set to 160. (a) Test accuracy, (b) Alignment, (c) Kernel distance, (d) Maximum eigenvalue of NTK, (e) Kernel velocity with dt=10, (f) Kernel velocity (zoomed-in).}
    \label{fig:kaiming_5}
\end{figure}

The addition results presented in Figure \ref{fig:kaiming_5} demonstrate that as the network width increases exponentially from 64 to 2048, the magnitude of changes in test accuracy (a), alignment (b), kernel distance (c), and the maximum eigenvalue of the NTK (d) decreases during task transitions in continual learning. 

% test accuracy, kernel velocity
% different parameter setting
\newpage
\subsection{Figures}

% 学习率 0.001

\vspace{-5mm}
% 学习率 0.0001

\end{document}